\documentclass[10pt,english,british,twocolumn]{article}
\usepackage[T1]{fontenc}
\usepackage[latin9]{inputenc}
\usepackage{xcolor}
\usepackage{array}
\usepackage{enumitem}
\usepackage{multirow}
\usepackage{amsmath}
\usepackage{graphicx}
\PassOptionsToPackage{normalem}{ulem}
\usepackage{ulem}

\makeatletter

\providecommand{\tabularnewline}{\\}
\usepackage{cvpr}
\usepackage{times}
\usepackage{epsfig}
\usepackage{graphicx}
\usepackage{amsmath}
\usepackage{amssymb}

\usepackage{hyphenat}
\usepackage{pdflscape}

\usepackage{color}
\usepackage{colortbl}
\usepackage{rotating}
\definecolor{lightgray}{gray}{0.8}
\definecolor{verylightgray}{gray}{0.9}
\usepackage[pagebackref=true,breaklinks=true,letterpaper=true,colorlinks,bookmarks=false]{hyperref}

 \cvprfinalcopy % *** Uncomment this line for the final submission

\ifcvprfinal\fi

\@ifundefined{showcaptionsetup}{}{%
 \PassOptionsToPackage{caption=false}{subfig}}
\usepackage{subfig}
\makeatother

\usepackage{babel}
\begin{document}
% from http://tex.stackexchange.com/questions/4637/correct-use-of-paragraph-titles
\let\originalparagraph\paragraph 
\renewcommand{\paragraph}[2][.]{\originalparagraph{#2#1}}

\title{Simple Does It: Weakly Supervised Instance and Semantic Segmentation}

\author{Anna Khoreva\textsuperscript{1}\hspace{1.5em}Rodrigo Benenson\textsuperscript{1}\hspace{1.5em}Jan Hosang\textsuperscript{1}\hspace{1.5em}Matthias Hein\textsuperscript{2}\hspace{1.5em}Bernt
Schiele\textsuperscript{1}\\\\\textsuperscript{1}Max Planck Institute
for Informatics, Saarbr{\"u}cken, Germany\\ \textsuperscript{2}Saarland University,
Saarbr{\"u}cken, Germany}
\maketitle
\vspace{-0.5em}
% 
% \author{Anna Khoreva\textsuperscript{1}\hspace{0.3em}Rodrigo Benenson\textsuperscript{1}\hspace{0.3em}Jan Hosang\textsuperscript{1}\hspace{0.3em}Matthias Hein\textsuperscript{2}\hspace{0.3em}Bernt Schiele\textsuperscript{1}}
% 
% \institute{\textsuperscript{1}Max Planck Institute for Informatics, Saarbr{\"u}cken,
% Germany\\ \textsuperscript{2}Saarland University, Saarbr{\"u}cken,
% Germany}
\maketitle
\begin{abstract}
Semantic labelling and instance segmentation are two tasks that require
particularly costly annotations. Starting from weak supervision in
the form of bounding box detection annotations, we propose a new approach
that does not require modification of the segmentation training procedure.
We show that when carefully designing the input labels from given
bounding boxes, even a single round of training is enough to improve
over previously reported weakly supervised results. Overall, our weak
supervision approach reaches $\sim\!95\%$ of the quality of the fully
supervised model, both for semantic labelling and instance segmentation.
\end{abstract}

\section{\label{sec:Introduction}Introduction}

Convolutional networks (convnets) have become the de facto technique
for pattern recognition problems in computer vision. One of their
main strengths is the ability to profit from extensive amounts of
training data to reach top quality. However, one of their main weaknesses
is that they need a large number of training samples for high quality
results. This is usually mitigated by using pre-trained models (e.g.
with $\sim\negmedspace10^{6}$ training samples for ImageNet classification
\cite{Russakovsky2015Ijcv}), but still thousands of samples are needed
to shift from the pre-training domain to the application domain. Applications
such as semantic labelling (associating each image pixel to a given
class) or instance segmentation (grouping all pixels belonging to
the same object instance) are expensive to annotate, and thus significant
cost is involved in creating large enough training sets.

Compared to object bounding box annotations, pixel-wise mask annotations
are far more expensive, requiring $\sim\negmedspace15\times$ more
time \cite{Lin2014EccvCoco}. Cheaper and easier to define, box annotations
are more pervasive than pixel-wise annotations. In principle, a large
number of box annotations (and images representing the background
class) should convey enough information to understand which part of
the box content is foreground and which is background. In this paper
we explore how much one can close the gap between training a convnet
using full supervision for semantic labelling (or instance segmentation)
versus using only bounding box annotations.

Our experiments focus on the $20$ Pascal classes \cite{Everingham15}
and show that using only bounding box annotations over the same training
set we can reach $\sim\negmedspace95\%$ of the accuracy achievable
with full supervision. We show top results for (bounding box) weakly
supervised semantic labelling and, to the best of our knowledge, for
the first time report results for weakly supervised instance segmentation.

\begin{figure}
\begingroup
\setlength{\tabcolsep}{0.15em} \hspace*{\fill}%
\begin{tabular}{ccc}
\includegraphics[width=0.32\columnwidth,height=0.09\textheight]{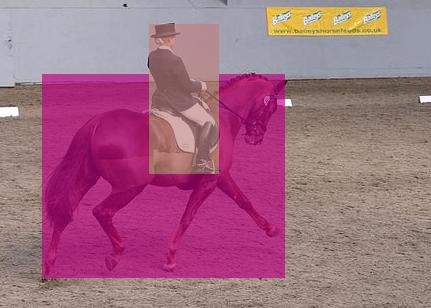} & \includegraphics[width=0.32\columnwidth,height=0.09\textheight]{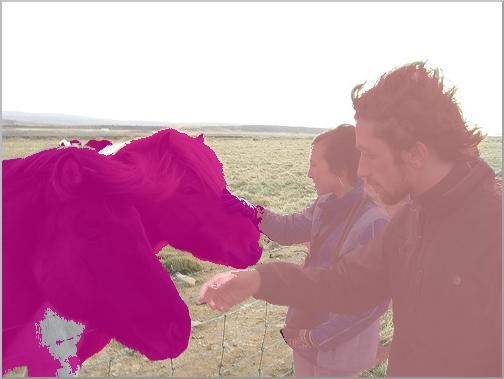} & \includegraphics[width=0.32\columnwidth,height=0.09\textheight]{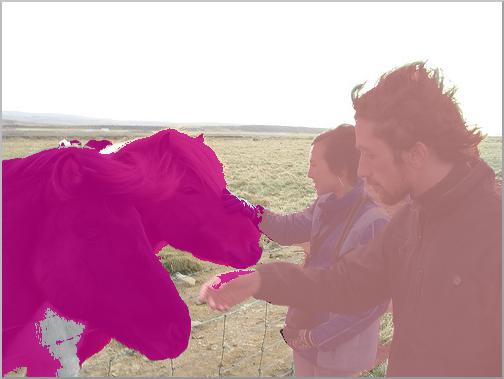}\tabularnewline
\textcolor{black}{\small{}}%
\begin{tabular}{c}
\textcolor{black}{\small{}Training sample,}\tabularnewline
\textcolor{black}{\small{}with box annotations}\tabularnewline
\end{tabular}\textcolor{black}{\small{} } & \textcolor{black}{\small{}}%
\begin{tabular}{c}
\textcolor{black}{\small{}Test image, fully }\tabularnewline
\textcolor{black}{\small{}supervised result}\tabularnewline
\end{tabular}\textcolor{black}{\small{} } & \textcolor{black}{\small{}}%
\begin{tabular}{c}
\textcolor{black}{\small{}Test image, weakly }\tabularnewline
\textcolor{black}{\small{}supervised result}\tabularnewline
\end{tabular}\textcolor{black}{\small{} }\tabularnewline
\end{tabular}\endgroup\hspace*{\fill}
\begin{centering}
\vspace{0em}
\par\end{centering}
\caption{\label{fig:teaser}We propose a technique to train semantic labelling
from bounding boxes, and reach $95\%$ of the quality obtained when
training from pixel-wise annotations.}
\vspace{-1em}
\end{figure}
We view the problem of weak supervision as an issue of input label
noise. We explore recursive training as a de-noising strategy, where
convnet predictions of the previous training round are used as supervision
for the next round. We also show that, when properly used, ``classic
computer vision'' techniques for box-guided instance segmentation
are a source of surprisingly effective supervision for convnet training.

In summary, our main contributions are:
\begin{itemize}[label=$-$,itemsep=0em]
\item We explore recursive training of convnets for weakly supervised semantic
labelling, discuss how to reach good quality results, and what are
the limitations of the approach (Section \ref{subsec:Box-baselines}).
\item We show that state of the art quality can be reached when properly
employing GrabCut-like algorithms to generate training labels from
given bounding boxes, instead of modifying the segmentation convnet
training procedure (Section \ref{subsec:GrabCut-baselines}).
\item We report the best known results when training using bounding boxes
only, both using Pascal VOC12 and VOC12+COCO training data, reaching
comparable quality with the fully supervised regime (Section \ref{subsec:Main-results}).
\item We are the first to show that similar results can be achieved for
the weakly supervised instance segmentation task (Section \ref{sec:Instance-segmentation-results}).
\end{itemize}

\section{\label{subsec:Related-work}Related work}

\paragraph{Semantic labelling}

Semantic labelling may be tackled via decision forests \cite{Shotton2009Ijcv}
or classifiers over hand-crafted superpixel features \cite{Gould2009Iccv}.
However, convnets have proven particularly effective for semantic
labelling. A flurry of variants have been proposed recently \cite{Pinheiro2014Icml,Long2015Cvpr,Chen2015Iclr,Lin2016CvprAdelaide,Zheng2015IccvCrfAsRnn,Kokkinos2016Iclr,Yu2016Iclr}.
In this work we use Deep\-Lab \cite{Chen2015Iclr} as our reference
implementation. This network achieves state-of-the-art performance
on the Pascal VOC12 semantic segmentation benchmark and the source
code is available online. \\
Almost all these methods include a post-processing step to enforce
a spatial continuity prior in the predicted segments, which provides
a non-negligible improvement on the results ($2\!\sim\!5$ points).
The most popular technique is DenseCRF \cite{Kraehenbuehl2011Nips},
but other variants are also considered \cite{Kolmogorov2004Pami,Barron2015ArXiv}.
\vspace{-1em}

\paragraph{Weakly supervised semantic labelling}

In order to keep annotation cost low, recent work has explored different
forms of supervision for semantic labelling: image labels \cite{Pathak2015Iclrw,Pathak2015Iccv,Papandreou2015Iccv,Pinheiro2015Cvpr,Wei2015ArXiv},
points \cite{Bearman2015ArXiv}, scribbles \cite{Xu2015CvprWeakSegmentation,Lin2016CvprScribbleSup},
and bounding boxes~\cite{Dai2015Iccv,Papandreou2015Iccv}. \cite{Dai2015Iccv,Papandreou2015Iccv,Hong2015Nips}
also consider the case where a fraction of images are fully supervised.
\cite{Xu2015CvprWeakSegmentation} proposes a framework to handle
all these types of annotations.\\
In this work we focus on box level annotations for semantic labelling
of objects. The closest related work are thus \cite{Dai2015Iccv,Papandreou2015Iccv}.
BoxSup \cite{Dai2015Iccv} proposes a recursive training procedure,
where the convnet is trained under supervision of segment object proposals
and the updated network in turn improves the segments used for training.
WSSL \cite{Papandreou2015Iccv} proposes an expectation-maximisation
algorithm with a bias to enable the network to estimate the foreground
regions. We compare with these works in the result sections. \textcolor{black}{Since
all implementations use slightly different networks and training procedures,
care should be taken during comparison.} Both \cite{Dai2015Iccv}
and \cite{Papandreou2015Iccv} propose new ways to train convnets
under weak supervision. In contrast, in this work we show that one
can reach better results without modifying the training procedure
(compared to the fully supervised case) by instead carefully generating
input labels for training from the bounding box annotations (Section
\ref{sec:Baselines-and-approach}).

\paragraph{Instance segmentation}

In contrast to instance agnostic semantic labelling that groups pixels
by object class, instance segmentation groups pixels by object instance
and ignores classes.\\
Object proposals \cite{PontTuset2015Iccv,Hosang2015Pami} that generate
segments (such as \cite{PontTuset2015ArxivMcg,Krahenbuhl2015Cvpr})
can be used for instance segmentation. Similarly, given a bounding
box (e.g. selected by a detector), GrabCut \cite{Rother2004TogGrabcut}
variants can be used to obtain an instance segmentation (e.g. \cite{Lempitsky2009Iccv,Cheng2015CgfDenseCut,Taniai2015Cvpr,Tang2015IccvSecretGrabCut,Yu2015ArXivLooseCut}).\\
To enable end-to-end training of detection and segmentation systems,
it has recently been proposed to train convnets for the task of instance
segmentation \cite{Hariharan2015Cvpr,Pinheiro2015Nips}. In this work
we explore weakly supervised training of an instance segmentation
convnet. We use DeepMask \cite{Pinheiro2015Nips} as a reference implementation
for this task. In addition we re-purpose Deep\-Lab\-v2 network \cite{Chen2016ArxivDeeplabv2},
originally designed for semantic segmentation, for the instance segmentation
task.

\section{\label{sec:Baselines-and-approach}From boxes to semantic labels}

The goal of this work is to provide high quality semantic labelling
starting from object bounding box annotations. We design our approach
aiming to exploit the available information at its best. There are
two sources of information: the annotated boxes and priors about the
objects. We integrate these in the following cues:

\paragraph{C1 Background}

Since the bounding boxes are expected to be exhaustive, any pixel
not covered by a box is labelled as background.

\paragraph{C2 Object~extend}

The box annotations bound the extent of each instance. Assuming a
prior on the objects shapes (e.g. oval-shaped objects are more likely
than thin bar or full rectangular objects), the box also gives information
on the expected object area. We employ this size information during
training.

\paragraph{C3 Objectness}

Other than extent and area, there are additional object priors at
hand. Two priors typically used are spatial continuity and having
a contrasting boundary with the background. In general we can harness
priors about object shape by using segment proposal techniques \cite{PontTuset2015Iccv},
which are designed to enumerate and rank plausible object shapes in
an area of the image.

\subsection{\label{subsec:Box-baselines}Box baselines}

We first describe a naive baseline that serves as starting point for
our exploration. Given an annotated bounding box and its class label,
we label all pixels inside the box with such given class. If two boxes
overlap, we assume the smaller one is in front. Any pixel not covered
by boxes is labelled as background.

\begin{figure*}
\begin{centering}
\begin{tabular}{ccccc}
\includegraphics[width=0.18\textwidth]{figures/teaser/2008_003379_rect_and_input_crop.jpg} & \includegraphics[width=0.18\textwidth]{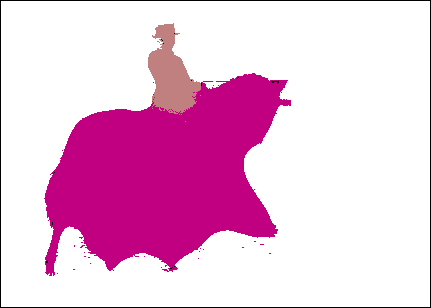}\hspace*{-0.5em} & \includegraphics[width=0.18\textwidth]{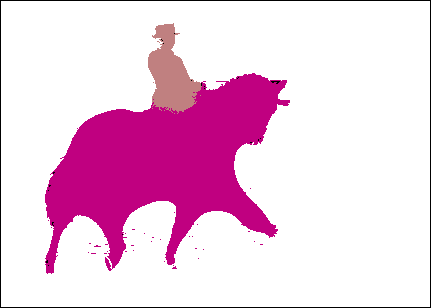}\hspace*{-0.5em} & \includegraphics[width=0.18\textwidth]{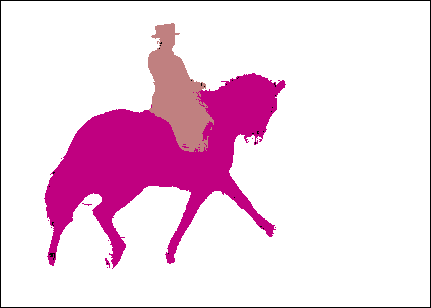} & \includegraphics[width=0.18\textwidth]{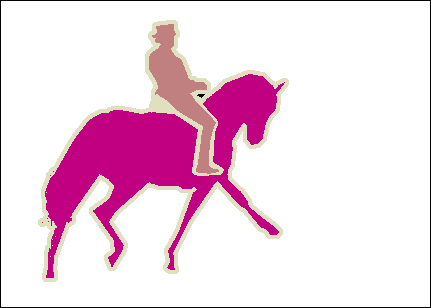}\tabularnewline
Example & Output after & After & After & Ground\tabularnewline
input rectangles & 1 training round & 5 rounds & 10 rounds & truth\tabularnewline
\end{tabular}\vspace{-0.5em}
\par\end{centering}
\caption{\label{fig:rectangles-training-progress}Example results of using
only rectangle segments and recursive training (using convnet predictions
as supervision for the next round), see Section \ref{subsec:Box-baselines}.}
\vspace{-1em}
\end{figure*}

Figure \ref{fig:rectangles-training-progress} left side and Figure
\ref{fig:grabcut-variant-rectangle} show such example annotations.
We use these labels to train a segmentation network with the standard
training procedure. We employ the Deep\-Lab\-v1 approach from \cite{Chen2015Iclr}
(details in Section \ref{subsec:labelling-experimental-setup}). 

\paragraph{Recursive training}

We observe that when applying the resulting model over the training
set, the network outputs capture the object shape significantly better
than just boxes (see Figure \ref{fig:rectangles-training-progress}).
This inspires us to follow a recursive training procedure, where these
new labels are fed in as ground truth for a second training round.
We name this recursive training approach \texttt{Naive}. 

The recursive training is enhanced by de-noising the convnet outputs
using extra information from the annotated boxes and object priors.
Between each round we improve the labels with three post-processing
stages:
\begin{enumerate}
\item Any pixel outside the box annotations is reset to background label
(cue C1).
\item If the area of a segment is too small compared to its corresponding
bounding box (e.g. IoU$<50\%$), the box area is reset to its initial
label (fed in the first round). This enforces a minimal area (cue
C2).
\item As it is common practice among semantic labelling methods, we filter
the output of the network to better respect the image boundaries.
(We use DenseCRF \cite{Kraehenbuehl2011Nips} with the Deep\-Lab\-v1
parameters \cite{Chen2015Iclr}). In our weakly supervised scenario,
boundary-aware filtering is particularly useful to improve objects
delineation (cue C3).
\end{enumerate}
The recursion and these three post-processing stages are crucial to
reach good performance. We name this recursive training approach \texttt{Box},
and show an example result in Figure \ref{fig:rectangles-training-progress}.

\paragraph{Ignore regions}

We also consider a second variant $\mathtt{Box^{i}}$ that, instead
of using filled rectangles as initial labels, we fill in the 20\%
inner region, and leave the remaining inner area of the bounding box
as ignore regions. See Figure \ref{fig:grabcut-variant-bbox20}. Following
cues C2 and C3 (shape and spatial continuity priors), the 20\% inner
box region should have higher chances of overlapping with the corresponding
object, reducing the noise in the generated input labels. The intuition
is that the convnet training might benefit from trading-off lower
recall (more ignore pixels) for higher precision (more pixels are
correctly labelled). Starting from this initial input, we use the
same recursive training procedure as for \texttt{Box}.

Despite the simplicity of the approach, as we will see in the experimental
section \ref{sec:Semantic-labelling-results}, $\mathtt{Box}$ / $\mathtt{Box^{i}}$
is already competitive with the current state of the art.

However, using rectangular shapes as training labels is clearly suboptimal.
Therefore, in the next section, we propose an approach that obtains
better results while avoiding multiple recursive training rounds.

\subsection{\label{subsec:GrabCut-baselines}Box-driven segments}

The box baselines are purposely simple. A next step in complexity
consists in utilising the box annotations to generate an initial guess
of the object segments. We think of this as ``old school meets new
school'': we use the noisy outputs of classic computer vision methods,
box-driven figure-ground segmentation \cite{Rother2004TogGrabcut}
and object proposal \cite{PontTuset2015Iccv} techniques, to feed
the training of a convnet. Although the output object segments are
noisy, they are more precise than simple rectangles, and thus should
provide improved results. A single training round will be enough to
reach good quality.

\subsubsection{GrabCut baselines}

GrabCut \cite{Rother2004TogGrabcut} is the established technique
to estimate an object segment from its bounding box. We propose to
use a modified version of GrabCut, which we call $\mathtt{GrabCut+}$,
where HED boundaries \cite{Xie2015Iccv} are used as pairwise term
instead of the typical RGB colour difference. (The HED boundary detector
is trained on the generic boundaries of BSDS500 \cite{ArbelaezMaireFowlkesMalikPAMI11}).
We considered other GrabCut variants, such as \cite{Cheng2015CgfDenseCut,Tang2015IccvSecretGrabCut};
however, the proposed $\mathtt{GrabCut+}$ gives higher quality segments
(see supplementary material).\\
Similar to $\mathtt{Box^{i}}$, we also consider a $\mathtt{GrabCut+^{i}}$
variant, which trades off recall for higher precision. For each annotated
box we generate multiple ($\sim\negmedspace150$) perturbed $\mathtt{GrabCut+}$\texttt{
}outputs. If $70\%$ of the segments mark the pixel as foreground,
the pixel is set to the box object class. If less than $20\%$ of
the segments mark the pixels as foreground, the pixel is set as background,
otherwise it is marked as ignore. The perturbed outputs are generated
by jittering the box coordinates ($\pm5\%$) as well as the size of
the outer background region considered by GrabCut (from $10\%$ to
$60\%$). An example result of $\mathtt{GrabCut+^{i}}$ can be seen
in Figure \ref{fig:grabcut-variant-grabcut+-perturbed}.

\subsubsection{Adding objectness}

With our final approach we attempt to better incorporate the object
shape priors by using segment proposals \cite{PontTuset2015Iccv}.
Segment proposals techniques are designed to generate a soup of likely
object segmentations, incorporating as many ``objectness'' priors
as useful (cue C3). 

We use the state of the art proposals from MCG \cite{PontTuset2015ArxivMcg}.
As final stage the MCG algorithm includes a ranking based on a decision
forest trained over the Pascal VOC 2012 dataset. We do \emph{not}
use this last ranking stage, but instead use all the (unranked) generated
segments. Given a box annotation, we pick the highest overlapping
proposal as a corresponding segment.

Building upon the insights from the baselines in Section \ref{subsec:Box-baselines}
and \ref{subsec:GrabCut-baselines}, we use the MCG segment proposals
to supplement $\mathtt{GrabCut+}$. Inside the annotated boxes, we
mark as foreground pixels where both MCG and $\mathtt{GrabCut+}$
agree; the remaining ones are marked as ignore. We denote this approach
as $\mathtt{MCG}\cap\mathtt{GrabCut+}$ or $\mathtt{M}\cap\mathtt{G+}$
for short.

\begin{figure*}[t]
\hspace*{\fill}\subfloat[\label{fig:grabcut-input-image}Input image]{\centering{}\includegraphics[width=0.14\textwidth,height=0.13\textheight]{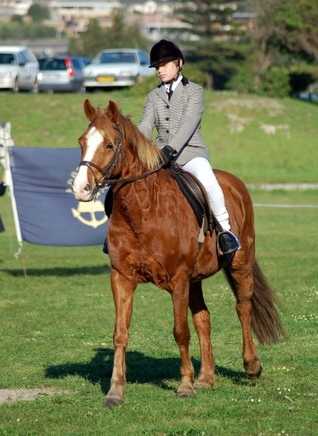}}\hspace*{\fill}
\subfloat[\label{fig:grabcut-gt-image} Ground~truth]{\centering{}\includegraphics[width=0.14\textwidth,height=0.13\textheight]{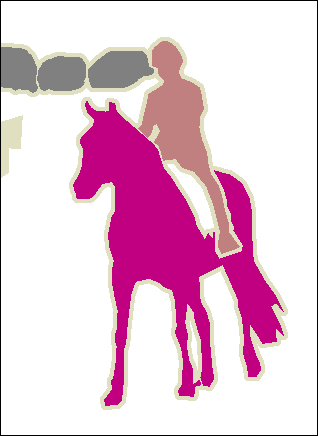}}\hspace*{\fill}
\subfloat[\label{fig:grabcut-variant-rectangle}$\mathtt{Box}$]{\centering{}\includegraphics[width=0.14\textwidth,height=0.13\textheight]{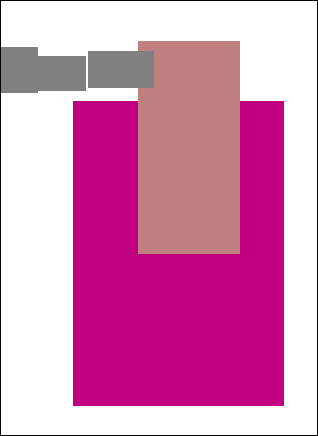}}\hspace*{\fill}
\subfloat[\label{fig:grabcut-variant-bbox20}$\mathtt{Box^{i}}$]{\centering{}\includegraphics[width=0.14\textwidth,height=0.13\textheight]{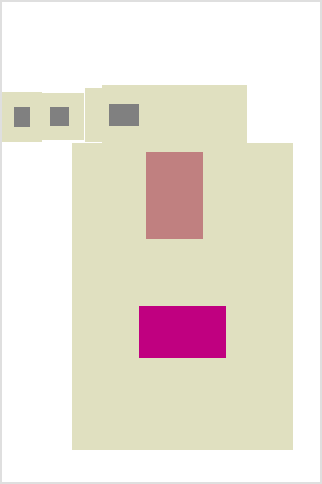}}\hspace*{\fill}\vspace{-0.5em}

\hspace*{\fill}\subfloat[\label{fig:grabcut-variant-grabcut}GrabCut]{\centering{}\includegraphics[width=0.14\textwidth,height=0.13\textheight]{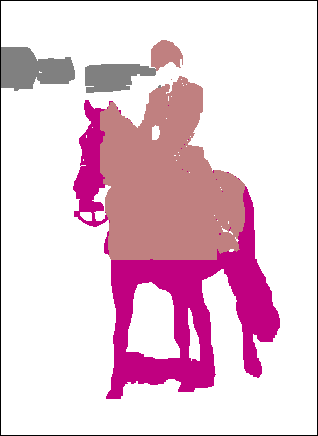}}\hspace*{\fill}
\subfloat[\label{fig:grabcut-variant-grabcut-with-strong-edges}GrabCut+]{\centering{}\includegraphics[width=0.14\textwidth,height=0.13\textheight]{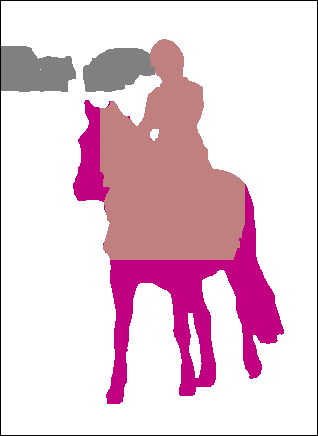}}\hspace*{\fill}
\subfloat[\label{fig:grabcut-variant-grabcut+-perturbed}$\mathtt{GrabCut+^{i}}$]{\centering{}\includegraphics[width=0.14\textwidth,height=0.13\textheight]{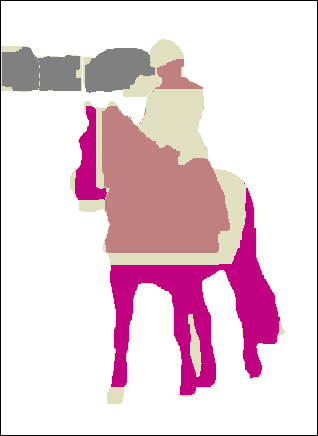}}\hspace*{\fill}
\subfloat[\label{fig:grabcut-variant-MCG}MCG]{\centering{}\includegraphics[width=0.14\textwidth,height=0.13\textheight]{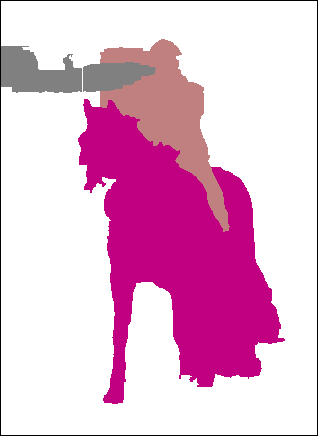}}\hspace*{\fill}
\subfloat[\label{fig:grabcut-variant-M=000026G}$\mbox{M}\,\cap\,\mbox{G+}$]{\centering{}\includegraphics[width=0.14\textwidth,height=0.13\textheight]{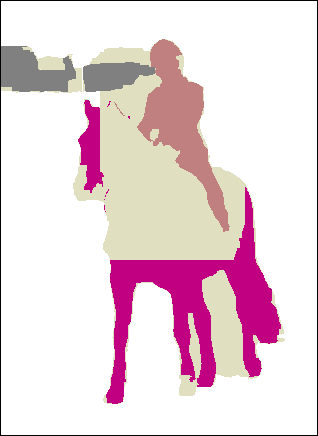}}\hspace*{\fill}\vspace{-0.5em}

\caption{\label{fig:grabcut-variants-examples}Example of the different segmentations
obtained starting from a bounding box annotation. Grey/pink/magenta
indicate different object classes, white is background, and ignore
regions are beige. $\mathtt{M}\cap\mathtt{G+}$ denotes $\mathtt{MCG}\cap\mathtt{GrabCut+}$.}
\vspace{-1em}
\end{figure*}

Because MCG and $\mathtt{GrabCut+}$provide complementary information,
we can think of $\mathtt{M}\cap\mathtt{G+}$ as an improved version
of $\mathtt{GrabCut+^{i}}$ providing a different trade-off between
precision and recall on the generated labels (see Figure \ref{fig:grabcut-variant-M=000026G}). 

The BoxSup method \cite{Dai2015Iccv} also uses MCG object proposals
during training; however, there are important differences. They modify
the training procedure so as to denoise intermediate outputs by randomly
selecting high overlap proposals. In comparison, our approach keeps
the training procedure unmodified and simply generates input labels.
Our approach also uses ignore regions, while BoxSup does not explore
this dimension. Finally, BoxSup uses a longer training than our approach.

Section \ref{sec:Semantic-labelling-results} shows results for the
semantic labelling task, compares different methods and different
supervision regimes. In Section \ref{sec:Instance-segmentation} we
show that the proposed approach is also suitable for the instance
segmentation task. 

\section{\label{sec:Semantic-labelling-results}Semantic labelling results}

Our approach is equally suitable (and effective) for weakly supervised
instance segmentation as well as for semantic labelling. However,
only the latter has directly comparable related work. We thus focus
our experimental comparison efforts on the semantic labelling task.
Results for instance segmentation are presented in Section \ref{sec:Instance-segmentation-results}.

Section \ref{subsec:labelling-experimental-setup} discusses the experimental
setup, evaluation, and implementation details for semantic labelling.
Section \ref{subsec:Main-results} presents our main results, contrasting
the methods from Section \ref{sec:Baselines-and-approach} with the
current state of the art. Section \ref{subsec:Additional-semantic-labelling-results}
further expands these results with a more detailed analysis, and presents
results when using more supervision (semi-supervised case).

\subsection{\label{subsec:labelling-experimental-setup}Experimental setup}

\paragraph{Datasets}

We evaluate the proposed methods on the Pascal VOC12 segmentation
benchmark \cite{Everingham15}. The dataset consists of $20$ foreground
object classes and one background class. The segmentation part of
the VOC12 dataset contains $1\,464$ training, $1\,449$ validation,
and $1\,456$ test images. Following previous work \cite{Chen2015Iclr,Dai2015Iccv},
we extend the training set with the annotations provided by \cite{Hariharan2011Iccv},
resulting in an augmented set of $10\,582$ training images.\\
In some of our experiments, we use additional training images from
the COCO \cite{Lin2014EccvCoco} dataset. We only consider images
that contain any of the $20$ Pascal classes and (following \cite{Zheng2015IccvCrfAsRnn})
only objects with a bounding box area larger than $200$ pixels. After
this filtering, $99\,310$ images remain (from training and validation
sets), which are added to our training set. When using COCO data,
we first pre-train on COCO and then fine-tune over the Pascal VOC12
training set.\\
All of the COCO and Pascal training images come with semantic labelling
annotations (for fully supervised case) and bounding box annotations
(for weakly supervised case).

\paragraph{Evaluation}

We use the ``comp6'' evaluation protocol. The performance is measured
in terms of pixel intersection-over-union averaged across $21$ classes
(mIoU). Most of our results are shown on the validation set, which
we use to guide our design choices.\textcolor{brown}{{} }Final results
are reported on the test set (via the evaluation server) and compared
with other state-of-the-art methods.

\paragraph{Implementation details}

For all our experiments we use the Deep\-Lab\--Large\-FOV network,
using the same train and test parameters as \cite{Chen2015Iclr}.
The model is initialized from a VGG16 network pre-trained on ImageNet
\cite{Simonyan2015Iclr}. We use a mini-batch of $30$ images for
SGD and initial learning rate of $0.001$, which is divided by $10$
after a $2\mbox{k}$/$20\mbox{k}$ iterations (for Pascal/COCO). At
test time, we apply DenseCRF~\cite{Kraehenbuehl2011Nips}. Our network
and post-processing are comparable to the ones used in \cite{Dai2015Iccv,Papandreou2015Iccv}.

Note that multiple strategies have been considered to boost test time
results, such as multi-resolution or model ensembles \cite{Chen2015Iclr,Kokkinos2016Iclr}.
Here we keep the approach simple and fixed. In all our experiments
we use a fixed training and test time procedure. Across experiments
we only change the input training data that the networks gets to see.

\subsection{\label{subsec:Main-results}Main results}

\begin{figure}
\begin{centering}
\includegraphics[width=1\columnwidth]{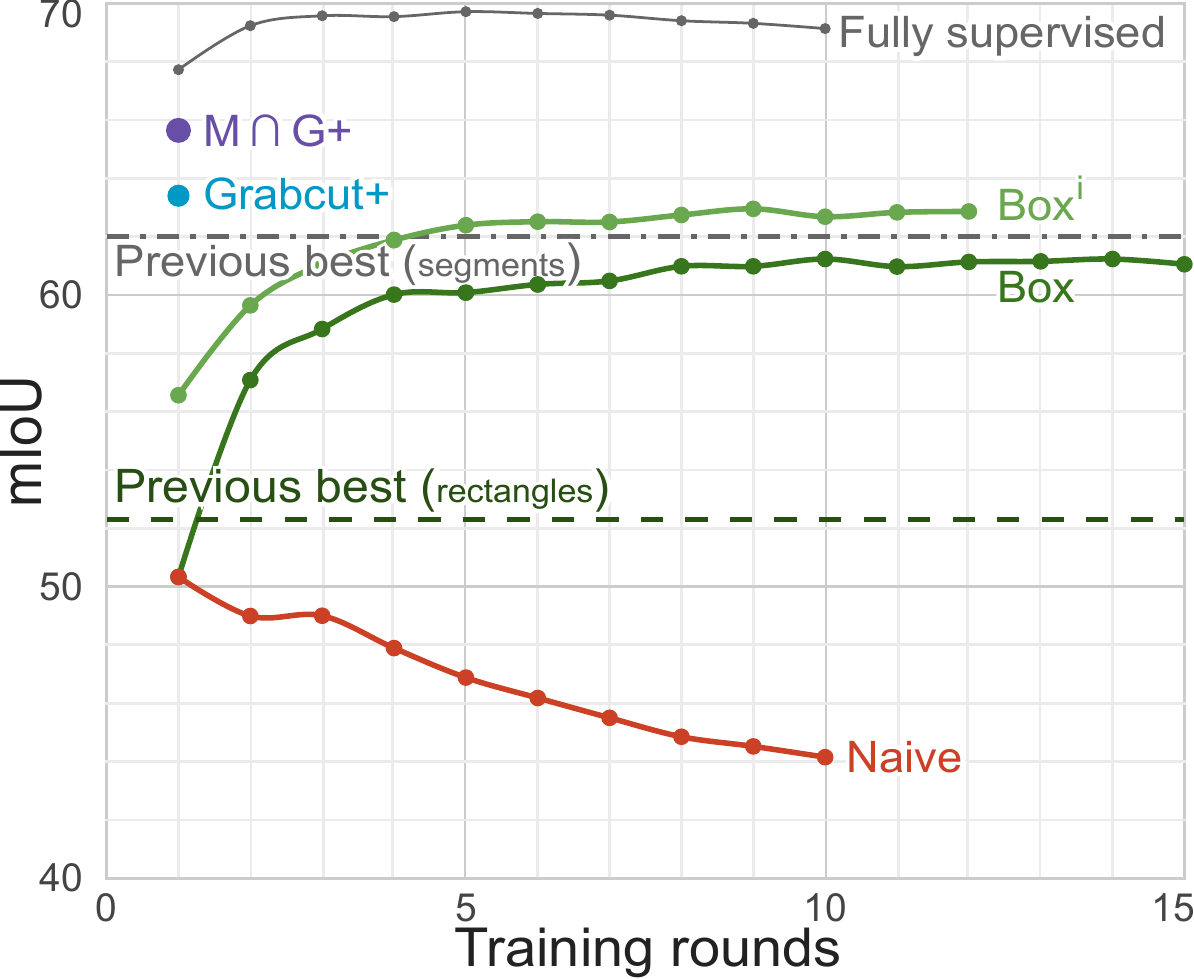}\vspace{-0.5em}
\par\end{centering}
\caption{\label{fig:Segmentation-quality-versus-training-round}Segmentation
quality versus training round for different approaches, see also Tables
\ref{tab:Semantic-labelling-results-boxes-ours} and \ref{tab:Semantic-labelling-results-boxes-others}.
Pascal VOC12 validation set results. ``Previous best (rectangles/segments)''
corresponds to $\mathrm{WSSL_{R}}$/$\mathrm{BoxSup_{MCG}}$ in Table
\ref{tab:Semantic-labelling-results-boxes-others}.}
\vspace{-1.5em}
\end{figure}
\begin{table}
\begin{centering}
\vspace{-0.5em}
\begin{tabular}{cc|c}
\multicolumn{2}{c|}{Method} & val. mIoU\tabularnewline
\hline 
\hline 
\multirow{2}{*}{-} & Fast-RCNN & 44.3\tabularnewline
 & GT Boxes & 62.2\tabularnewline
\hline 
\multirow{6}{*}{%
\begin{tabular}{c}
Weakly\tabularnewline
supervised\tabularnewline
\end{tabular}} & $\mathtt{Box^{\ }}$\textsuperscript{} & 61.2\tabularnewline
 & $\mathtt{Box^{i}}$ & 62.7\tabularnewline
\cline{2-3} 
 & \texttt{$\mathtt{MCG}$} & 62.6\tabularnewline
 & $\mathtt{GrabCut+}^{\ }$ & 63.4\tabularnewline
 & $\mathtt{GrabCut+^{i}}$ & 64.3\tabularnewline
 & $\mathtt{M}\cap\mathtt{G+}$ & \textbf{65.7}\tabularnewline
\hline 
Fully supervised & $\text{DeepLab}_{ours}$\hspace*{0.1em}\cite{Chen2015Iclr} & \uline{69.1}\tabularnewline
\end{tabular}\vspace{-0.5em}
\par\end{centering}
\centering{}\caption{\label{tab:Semantic-labelling-results-boxes-ours}Weakly supervised
semantic labelling results for our baselines. Trained using Pascal
VOC12 bounding boxes alone, validation set results. $\text{DeepLab}_{ours}$
indicates our fully supervised result.}
\vspace{-1.5em}
\end{table}

\paragraph{Box results}

Figure \ref{fig:Segmentation-quality-versus-training-round} presents
the results for the recursive training of the box baselines from Section
\ref{subsec:Box-baselines}. We see that the $\mathtt{Naive}$ scheme,
a recursive training from rectangles disregarding post-processing
stages, leads to poor quality. However, by using the suggested three
post-processing stages, the $\mathtt{Box}$ baseline obtains a significant
gain, getting tantalisingly close to the best reported results on
the task \cite{Dai2015Iccv}. Details of the contribution of each
post-processing stage are presented in the supplementary material.
Adding ignore regions inside the rectangles ($\mathtt{Box}\rightarrow\mathtt{Box^{i}}$)
provides a clear gain and leads by itself to state of the art results.\\
Figure \ref{fig:Segmentation-quality-versus-training-round} also
shows the result of using longer training for fully supervised case.
When using ground truth semantic segmentation annotations, one training
round is enough to achieve good performance; longer training brings
marginal improvement. As discussed in Section \ref{subsec:Box-baselines},
reaching good quality for $\mathtt{Box}/\mathtt{Box^{i}}$ requires
multiple training rounds instead, and performance becomes stable from
round 5 onwards. Instead, $\mathtt{GrabCut+}$/$\mathtt{M}\cap\mathtt{G+}$
do not benefit from additional training rounds.

\paragraph{Box-driven segment results}

Table \ref{tab:Semantic-labelling-results-boxes-ours} evaluates results
on the Pascal VOC12 validation set. It indicates the $\mathtt{Box}/\mathtt{Box^{i}}$
results after $10$ rounds, and \texttt{$\mathtt{MCG}$}/$\mathtt{GrabCut+}$/$\mathtt{GrabCut+^{i}}$/$\mathtt{M}\cap\mathtt{G+}$
results after one round. ``Fast-RCNN'' is the result using detections
\cite{Girshick2015IccvFastRCNN} to generate semantic labels (lower-bound),
``GT Boxes'' considers the box annotations as labels, and $\text{DeepLab}_{ours}$
indicates our fully supervised segmentation network result obtained
with a training length equivalent to three training rounds (upper-bound
for our results). We see in the results that using ignore regions
systematically helps (trading-off recall for precision), and that
$\mathtt{M}\cap\mathtt{G+}$ provides better results than \texttt{$\mathtt{MCG}$}
and $\mathtt{GrabCut+}$ alone.\\
Table \ref{tab:Semantic-labelling-results-boxes-others} indicates
the box-driven segment results after $1$ training round and shows
comparison with other state of the art methods, trained from boxes
only using either Pascal VOC12, or VOC12+COCO data. $\mathrm{BoxSup_{R}}$
and $\mathrm{WSSL_{R}}$ both feed the network with rectangle segments
(comparable to $\mathtt{Box^{i}}$), while $\mathrm{WSSL_{S}}$ and
$\mathrm{BoxSup_{MCG}}$ exploit arbitrary shaped segments (comparable
to $\mathtt{M}\cap\mathtt{G+}$). Although our network and post-processing
is comparable to the ones in \cite{Dai2015Iccv,Papandreou2015Iccv},
there are differences in the exact training procedure and parameters
(details in supplementary material).\\
Overall, our results indicate that \emph{- }without modifying the
training procedure - $\mathtt{M}\cap\mathtt{G+}$ is able to improve
over previously reported results and reach $95\%$ of the fully-supervised
training quality. By training with COCO data \cite{Lin2014EccvCoco}
before fine-tuning for Pascal VOC12, we see that with enough additional
bounding boxes we can match the full supervision from Pascal VOC 12
($68.9$ versus $69.1$). This shows that the labelling effort could
be significantly reduced by replacing segmentation masks with bounding
box annotations.

\begin{table}
\begin{centering}
\begingroup
\setlength{\tabcolsep}{0.15em} 
\hspace*{-1em}%
\begin{tabular}{ccc|c|c|cc}
\begin{tabular}{c}
{\small{}Super-}\tabularnewline
{\small{}vision}\tabularnewline
\end{tabular} & %
\begin{tabular}{c}
{\small{}\#GT}\tabularnewline
{\small{}images}\tabularnewline
\end{tabular}  & %
\begin{tabular}{c}
{\small{}\#Weak}\tabularnewline
{\small{}images}\tabularnewline
\end{tabular}  & Method & %
\begin{tabular}{c}
{\small{}val. set}\tabularnewline
mIoU\tabularnewline
\end{tabular} & \multicolumn{2}{c}{%
\begin{tabular}{cc}
\multicolumn{2}{c}{{\small{}test set}}\tabularnewline
mIoU & \,$\text{FS}\%$\tabularnewline
\end{tabular}}\tabularnewline
\hline 
\hline 
\multicolumn{7}{c}{\cellcolor{verylightgray}VOC12 (V)}\tabularnewline
\multirow{7}{*}{Weak} & \multirow{7}{*}{-} & \multirow{7}{*}{V\hspace*{0.1em}10k} & Bearman et al. \cite{Bearman2015ArXiv} & 45.1 & - & -\tabularnewline
 &  &  & $\mathrm{BoxSup_{R}}$ \cite{Dai2015Iccv} & 52.3 & - & -\tabularnewline
 &  &  & $\mathrm{WSSL_{R}}$\cite{Papandreou2015Iccv} & 52.5 & 54.2 & 76.9\tabularnewline
 &  &  & $\mathrm{WSSL_{S}}$\cite{Papandreou2015Iccv} & 60.6 & 62.2 & 88.2\tabularnewline
 &  &  & $\mathrm{BoxSup_{MCG}}$\cite{Dai2015Iccv} & 62.0 & 64.6 & 91.6\tabularnewline
 &  &  & $\mathtt{Box^{i}}$\textsuperscript{} & 62.7 & 63.5 & 90.0\tabularnewline
 &  &  & $\mathtt{M}\cap\mathtt{G+}$ & \textbf{65.7} & \textbf{67.5} & \textbf{95.7}\tabularnewline
\hline 
\multirow{4}{*}{Semi} & \multirow{4}{*}{V\hspace*{0.1em}1.4k} & \multirow{4}{*}{V\hspace*{0.1em}9k} & $\mathrm{WSSL_{R}}$\cite{Papandreou2015Iccv} & 62.1 & - & -\tabularnewline
 &  &  & $\mathrm{BoxSup_{MCG}}$\cite{Dai2015Iccv} & 63.5 & 66.2 & 93.9\tabularnewline
 &  &  & $\mathrm{WSSL_{S}}$\cite{Papandreou2015Iccv} & 65.1 & 66.6 & 94.5\tabularnewline
 &  &  & $\mathtt{M}\cap\mathtt{G+}$ & \textbf{65.8} & \textbf{66.9} & \textbf{94.9}\tabularnewline
\hline 
\multirow{3}{*}{Full} & \multirow{3}{*}{V\hspace*{0.1em}10k} & \multirow{3}{*}{-} & BoxSup \cite{Dai2015Iccv} & 63.8 & - & -\tabularnewline
 &  &  & WSSL \cite{Papandreou2015Iccv} & 67.6 & 70.3 & 99.7\tabularnewline
 &  &  & $\text{DeepLab}_{ours}$\hspace*{0.1em}\cite{Chen2015Iclr} & \uline{69.1} & \uline{70.5} & 100\tabularnewline
\hline 
\hline 
\multicolumn{7}{c}{\cellcolor{verylightgray}VOC12 + COCO (V+C)}\tabularnewline
\multirow{2}{*}{Weak} & \multirow{2}{*}{-} & \multirow{2}{*}{%
\begin{tabular}{c}
V+C\tabularnewline
110k\tabularnewline
\end{tabular}} & $\mathtt{Box^{i}}$\textsuperscript{} & \multicolumn{1}{c}{65.3} & 66.7 & 91.1\tabularnewline
 &  &  & $\mathtt{M}\cap\mathtt{G+}$ & \multicolumn{1}{c}{\textbf{68.9}} & \textbf{69.9} & \textbf{95.5}\tabularnewline
\hline 
\multirow{2}{*}{Semi} & \multirow{2}{*}{V\hspace*{0.1em}10k} & C\hspace*{0.1em}123k & $\mathrm{BoxSup_{MCG}}$\cite{Dai2015Iccv} & \multicolumn{1}{c}{68.2} & 71.0 & 97.0\tabularnewline
 &  & C\hspace*{0.1em}100k & $\mathtt{M}\cap\mathtt{G+}$ & \multicolumn{1}{c}{\textbf{71.6}} & \textbf{72.8} & \textbf{99.5}\tabularnewline
\hline 
\multirow{3}{*}{Full} & \multirow{2}{*}{\hspace*{-1em}{\small{}V+C\hspace*{0.1em}133k}\hspace*{-1em} } & \multirow{3}{*}{-} & BoxSup \cite{Dai2015Iccv} & \multicolumn{1}{c}{68.1} & - & -\tabularnewline
 &  &  & WSSL \cite{Papandreou2015Iccv} & \multicolumn{1}{c}{71.7} & 73 & 99.7\tabularnewline
 & \hspace*{-1em}{\small{}V+C\hspace*{0.1em}110k}\hspace*{-1em} &  & $\text{DeepLab}_{ours}$\hspace*{0.1em}\cite{Chen2015Iclr} & \multicolumn{1}{c}{\uline{72.3}} & \uline{73.2} & 100\tabularnewline
\end{tabular}\vspace{-0.5em}
\endgroup
\par\end{centering}
\caption{\label{tab:Semantic-labelling-results-boxes-others}\label{tab:Semantic-labelling-semi-results}Semantic
labelling results for validation and test set; under different training
regimes with VOC12 (V) and COCO data (C). Underline indicates full
supervision baselines, and bold are our best weakly- and semi-supervised
results. $\text{FS}\%$: performance relative to the best fully supervised
model ($\text{DeepLab}_{ours}$). Discussion in Sections \ref{subsec:Main-results}
and \ref{subsec:Additional-semantic-labelling-results}.}
\vspace{-1.5em}
\end{table}
\begin{figure*}[t]
\begingroup{
\setlength{\tabcolsep}{0pt} 
\renewcommand{\arraystretch}{0.2}

\begin{tabular}[b]{ccccccccccccc}
\includegraphics[width=0.14\textwidth]{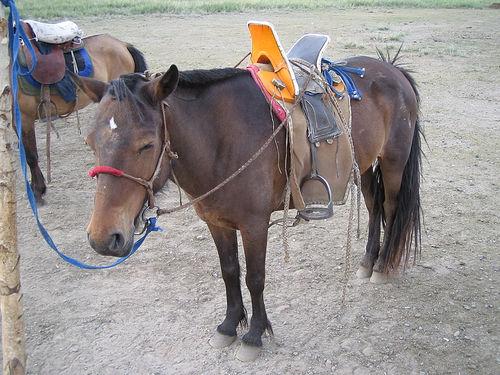} & \hspace*{0.1em} & \includegraphics[width=0.14\textwidth]{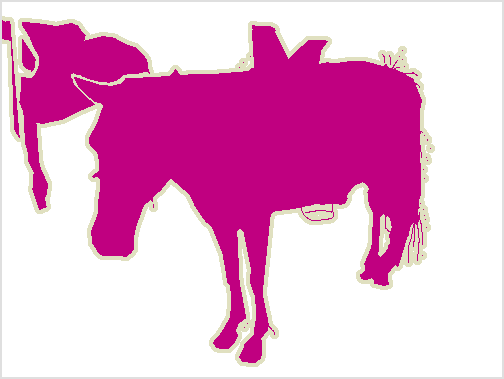} & \hspace*{0.1em} & \includegraphics[width=0.14\textwidth]{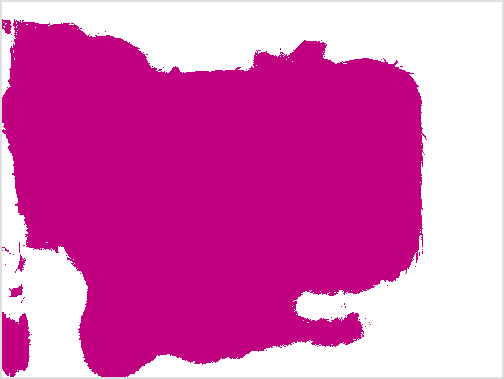} & \hspace*{0.1em} & \includegraphics[width=0.14\textwidth]{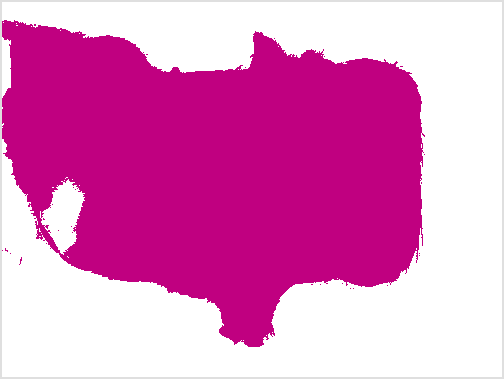} & \hspace*{0.1em} & \includegraphics[width=0.14\textwidth]{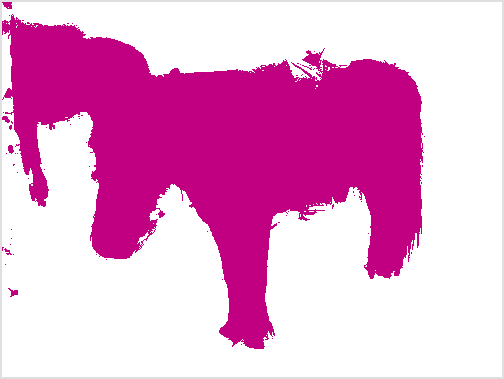} & \hspace*{0.1em} & \includegraphics[width=0.14\textwidth]{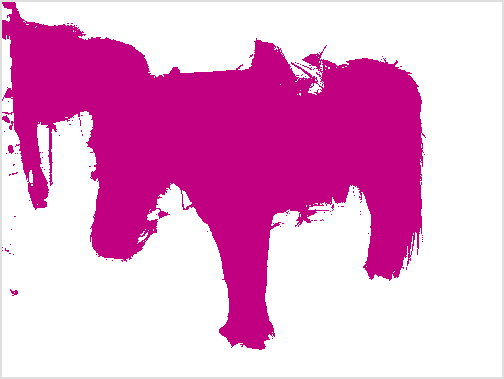} & \hspace*{0.1em} & \includegraphics[width=0.14\textwidth]{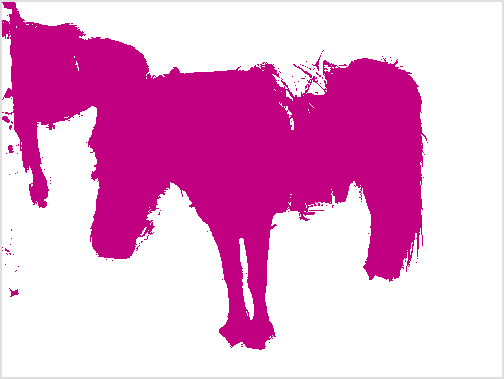}\tabularnewline
\includegraphics[width=0.14\textwidth]{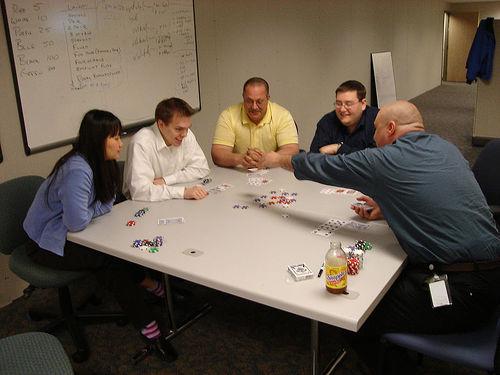} & \hspace*{0.1em} & \includegraphics[width=0.14\textwidth]{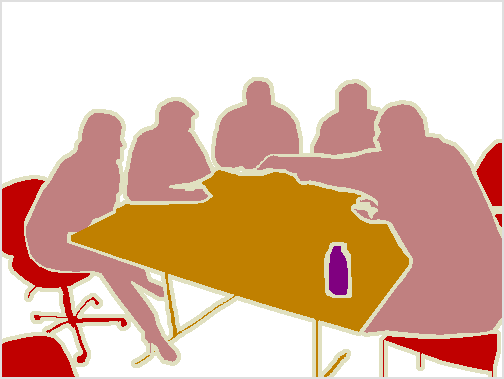} & \hspace*{0.1em} & \includegraphics[width=0.14\textwidth]{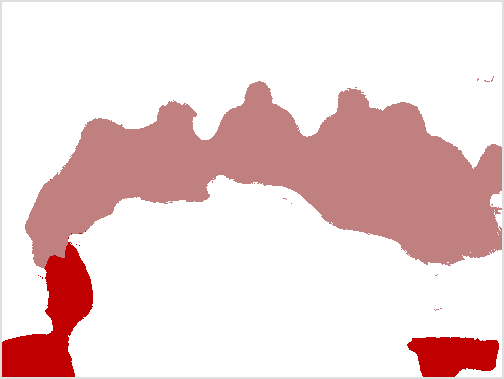} & \hspace*{0.1em} & \includegraphics[width=0.14\textwidth]{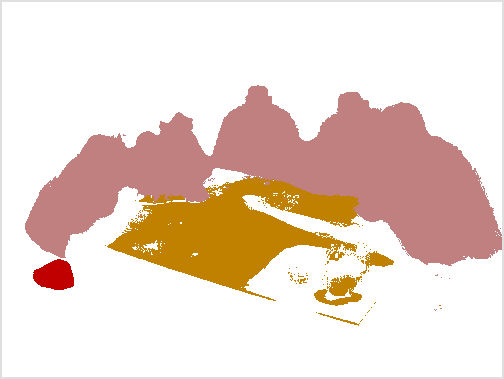} & \hspace*{0.1em} & \includegraphics[width=0.14\textwidth]{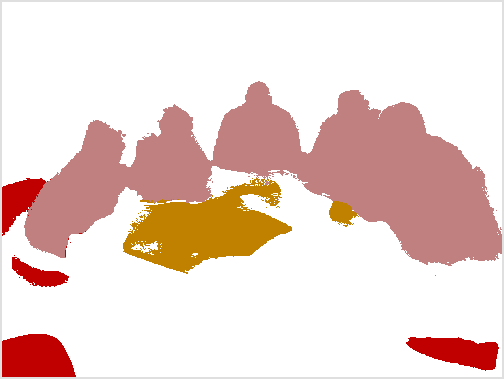} & \hspace*{0.1em} & \includegraphics[width=0.14\textwidth]{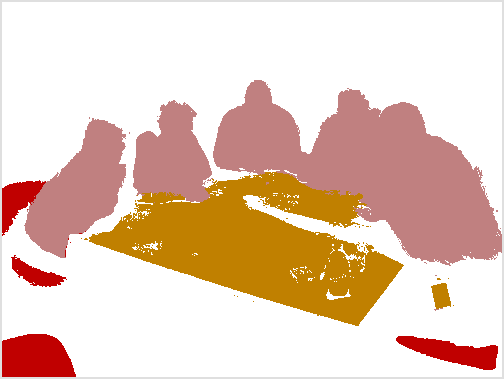} & \hspace*{0.1em} & \includegraphics[width=0.14\textwidth]{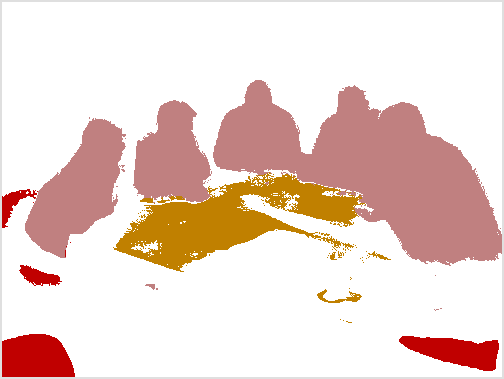}\tabularnewline
\includegraphics[width=0.14\textwidth]{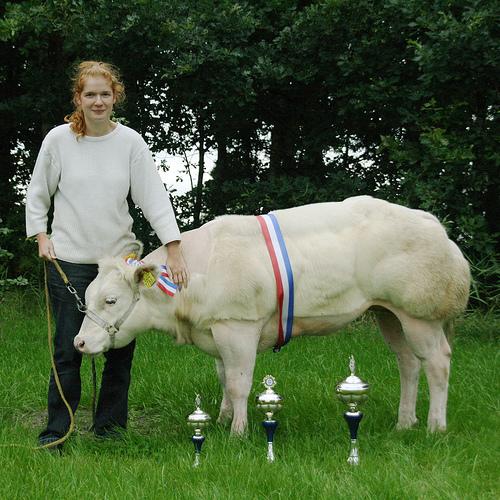} & \hspace*{0.1em} & \includegraphics[width=0.14\textwidth]{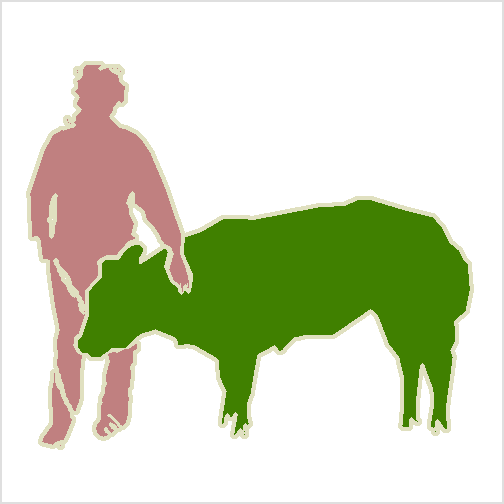} & \hspace*{0.1em} & \includegraphics[width=0.14\textwidth]{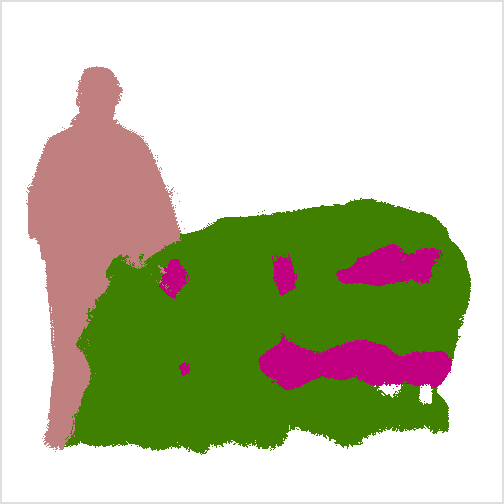} & \hspace*{0.1em} & \includegraphics[width=0.14\textwidth]{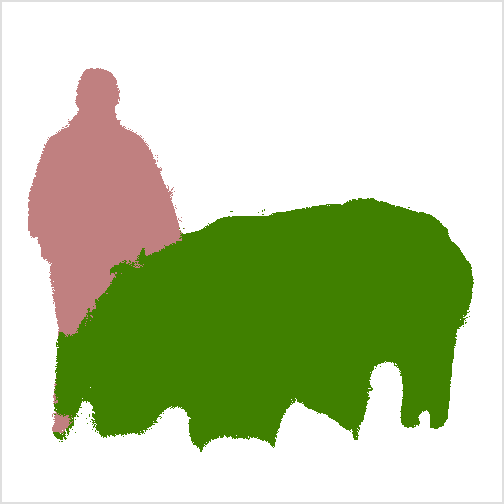} & \hspace*{0.1em} & \includegraphics[width=0.14\textwidth]{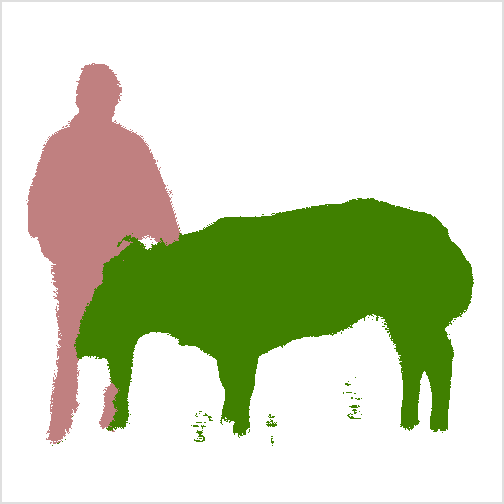} & \hspace*{0.1em} & \includegraphics[width=0.14\textwidth]{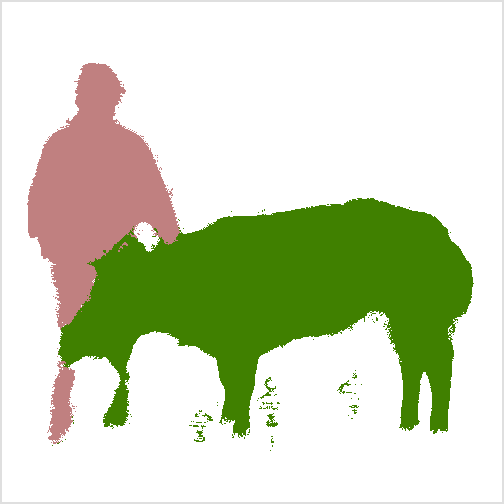} & \hspace*{0.1em} & \includegraphics[width=0.14\textwidth]{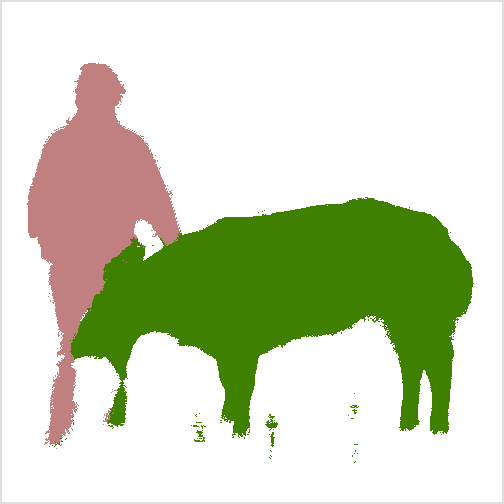}\tabularnewline
\includegraphics[width=0.14\textwidth]{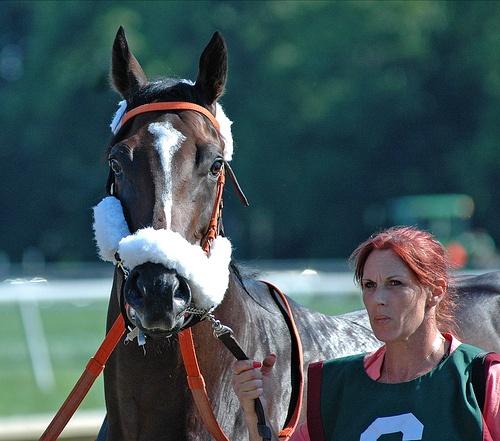} & \hspace*{0.1em} & \includegraphics[width=0.14\textwidth]{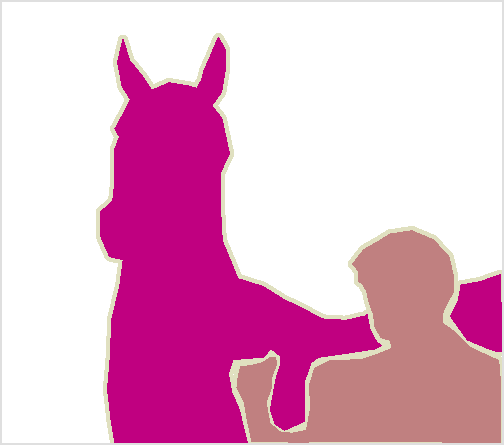} & \hspace*{0.1em} & \includegraphics[width=0.14\textwidth]{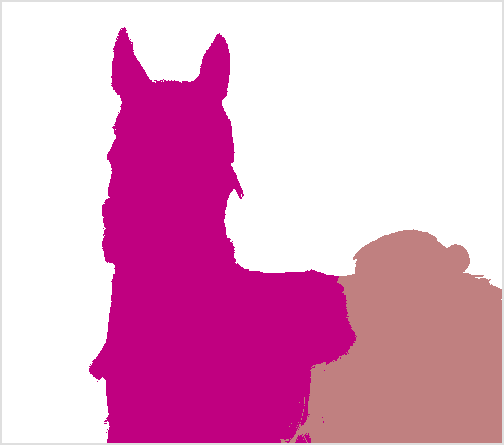} & \hspace*{0.1em} & \includegraphics[width=0.14\textwidth]{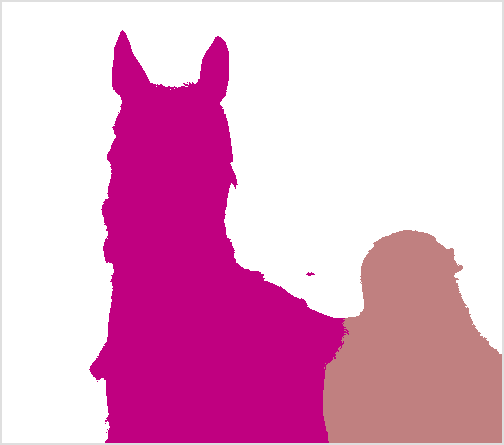} & \hspace*{0.1em} & \includegraphics[width=0.14\textwidth]{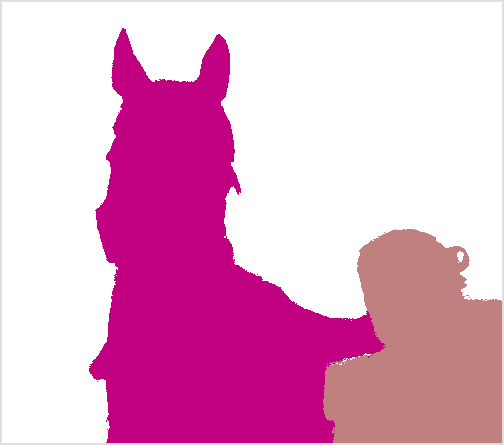} & \hspace*{0.1em} & \includegraphics[width=0.14\textwidth]{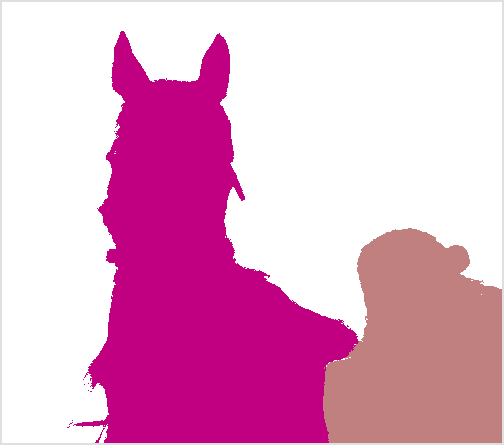} & \hspace*{0.1em} & \includegraphics[width=0.14\textwidth]{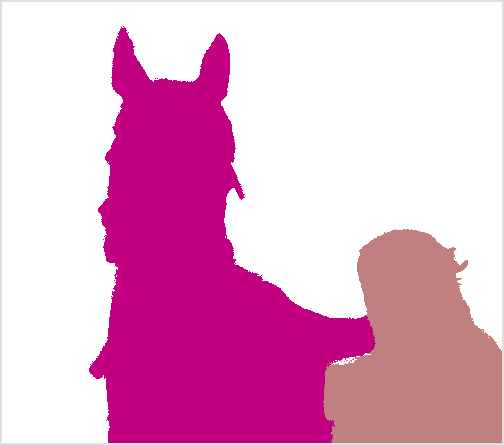}\tabularnewline
\vspace{-0.25em}
 &  &  &  &  &  &  &  &  &  &  &  & \tabularnewline
Image &  & %
\begin{tabular}{c}
Ground\tabularnewline
truth\tabularnewline
\end{tabular}  &  & $\mathtt{Box}$ &  & $\mathtt{Box^{i}}$ &  & $\mathtt{M}\cap\mathtt{G+}$ &  & %
\begin{tabular}{c}
Semi\tabularnewline
supervised\tabularnewline
$\mathtt{M}\cap\mathtt{G+}$\tabularnewline
\end{tabular} &  & %
\begin{tabular}{c}
Fully\tabularnewline
supervised\tabularnewline
\end{tabular}\tabularnewline
\end{tabular}\vspace{-0.5em}

\caption{\label{fig:Qualitative-results}Qualitative results on VOC12. Visually,
the results from our weakly supervised method $\mathtt{M}\cap\mathtt{G+}$
are hardly distinguishable from the fully supervised ones.}
\vspace{-1em}
}\endgroup
\end{figure*}

\subsection{\label{subsec:Additional-semantic-labelling-results}Additional results}

\paragraph{Semi-supervised case}

Table \ref{tab:Semantic-labelling-semi-results} compares results
in the semi-supervised modes considered by \cite{Dai2015Iccv,Papandreou2015Iccv},
where some of the images have full supervision, and some have only
bounding box supervision. Training with $10\%$ of Pascal VOC12 semantic
labelling annotations does not bring much gain to the performance
($65.7$ versus $65.8$), this hints at the high quality of the generated
$\mathtt{M}\cap\mathtt{G+}$ input data. 

By using ground-truth annotations on Pascal plus bounding box annotations
on COCO, we observe $2.5$ points gain ($69.1\negmedspace\rightarrow\negmedspace71.6$
, see Table \ref{tab:Semantic-labelling-semi-results}). This suggests
that the overall performance could be further improved by using extra
training data with bounding box annotations.

\paragraph{Boundaries supervision}

Our results from $\mathtt{MCG}$, $\mathtt{GrabCut+}$, and $\mathtt{M}\cap\mathtt{G+}$
all indirectly include information from the BSDS500 dataset \cite{ArbelaezMaireFowlkesMalikPAMI11}
via the HED boundary detector \cite{Xie2015Iccv}. These results are
fully comparable to BoxSup-MCG \cite{Dai2015Iccv}, to which we see
a clear improvement. Nonetheless one would like to know how much using
dense boundary annotations from BSDS500 contributes to the results.
We use the weakly supervised boundary detection technique from \cite{Khoreva2016Cvpr}
to learn boundaries directly from the Pascal VOC12 box annotations.
Training $\mathtt{M}\cap\mathtt{G+}$ using weakly supervised HED
boundaries results in $1$ point loss compared to using the BSDS500
($64.8$ versus $65.7$ mIoU on Pascal VOC12 validation set). We see
then that although the additional supervision does bring some help,
it has a minor effect and our results are still rank at the top even
when we use only Pascal VOC12 + ImageNet pre-training.

\paragraph{Different convnet results}

For comparison purposes with \cite{Dai2015Iccv,Papandreou2015Iccv}
we used Deep\-Lab\-v1 with a VGG-16 network in our experiments.
To show that our approach also generalizes across different convnets,
we also trained Deep\-Lab\-v2 with a ResNet101 network \cite{Chen2016ArxivDeeplabv2}.
Table \ref{tab:Semantic-labelling-results-ResNet} presents the results.\\
Similar to the case with VGG-16, our weakly supervised approach $\mathtt{M}\cap\mathtt{G+}$
reaches $93\%\text{/}95\%$ of the fully supervised case when training
with VOC12/VOC12+COCO, and the weakly supervised results with COCO
data reach similar quality to full supervision with VOC12 only.

\begin{table}
\begin{centering}
\begin{tabular}{c|c|cc}
Supervision & Method & mIoU & $\text{FS}\%$\tabularnewline
\hline 
\hline 
\multicolumn{4}{c}{\cellcolor{verylightgray}VOC12}\tabularnewline
\multirow{1}{*}{Weak} & $\mathtt{M}\cap\mathtt{G+}$ & 69.4 & 93.2\tabularnewline
Full & DeepLabv2-ResNet101 \cite{Chen2016ArxivDeeplabv2} & \uline{74.5} & 100\tabularnewline
\hline 
\hline 
\multicolumn{4}{c}{\cellcolor{verylightgray}VOC12 + COCO}\tabularnewline
\multirow{1}{*}{Weak} & $\mathtt{M}\cap\mathtt{G+}$ & 74.2 & 95.5\tabularnewline
\multirow{1}{*}{Full} & DeepLabv2-ResNet101 \cite{Chen2016ArxivDeeplabv2} & \uline{77.7} & 100\tabularnewline
\end{tabular}\vspace{-0.5em}
\par\end{centering}
\caption{\label{tab:Semantic-labelling-results-ResNet}Deep\-Lab\-v2-ResNet101
network semantic labelling results on VOC12 validation set, using
VOC12 or VOC12+COCO training data. $\text{FS}\%$: performance relative
to the full supervision. Discussion in Section \ref{subsec:Additional-semantic-labelling-results}.}

\vspace{-1em}
\end{table}

\section{\label{sec:Instance-segmentation}From boxes to instance segmentation}

Complementing the experiments of the previous sections, we also explore
a second task: weakly supervised instance segmentation. To the best
of our knowledge, these are the first reported experiments on this
task. 

As object detection moves forward, there is a need to provide richer
output than a simple bounding box around objects. Recently \cite{Hariharan2015Cvpr,Pinheiro2015Nips,Pinheiro2016Eccv}
explored training convnets to output a foreground versus background
segmentation of an instance inside a given bounding box. Such networks
are trained using pixel-wise annotations that distinguish between
instances. These annotations are more detailed and expensive than
semantic labelling, and thus there is interest in weakly supervised
training.

The segments used for training, as discussed in Section \ref{subsec:GrabCut-baselines},
are generated starting from individual object bounding boxes. Each
segment represents a different object instance and thus can be used
directly to train an instance segmentation convnet. For each annotated
bounding box, we generate a foreground versus background segmentation
using the $\mathtt{GrabCut+}$ method (Section \ref{subsec:GrabCut-baselines}),
and train a convnet to regress from the image and bounding box information
to the instance segment.

\section{\label{sec:Instance-segmentation-results}Instance segmentation results}

\paragraph{\label{sec:Instance-segmentation-setup}Experimental setup}

We choose a purposely simple instance segmentation pipeline, based
on the ``hyper-columns system 2'' architecture \cite{Hariharan2015Cvpr}.
We use Fast-RCNN \cite{Girshick2015IccvFastRCNN} detections (post-NMS)
with their class score, and for each detection estimate an associated
foreground segment. We estimate the foreground using either some baseline
method (e.g. GrabCut) or using convnets trained for the task \cite{Pinheiro2015Nips,Chen2016ArxivDeeplabv2}. 

For our experiments we use a re-implementation of the $\mathrm{DeepMask}$
\cite{Pinheiro2015Nips} architecture, and additionally we re-purpose
a Deep\-Lab\-v2 VGG-16 network \cite{Chen2016ArxivDeeplabv2} for
the instance segmentation task, which we name $\mathrm{DeepLab_{BOX}}$. 

Inspired by \cite{Xu2016Cvpr,Carreira2016Cvpr}, we modify Deep\-Lab
to accept four input channels: the input image RGB channels, plus
a binary map with a bounding box of the object instance to segment.
We train the network $\mathrm{DeepLab_{BOX}}$ to output the segmentation
mask of the object corresponding to the input bounding box. The additional
input channel guides the network so as to segment only the instance
of interest instead of all objects in the scene. The input box rectangle
can also be seen as an initial guess of the desired output. We train
using ground truth bounding boxes, and at test time Fast-RCNN detection
boxes are used. 

We train $\mathrm{DeepMask}$ and $\mathrm{DeepLab_{BOX}}$ using
$\mathtt{GrabCut+}$ results either over Pascal VOC12 or VOC12+COCO
data (1 training round, no recursion like in Section \ref{subsec:Box-baselines}),
and test on the VOC12 validation set, the same set of images used
in Section \ref{sec:Semantic-labelling-results}. The augmented annotation
from \cite{Hariharan2011Iccv} provides per-instance segments for
VOC12. We do not use CRF post-processing for neither of the networks.\\
Following instance segmentation literature \cite{Hariharan2014Eccv,Hariharan2015Cvpr}
we report in Table \ref{tab:Instance-segmentation-results} $\text{mAP}^{r}$
at IoU threshold $0.5$ and $0.75$. $\text{mAP}^{r}$ is similar
to the traditional VOC12 evaluation, but using IoU between segments
instead of between boxes. Since we have a fixed set of windows, we
can also report the average best overlap (ABO) \cite{PontTuset2015Iccv}
metric to give a different perspective on the results.
\begin{table}
\begin{centering}
\begin{tabular}{cc|cc|c}
Supervision & Method & $\text{mAP}_{0.5}^{r}$\hspace*{-1em} & $\text{mAP}_{0.75}^{r}$ & ABO\tabularnewline
\hline 
\hline 
\multirow{5}{*}{- } & Rectangle\textsuperscript{} & 21.6 & 1.8 & 38.5\tabularnewline
 & Ellipse & 29.5 & 3.9 & 41.7\tabularnewline
 & MCG & 28.3 & 5.9 & 44.7\tabularnewline
 & GrabCut & 38.5 & 13.9 & 45.8\tabularnewline
 & GrabCut+ & 41.1 & 17.8 & 46.4\tabularnewline
\hline 
\hline 
\multicolumn{5}{c}{\cellcolor{verylightgray}VOC12}\tabularnewline
\multirow{2}{*}{Weak} & $\mathrm{DeepMask}$ & 39.4 & 8.1 & 45.8\tabularnewline
 & $\mathrm{DeepLab_{BOX}}$ & 44.8 & 16.3 & \textbf{49.1}\tabularnewline
\hline 
\multirow{2}{*}{Full} & $\mathrm{DeepMask}$ & 41.7 & 9.7 & 47.1\tabularnewline
 & $\mathrm{DeepLab_{BOX}}$ & 47.5 & 20.2 & \uline{51.1}\tabularnewline
\hline 
\hline 
\multicolumn{5}{c}{\cellcolor{verylightgray}VOC12 + COCO}\tabularnewline
\multirow{2}{*}{Weak} & $\mathrm{DeepMask}$ & 42.9 & 11.5 & 48.8\tabularnewline
 & $\mathrm{DeepLab_{BOX}}$ & 46.4 & 18.5 & \textbf{51.4}\tabularnewline
\hline 
\multirow{2}{*}{Full} & $\mathrm{DeepMask}$ & 44.7 & 13.1 & 49.7\tabularnewline
 & $\mathrm{DeepLab_{BOX}}$ & 49.4 & 23.7 & \uline{53.1}\tabularnewline
\end{tabular}\vspace{-0.5em}
\caption{\label{tab:Instance-segmentation-results}Instance segmentation results
on VOC12 validation set. Underline indicates the full supervision
baseline, and bold are our best weak supervision results. Weakly supervised
$\mathrm{DeepMask}$ and $\mathrm{DeepLab_{BOX}}$ reach comparable
results to full supervision. See Section \ref{sec:Instance-segmentation-setup}
for details. }
\par\end{centering}
\vspace{-1em}
\end{table}
\begin{figure}
\begin{centering}
\begin{tabular}{c}
\includegraphics[width=0.75\columnwidth]{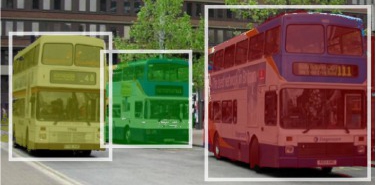}\tabularnewline
\includegraphics[width=0.75\columnwidth]{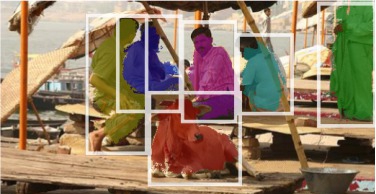}\tabularnewline
\end{tabular}\vspace{-0.5em}
\par\end{centering}
\caption{\label{fig:Instance-segmentation-example}Example result from our
weakly supervised $\mathrm{DeepMask}$ (VOC12+COCO) model.}
\vspace{-1em}
\end{figure}

\paragraph{Baselines}

We consider five training-free baselines: simply filling in the detection
rectangles (boxes) with foreground labels, fitting an ellipse inside
the box, using the MCG proposal with best bounding box IoU, and using
GrabCut and $\mathtt{GrabCut+}$ (see Section \ref{subsec:GrabCut-baselines}),
initialized from the detection box. 

\paragraph{Analysis}

The results table \ref{tab:Instance-segmentation-results} follows
the same trend as the semantic labelling results in Section \ref{sec:Semantic-labelling-results}.
$\mathtt{GrabCut+}$ provides the best results among the baselines
considered and shows comparable performance to $\mathrm{DeepMask}$,
while our proposed $\mathrm{DeepLab_{BOX}}$ outperforms both techniques.
We see that our weakly supervised approach reaches $\sim\negmedspace95\%$
of the quality of fully-supervised case (both on $\text{mAP}_{0.5}^{r}$
and ABO metrics) using two different convnets, $\mathrm{DeepMask}$
and $\mathrm{DeepLab_{BOX}}$, both when training with VOC12 or VOC12+COCO. 

Examples of the instance segmentation results from weakly supervised
$\mathrm{DeepMask}$ (VOC12+COCO) are shown in Figure \ref{fig:Instance-segmentation-example}.
Additional example results are presented in the supplementary material.

\section{\label{sec:Conclusion}Conclusion}

The series of experiments presented in this paper provides new insights
on how to train pixel-labelling convnets from bounding box annotations
only. We showed that when carefully employing the available cues,
recursive training using only rectangles as input can be surprisingly
effective ($\mathtt{Box^{i}}$). Even more, when using box-driven
segmentation techniques and doing a good balance between accuracy
and recall in the noisy training segments, we can reach state of the
art performance without modifying the segmentation network training
procedure ($\mathtt{M}\cap\mathtt{G+}$). Our results improve over
previously reported ones on the semantic labelling task and reach
$\sim\!95\%$ of the quality of the same network trained on the ground
truth segmentation annotations (over the same data). By employing
extra training data with bounding box annotations from COCO we are
able to match the full supervision results. We also report the first
results for weakly supervised instance segmentation, where we also
reach $\sim\!95\%$ of the quality of the fully-supervised training.

Our current approach exploits existing box-driven segmentation techniques,
treating each annotated box individually. In future work we would
like to explore co-segmentation ideas (treating the set of annotations
as a whole), and consider even weaker forms of supervision.

\bibliographystyle{ieee}
\bibliography{2017_cvpr_box_supervised_segmentation}

\begin{thebibliography}{10}\itemsep=-1pt

\bibitem{ArbelaezMaireFowlkesMalikPAMI11}
P.~Arbel\'{a}ez, M.~Maire, C.~Fowlkes, and J.~Malik.
\newblock Contour detection and hierarchical image segmentation.
\newblock {\em PAMI}, 2011.

\bibitem{Barron2015ArXiv}
J.~Barron and B.~Poole.
\newblock The fast bilateral solver.
\newblock {\em arXiv preprint arXiv:1511.03296}, 2015.

\bibitem{Bearman2015ArXiv}
A.~Bearman, O.~Russakovsky, V.~Ferrari, and L.~Fei-Fei.
\newblock What's the point: Semantic segmentation with point supervision.
\newblock {\em arXiv preprint arXiv:1506.02106}, 2015.

\bibitem{Carreira2016Cvpr}
J.~Carreira, P.~Agrawal, K.~Fragkiadaki, and J.~Malik.
\newblock Human pose estimation with iterative error feedback.
\newblock In {\em CVPR}, 2016.

\bibitem{Chen2015Iclr}
L.~Chen, G.~Papandreou, I.~Kokkinos, K.~Murphy, and A.~Yuille.
\newblock Semantic image segmentation with deep convolutional nets and fully
  connected crfs.
\newblock In {\em ICLR}, 2015.

\bibitem{Chen2016ArxivDeeplabv2}
L.-C. Chen, G.~Papandreou, I.~Kokkinos, K.~Murphy, and A.~L. Yuille.
\newblock Deeplab: Semantic image segmentation with deep convolutional nets,
  atrous convolution, and fully connected crfs.
\newblock {\em arXiv:1606.00915}, 2016.

\bibitem{Cheng2015CgfDenseCut}
M.~Cheng, V.~Prisacariu, S.~Zheng, P.~Torr, and C.~Rother.
\newblock Densecut: Densely connected crfs for realtime grabcut.
\newblock {\em Computer Graphics Forum}, 2015.

\bibitem{Dai2015Iccv}
J.~Dai, K.~He, and J.~Sun.
\newblock Boxsup: Exploiting bounding boxes to supervise convolutional networks
  for semantic segmentation.
\newblock In {\em ICCV}, 2015.

\bibitem{Everingham15}
M.~Everingham, S.~M.~A. Eslami, L.~Van~Gool, C.~K.~I. Williams, J.~Winn, and
  A.~Zisserman.
\newblock The pascal visual object classes challenge: A retrospective.
\newblock {\em IJCV}, 2015.

\bibitem{Girshick2015IccvFastRCNN}
R.~Girshick.
\newblock Fast {R-CNN}.
\newblock In {\em ICCV}, 2015.

\bibitem{Gould2009Iccv}
S.~Gould, R.~Fulton, and D.~Koller.
\newblock Decomposing a scene into geometric and semantically consistent
  regions.
\newblock In {\em ICCV}, 2009.

\bibitem{Hariharan2011Iccv}
B.~Hariharan, P.~Arbel\'{a}ez, L.~Bourdev, S.~Maji, and J.~Malik.
\newblock Semantic contours from inverse detectors.
\newblock In {\em ICCV}, 2011.

\bibitem{Hariharan2014Eccv}
B.~Hariharan, P.~Arbel\'{a}ez, R.~Girshick, and J.~Malik.
\newblock Simultaneous detection and segmentation.
\newblock In {\em ECCV}, 2014.

\bibitem{Hariharan2015Cvpr}
B.~Hariharan, P.~Arbel{\'a}ez, R.~Girshick, and J.~Malik.
\newblock Hypercolumns for object segmentation and fine-grained localization.
\newblock In {\em CVPR}, 2015.

\bibitem{Hong2015Nips}
S.~Hong, H.~Noh, and B.~Han.
\newblock Decoupled deep neural network for semi-supervised semantic
  segmentation.
\newblock In {\em NIPS}, 2015.

\bibitem{Hosang2015Pami}
J.~Hosang, R.~Benenson, P.~Doll\'{a}r, and B.~Schiele.
\newblock What makes for effective detection proposals?
\newblock {\em PAMI}, 2015.

\bibitem{Khoreva2016Cvpr}
A.~Khoreva, R.~Benenson, M.~Omran, M.~Hein, and B.~Schiele.
\newblock Weakly supervised object boundaries.
\newblock In {\em CVPR}, 2016.

\bibitem{Kokkinos2016Iclr}
I.~Kokkinos.
\newblock Pushing the boundaries of boundary detection using deep learning.
\newblock In {\em ICLR}, 2016.

\bibitem{Kolmogorov2004Pami}
V.~Kolmogorov and R.~Zabih.
\newblock What energy functions can be minimized via graph cuts?.
\newblock {\em PAMI}, 2004.

\bibitem{Kraehenbuehl2011Nips}
P.~Kr\"ahenb\"uhl and V.~Koltun.
\newblock Efficient inference in fully connected crfs with gaussian edge
  potentials.
\newblock In {\em NIPS}. 2011.

\bibitem{Krahenbuhl2015Cvpr}
P.~Kr\"ahenb\"uhl and V.~Koltun.
\newblock Learning to propose objects.
\newblock In {\em CVPR}, 2015.

\bibitem{Lempitsky2009Iccv}
V.~Lempitsky, P.~Kohli, C.~Rother, and T.~Sharp.
\newblock Image segmentation with a bounding box prior.
\newblock In {\em ICCV}, 2009.

\bibitem{Lin2016CvprScribbleSup}
D.~Lin, J.~Dai, J.~Jia, K.~He, and J.~Sun.
\newblock Scribblesup: Scribble-supervised convolutional networks for semantic
  segmentation.
\newblock In {\em CVPR}, 2016.

\bibitem{Lin2016CvprAdelaide}
G.~Lin, C.~Shen, A.~van~dan Hengel, and I.~Reid.
\newblock Efficient piecewise training of deep structured models for semantic
  segmentation.
\newblock In {\em CVPR}, 2016.

\bibitem{Lin2014EccvCoco}
T.~Lin, M.~Maire, S.~Belongie, J.~Hays, P.~Perona, D.~Ramanan, P.~Doll\'ar, and
  C.~L. Zitnick.
\newblock Microsoft coco: Common objects in context.
\newblock In {\em ECCV}, 2014.

\bibitem{Long2015Cvpr}
J.~Long, E.~Shelhamer, and T.~Darrell.
\newblock Fully convolutional networks for semantic segmentation.
\newblock In {\em CVPR}, 2015.

\bibitem{Papandreou2015Iccv}
G.~Papandreou, L.~Chen, K.~Murphy, , and A.~L. Yuille.
\newblock Weakly- and semi-supervised learning of a dcnn for semantic image
  segmentation.
\newblock In {\em ICCV}, 2015.

\bibitem{Pathak2015Iccv}
D.~Pathak, P.~Kraehenbuehl, and T.~Darrell.
\newblock Constrained convolutional neural networks for weakly supervised
  segmentation.
\newblock In {\em ICCV}, 2015.

\bibitem{Pathak2015Iclrw}
D.~Pathak, E.~Shelhamer, J.~Long, and T.~Darrell.
\newblock Fully convolutional multi-class multiple instance learning.
\newblock In {\em ICLR workshop}, 2015.

\bibitem{Pinheiro2015Cvpr}
P.~Pinheiro and R.~Collobert.
\newblock From image-level to pixel-level labeling with convolutional network.
\newblock In {\em CVPR}, 2015.

\bibitem{Pinheiro2016Eccv}
P.~Pinheiro, T.-Y. Lin, R.~Collobert, and P.~Doll\'ar.
\newblock Learning to refine object segments.
\newblock In {\em ECCV}, 2016.

\bibitem{Pinheiro2014Icml}
P.~O. Pinheiro and R.~Collobert.
\newblock Recurrent convolutional neural networks for scene labeling.
\newblock In {\em ICML}, 2014.

\bibitem{Pinheiro2015Nips}
P.~O. Pinheiro, R.~Collobert, and P.~Doll\'ar.
\newblock Learning to segment object candidates.
\newblock In {\em NIPS}, 2015.

\bibitem{PontTuset2015ArxivMcg}
J.~Pont-Tuset, P.~Arbel\'{a}ez, J.~Barron, F.~Marques, and J.~Malik.
\newblock Multiscale combinatorial grouping for image segmentation and object
  proposal generation.
\newblock {\em arXiv preprint arXiv:1503.00848}, 2015.

\bibitem{PontTuset2015Iccv}
J.~Pont-Tuset and L.~V. Gool.
\newblock Boosting object proposals: From pascal to coco.
\newblock In {\em ICCV}, 2015.

\bibitem{Rother2004TogGrabcut}
C.~Rother, V.~Kolmogorov, and A.~Blake.
\newblock Grabcut: Interactive foreground extraction using iterated graph cuts.
\newblock In {\em ACM Trans. Graphics}, 2004.

\bibitem{Russakovsky2015Ijcv}
O.~Russakovsky, J.~Deng, H.~Su, J.~Krause, S.~Satheesh, S.~Ma, Z.~Huang,
  A.~Karpathy, A.~Khosla, M.~Bernstein, A.~C. Berg, and L.~Fei-Fei.
\newblock {ImageNet Large Scale Visual Recognition Challenge}.
\newblock {\em IJCV}, 2015.

\bibitem{Shotton2009Ijcv}
J.~Shotton, J.~Winn, C.~Rother, and A.~Criminisi.
\newblock Textonboost for image understanding: Multi-class object recognition
  and segmentation by jointly modeling texture, layout, and context.
\newblock {\em IJCV}, 2009.

\bibitem{Simonyan2015Iclr}
K.~Simonyan and A.~Zisserman.
\newblock Very deep convolutional networks for large-scale image recognition.
\newblock In {\em ICLR}, 2015.

\bibitem{Tang2015IccvSecretGrabCut}
M.~Tang, I.~{Ben Ayed}, D.~Marin, and Y.~Boykov.
\newblock Secrets of grabcut and kernel k-means.
\newblock In {\em ICCV}, 2015.

\bibitem{Taniai2015Cvpr}
T.~Taniai, Y.~Matsushita, and T.~Naemura.
\newblock Superdifferential cuts for binary energies.
\newblock In {\em CVPR}, 2015.

\bibitem{Wei2015ArXiv}
Y.~Wei, X.~Liang, Y.~Chen, X.~Shen, M.-M. Cheng, Y.~Zhao, and S.~Yan.
\newblock Stc: A simple to complex framework for weakly-supervised semantic
  segmentation.
\newblock {\em arXiv preprint arXiv:1509.03150}, 2015.

\bibitem{Xie2015Iccv}
S.~Xie and Z.~Tu.
\newblock Holistically-nested edge detection.
\newblock In {\em ICCV}, 2015.

\bibitem{Xu2015CvprWeakSegmentation}
J.~Xu, A.~Schwing, and R.~Urtasun.
\newblock Learning to segment under various forms of weak supervision.
\newblock In {\em CVPR}, 2015.

\bibitem{Xu2016Cvpr}
N.~Xu, B.~Price, S.~Cohen, J.~Yang, and T.~S. Huang.
\newblock Deep interactive object selection.
\newblock In {\em CVPR}, 2016.

\bibitem{Yu2016Iclr}
F.~Yu and V.~Koltun.
\newblock Multi-scale context aggregation by dilated convolutions.
\newblock In {\em ICLR}, 2016.

\bibitem{Yu2015ArXivLooseCut}
H.~Yu, Y.~Zhou, H.~Qian, M.~Xian, Y.~Lin, D.~Guo, K.~Zheng, K.~Abdelfatah, and
  S.~Wang.
\newblock Loosecut: Interactive image segmentation with loosely bounded boxes.
\newblock {\em arXiv preprint arXiv:1507.03060}, 2015.

\bibitem{Zheng2015IccvCrfAsRnn}
S.~Zheng, S.~Jayasumana, B.~Romera-Paredes, V.~Vineet, Z.~Su, D.~Du, C.~Huang,
  and P.~Torr.
\newblock Conditional random fields as recurrent neural networks.
\newblock In {\em ICCV}, 2015.

\end{thebibliography}
\clearpage{}

\selectlanguage{english}%

\appendix
\renewcommand{\thetable}{S\arabic{table}}\renewcommand{\thefigure}{S\arabic{figure}}
\selectlanguage{british}%

\part*{Supplementary material}

\setcounter{figure}{0}\setcounter{table}{0}

\section{\label{sec:Content}Content}

This supplementary material provides additional quantitative and qualitative
results:
\begin{itemize}
\item Section \ref{sec:Recursive-training-with} analyses the contribution
of the post-processing stages during recursive training (Figure \ref{fig:Training-labelling-from-rectangles}).
\item Section \ref{sec:Training-details-in} discusses training differences
of our approach in contrast to the related work.
\item We report a comparison of different GrabCut-like methods on Pascal
VOC12 boxes in Section \ref{sec:Box-driven-segments}.
\item Section \ref{sec:Examples-of-input} (Figure \ref{fig:grabcut-variants-examples-2-1})
shows visualization of the different variants of the proposed segmentation
inputs obtained from bounding box annotations for weakly supervised
semantic segmentation. 
\item Detailed performance of each class for semantic labelling is reported
in Section \ref{sec:Quantitative-results-for} (Table \ref{tab:Semantic-labelling-results-boxes-only}).
\item Section \ref{sec:Qualitative-results-for} provides additional qualitative
results for weakly supervised semantic segmentation on Pascal VOC12
(Figure \ref{fig:Qualitative-results-1}). 
\item Qualitative results for instance segmentation are shown in Section
\ref{sec:Qualitative-results-for-1} (Figure \ref{fig:Qualitative-results-1-1}
and Figure \ref{fig:Qualitative-results-DeepMask}). 
\end{itemize}

\section{\label{sec:Recursive-training-with}Recursive training with boxes}

\begin{figure}
\begin{centering}
\includegraphics[width=1\columnwidth]{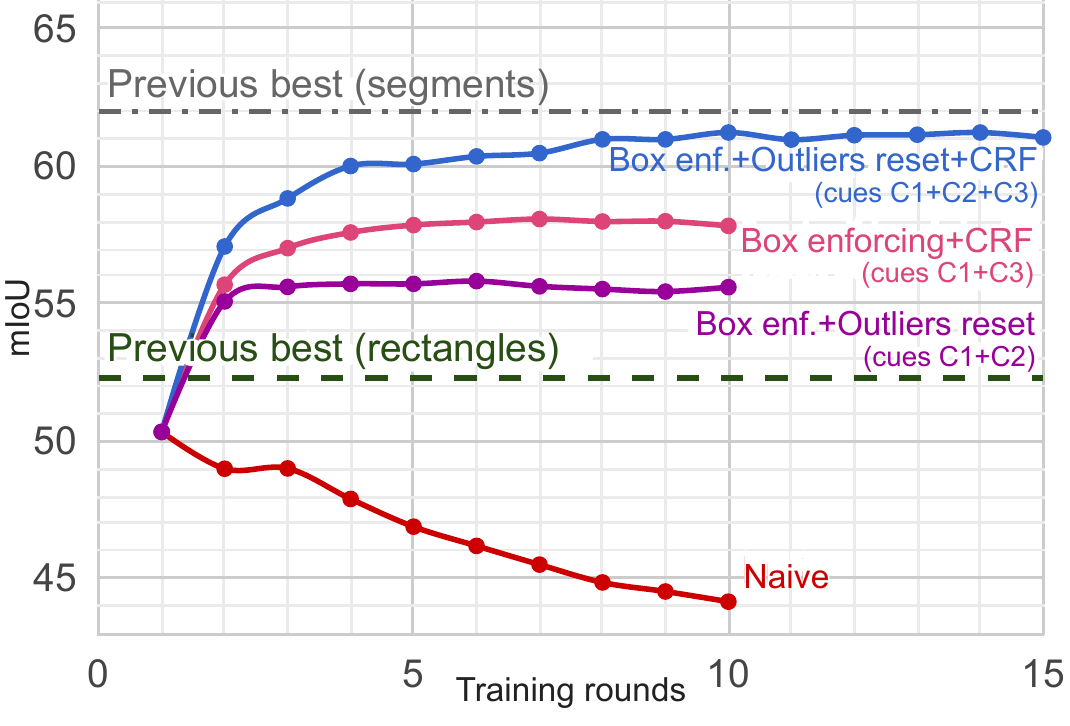}
\par\end{centering}
\caption{\label{fig:Training-labelling-from-rectangles}Recursive training
from rectangles only as input. Validation set results. All methods
use only rectangles as initial input, except ``previous best (segments)``. }
\end{figure}

In Section 3 of the main paper we recursively train a convnet directly
on the full extend of bounding box annotations as foreground labels,
disregarding post-processing stages. We name this recursive training
approach \texttt{Naive}. Using this supervision and directly applying
recursive training leads to significant degradation of the segmentation
output quality, see Figure \ref{fig:Training-labelling-from-rectangles}.

To improve the labels between the training rounds three post-processing
stages are proposed. Here we discuss them in more detail:
\begin{enumerate}
\item \textbf{Box enforcing}: Any pixel outside the box annotations is reset
to background label (cue C1, see Section 3 in the main paper).
\item \textbf{Outliers reset}: If the area of a segment is too small compared
to its corresponding bounding box (e.g. IoU$<50\%$), the box area
is reset to its initial label (fed in the first round). This enforces
a minimal area (cue C2).
\item \textbf{CRF}: As it is common practice among semantic labelling methods,
we filter the output of the network to better respect the image boundaries.
(We use DenseCRF \cite{Kraehenbuehl2011Nips} with the Deep\-Lab\-v1
parameters \cite{Chen2015Iclr}). In our weakly supervised scenario,
boundary-aware filtering is particularly useful to improve objects
delineation (cue C3).
\end{enumerate}
\vspace{-1em}

\paragraph{Results}

Figure \ref{fig:Training-labelling-from-rectangles} presents results
of the recursive training using boxes as input and shows the contribution
of the post-processing stages. We see that the naive recursive training
is ineffectual. However as soon as some constraints (box enforcing
and outliers reset, cues C1+C2) are enforced, the quality improves
dramatically after the first round of recursive training. These results
already improve over previous work considering rectangles only input
\cite{Dai2015Iccv,Papandreou2015Iccv} (both using a similar convnet
to ours) and achieve $3$ points improvement over \cite{Papandreou2015Iccv}
(from $52.5$ to $55.6$ mIoU, see Figure \ref{fig:Training-labelling-from-rectangles}
``Box enf.+Outliers reset'').

Even more, when also adding CRF filtering (+ cue C3) over the training
set, we see a steady grow after each round, stabilizing around $61\%\ \mbox{mIoU}$.
This number is surprisingly close to the best results obtained using
more sophisticated techniques \cite{Dai2015Iccv}, which achieve around
$62\%\ \mbox{mIoU}$ (see Figure \ref{fig:Training-labelling-from-rectangles}
and Table \ref{tab:Semantic-labelling-results-boxes-only}). 

Our results indicate that recursive training of a convnet is robust
to input noise as soon as appropriate care is taken to de-noise the
output between rounds, enabled by given bounding boxes and object
priors. 

\section{\label{sec:Training-details-in}Training details in comparison with
BoxSup and WSSL}

In this work we focus on box level annotations for semantic labelling
of objects. The closest related work are thus \cite{Dai2015Iccv,Papandreou2015Iccv}.
\textcolor{black}{Since all implementations use slightly different
networks and training procedures, care should be taken during comparison.}
Both \cite{Dai2015Iccv} and \cite{Papandreou2015Iccv} propose new
ways to train convnets under weak supervision. Both of the approaches
build upon the DeepLab network \cite{Chen2015Iclr}, however, there
are a few differences in the network architecture. 

WSSL \cite{Papandreou2015Iccv} employs 2 different variants of the
DeepLab architecture with small and large receptive field of view
(FOV) size. For each experiment WSSL evaluates with both architectures
and reports the best result obtained (using boxes or segments as input).
BoxSup \cite{Dai2015Iccv} uses their own implementation of the DeepLab
with the small FOV. In our approach all the experiments employ the
DeepLab architecture with the large FOV.

There are also differences in the training procedure. For SGD WSSL
uses a mini-batch of 20-30 images and fine-tunes the network for about
12 hours (number of epochs is not specified) with the standard learning
parameters (following \cite{Chen2015Iclr}). In the SGD training BoxSup
uses a mini-batch size of 20 and the learning rate is divided by 10
after every 15 epochs. The training is terminated after 45 epochs.
We use a mini-batch of 30 images for SGD and the learning rate is
divided by 10 after every 2k iterations, \textasciitilde{}6 epochs.
Our network is trained for 6k iterations, \textasciitilde{}18 epochs.

Similarly to our approach, the BoxSup method \cite{Dai2015Iccv} uses
MCG object proposals during training. However, there are important
differences. They modify the training procedure so as to denoise intermediate
outputs by randomly selecting high overlap proposals. In comparison,
our approach keeps the training procedure unmodified and simply generates
input labels. Our approach also uses ignore regions, while BoxSup
does not explore this dimension. 

WSSL \cite{Papandreou2015Iccv} proposes an expectation-maximisation
algorithm with a bias to enable the network to estimate the foreground
regions. In contrast, in our work we show that one can reach better
results without modifying the training procedure (compared to the
fully supervised case) by instead carefully generating input labels
for training from the bounding box annotations (Section 3.2 in the
main paper).

\section{\label{sec:Box-driven-segments}GrabCut variants}

As discussed in Section 3.2 in the main paper we propose to employ
box-guided instance segmentation to increase quality of the input
data. Our goal is to have weak annotations with maximal quality and
minimal loss in recall. In Section 3.1 in the main paper we explored
how far could we get with just using boxes as foreground labels. However,
to obtain results of higher quality several rounds of recursive training
are needed. Starting from less noisier object segments we would like
to reach better performance with just one training round.

For this purpose we explore different GrabCut-like \cite{Rother2004TogGrabcut}
techniques, the corresponding quantitative results are in Table \ref{tab:Grabcut-variants}.
For evaluation we use the mean IoU measure. Previous work evaluated
using the 50 images from the GrabCut dataset \cite{Rother2004TogGrabcut},
or $1\mbox{k}$ images with one salient object \cite{Cheng2015CgfDenseCut}.
The evaluation of Table \ref{tab:Grabcut-variants} compares multiple
methods over $3.4\mbox{k}$ object windows, where the objects are
not salient, have diverse sizes and occlusions level. This is a more
challenging scenario than usually considered for GrabCut-like methods.
\begin{table}[h]
\begin{centering}
\begin{tabular}{c|c|c}
 & \multirow{1}{*}{Method} & mIoU\tabularnewline
\hline 
\hline 
\multirow{5}{*}{%
\begin{tabular}{c}
GrabCut\tabularnewline
variants\tabularnewline
\end{tabular}} & DenseCut \cite{Cheng2015CgfDenseCut} & 52.5\tabularnewline
 & Bbox-Seg+CRF \cite{Papandreou2015Iccv} & 71.1\tabularnewline
 & GrabCut \cite{Rother2004TogGrabcut} & 72.9\tabularnewline
 & KGrabCut \cite{Tang2015IccvSecretGrabCut} & 73.5\tabularnewline
 & GrabCut+ & 75.2\tabularnewline
\end{tabular}
\par\end{centering}
\caption{\label{tab:Grabcut-variants}GrabCut variants, evaluated on Pascal
VOC12 validation set. See Section \ref{sec:Box-driven-segments} for
details.}
\end{table}

\texttt{GrabCut} \cite{Rother2004TogGrabcut} is the established technique
to estimate an object segment from its bounding box. To further improve
its quality we propose to use better pairwise terms. We name this
variant \texttt{GrabCut+}. Instead of the typical RGB colour difference
the pairwise terms in \texttt{GrabCut+ }are replaced by probability
of boundary as generated by HED \cite{Xie2015Iccv}. The HED boundary
detector is trained on the generic boundaries of BSDS500 \cite{ArbelaezMaireFowlkesMalikPAMI11}.
Moving from \texttt{GrabCut} to\texttt{ GrabCut+} brings a $\sim2$
points improvement, see Table \ref{tab:Grabcut-variants}.\\
We also experimented with other variants such as \texttt{DenseCut}
\cite{Cheng2015CgfDenseCut} and \texttt{KGrabCut} \cite{Tang2015IccvSecretGrabCut}
but did not obtain significant gains.

\cite{Papandreou2015Iccv} proposed to perform foreground/background
segmentation by using DenseCRF and the $20\%$ of the centre area
of the bounding box as foreground prior. This approach is denoted
\texttt{Bbox-Seg+CRF} in Table \ref{tab:Grabcut-variants} and under-performs
compared to \texttt{GrabCut }and\texttt{ GrabCut+}. 

\section{\label{sec:Examples-of-input}Examples of input segmentations}

Figure \ref{fig:grabcut-variants-examples-1} presents examples of
the considered weak annotations. This figure extends Figure 3 of the
main paper.

\section{\label{sec:Quantitative-results-for}Detailed test set results for
semantic labelling}

In Table \ref{tab:Semantic-labelling-results-boxes-only}, we present
per class results on the Pascal VOC12 test set for the methods reported
in the main paper in Table 2.

On average with our weakly supervised results we achieve $\sim95\%$
quality of full supervision across all classes when training with
VOC12 only or VOC12+COCO.

\section{\label{sec:Qualitative-results-for}Qualitative results for semantic
labelling}

Figure \ref{fig:Qualitative-results-1} presents qualitative results
for semantic labelling on Pascal VOC12. The presented semantic labelling
examples show that high quality segmentation can be achieved using
only detection bounding box annotations. This figure extends Figure
5 of the main paper.

\section{\label{sec:Qualitative-results-for-1}Qualitative results for instance
segmentations}

Figure \ref{fig:Qualitative-results-1-1} illustrates additional qualitative
results for instance segmentations given by the weakly supervised
$\mathrm{DeepMask}$ and $\mathrm{DeepLab_{BOX}}$ models. This figure
complements Figure 6 from the main paper.

Figure \ref{fig:Qualitative-results-DeepMask} shows examples of instance
segmentation given by different methods. Our proposed weakly supervised
$\mathrm{DeepMask}$ model achieves competitive performance with fully
supervised results and provides higher quality output in comparison
with box-guided segmentation techniques. The $\mathrm{DeepLab_{BOX}}$
model also provides similar results, see Table 4 in the main paper.

\begin{landscape}{\scriptsize{}}
\begin{table}[t]
\begin{centering}
{\scriptsize{}\hspace{-2.5em}}\hspace*{\fill}{\scriptsize{}}%
\begin{tabular}{c|c|c|c|c|c|c|c|c|c|c|c|c|c|c|c|c|c|c|c|c|c|c|c}
{\scriptsize{}\hspace{-0.5cm}}%
\begin{tabular}{c}
{\scriptsize{}Training}\tabularnewline
{\scriptsize{}data}\tabularnewline
\end{tabular}{\scriptsize{} \hspace{-1.8em}} & {\scriptsize{}\hspace{-1em}}%
\begin{tabular}{c}
{\scriptsize{}Super-}\tabularnewline
{\scriptsize{}vision}\tabularnewline
\end{tabular}{\scriptsize{}\hspace{-1em}} & {\scriptsize{}\hspace{-2em}Method\hspace{-2em}} & {\scriptsize{}mean} & {\scriptsize{}plane} & {\scriptsize{}bike} & {\scriptsize{}bird} & {\scriptsize{}boat} & {\scriptsize{}bottle} & {\scriptsize{}bus} & {\scriptsize{}car} & {\scriptsize{}cat} & {\scriptsize{}chair} & {\scriptsize{}cow} & {\scriptsize{}table} & {\scriptsize{}dog} & {\scriptsize{}horse} & {\scriptsize{}\hspace{-1em}}%
\begin{tabular}{c}
{\scriptsize{}motor}\tabularnewline
{\scriptsize{}bike}\tabularnewline
\end{tabular}{\scriptsize{}\hspace{-1em}} & {\scriptsize{}\hspace{-1em}}%
\begin{tabular}{c}
{\scriptsize{}per}\tabularnewline
{\scriptsize{}son}\tabularnewline
\end{tabular}{\scriptsize{}\hspace{-1em}} & {\scriptsize{}plant} & {\scriptsize{}sheep} & {\scriptsize{}sofa} & {\scriptsize{}train} & {\scriptsize{}tv}\tabularnewline
\hline 
\multirow{6}{*}{{\scriptsize{}\hspace{-0.5em}VOC12\hspace{-1em}}} & \multirow{3}{*}{{\scriptsize{}weak}} & {\scriptsize{}$\mathtt{Box}$} & {\scriptsize{}62.2} & {\scriptsize{}62.6} & {\scriptsize{}24.5} & {\scriptsize{}63.7} & {\scriptsize{}56.7} & {\scriptsize{}68.1} & {\scriptsize{}84.3} & {\scriptsize{}75.0} & {\scriptsize{}72.3} & {\scriptsize{}27.2} & {\scriptsize{}63.5} & {\scriptsize{}61.7} & {\scriptsize{}68.2} & {\scriptsize{}56.0} & {\scriptsize{}70.9} & {\scriptsize{}72.8} & {\scriptsize{}49.0} & {\scriptsize{}66.7} & {\scriptsize{}45.2} & {\scriptsize{}71.8} & {\scriptsize{}58.3}\tabularnewline
 &  & {\scriptsize{}$\mathtt{Box^{i}}$} & {\scriptsize{}63.5} & {\scriptsize{}67.7} & {\scriptsize{}25.5} & {\scriptsize{}67.3} & {\scriptsize{}58.0} & {\scriptsize{}62.8} & {\scriptsize{}83.1} & {\scriptsize{}75.1} & {\scriptsize{}78.0} & {\scriptsize{}25.5} & {\scriptsize{}64.7} & {\scriptsize{}60.8} & {\scriptsize{}74.0} & {\scriptsize{}62.9} & {\scriptsize{}74.6} & {\scriptsize{}73.3} & {\scriptsize{}50.0} & {\scriptsize{}68.5} & {\scriptsize{}43.5} & {\scriptsize{}71.6} & {\scriptsize{}56.7}\tabularnewline
 &  & {\scriptsize{}$\mathtt{M}\cap\mathtt{G+}$} & \textbf{\scriptsize{}67.5} & {\scriptsize{}78.1} & {\scriptsize{}31.1} & {\scriptsize{}72.4} & {\scriptsize{}61.0} & {\scriptsize{}67.2} & {\scriptsize{}84.2} & {\scriptsize{}78.2} & {\scriptsize{}81.7} & {\scriptsize{}27.6} & {\scriptsize{}68.5} & {\scriptsize{}62.1} & {\scriptsize{}76.9} & {\scriptsize{}70.8} & {\scriptsize{}78.0} & {\scriptsize{}76.3} & {\scriptsize{}51.7} & {\scriptsize{}78.3} & {\scriptsize{}48.3} & {\scriptsize{}74.2} & {\scriptsize{}58.6}\tabularnewline
\cline{2-24} 
 & \multirow{1}{*}{{\scriptsize{}semi}} & {\scriptsize{}$\mathtt{M}\cap\mathtt{G+}$} & {\scriptsize{}66.9} & {\scriptsize{}75.8} & {\scriptsize{}32.3} & {\scriptsize{}75.9} & {\scriptsize{}60.1} & {\scriptsize{}65.7} & {\scriptsize{}82.9} & {\scriptsize{}75.0} & {\scriptsize{}79.5} & {\scriptsize{}29.5} & {\scriptsize{}68.5} & {\scriptsize{}60.6} & {\scriptsize{}76.2} & {\scriptsize{}68.6} & {\scriptsize{}76.9} & {\scriptsize{}75.2} & {\scriptsize{}53.2} & {\scriptsize{}76.6} & {\scriptsize{}49.5} & {\scriptsize{}73.8} & {\scriptsize{}58.6}\tabularnewline
\cline{2-24} 
 & \multirow{2}{*}{{\scriptsize{}full}} & {\scriptsize{}WSSL \cite{Papandreou2015Iccv}} & {\scriptsize{}70.3} & {\scriptsize{}83.5} & {\scriptsize{}36.6} & {\scriptsize{}82.5} & {\scriptsize{}62.3} & {\scriptsize{}66.5} & {\scriptsize{}85.4} & {\scriptsize{}78.5} & {\scriptsize{}83.7} & {\scriptsize{}30.4} & {\scriptsize{}72.9} & {\scriptsize{}60.4} & {\scriptsize{}78.5} & {\scriptsize{}75.5} & {\scriptsize{}82.1} & {\scriptsize{}79.7} & {\scriptsize{}58.2} & {\scriptsize{}82.0} & {\scriptsize{}48.8} & {\scriptsize{}73.7} & {\scriptsize{}63.3}\tabularnewline
 &  & {\scriptsize{}\hspace{-0.6em}$\text{DeepLab}_{ours}$\hspace*{0.1em}\cite{Chen2015Iclr}\hspace{-0.6em}} & {\scriptsize{}\uline{70.5}} & {\scriptsize{}85.3} & {\scriptsize{}38.3} & {\scriptsize{}79.4} & {\scriptsize{}61.4} & {\scriptsize{}68.9} & {\scriptsize{}86.4} & {\scriptsize{}82.1} & {\scriptsize{}83.6} & {\scriptsize{}30.3} & {\scriptsize{}74.5} & {\scriptsize{}53.8} & {\scriptsize{}78.0} & {\scriptsize{}77.0} & {\scriptsize{}83.7} & {\scriptsize{}81.8} & {\scriptsize{}55.6} & {\scriptsize{}79.8} & {\scriptsize{}45.9} & {\scriptsize{}79.3} & {\scriptsize{}63.4}\tabularnewline
\hline 
\multirow{6}{*}{{\scriptsize{}\hspace{-0.5em}}%
\begin{tabular}{c}
{\scriptsize{}VOC12}\tabularnewline
{\scriptsize{}+}\tabularnewline
{\scriptsize{}COCO}\tabularnewline
\end{tabular}{\scriptsize{} \hspace{-1em}}} & \multirow{2}{*}{{\scriptsize{}weak}} & {\scriptsize{}$\mathtt{Box^{i}}$\textsuperscript{}} & {\scriptsize{}66.7} & {\scriptsize{}69.0} & {\scriptsize{}27.5} & {\scriptsize{}77.1} & {\scriptsize{}61.9} & {\scriptsize{}65.3} & {\scriptsize{}84.2} & {\scriptsize{}75.5} & {\scriptsize{}83.2} & {\scriptsize{}25.7} & {\scriptsize{}73.6} & {\scriptsize{}63.6} & {\scriptsize{}78.2} & {\scriptsize{}69.3} & {\scriptsize{}75.3} & {\scriptsize{}75.2} & {\scriptsize{}51.0} & {\scriptsize{}73.5} & {\scriptsize{}46.2} & {\scriptsize{}74.4} & {\scriptsize{}60.4}\tabularnewline
 &  & {\scriptsize{}$\mathtt{M}\cap\mathtt{G+}$} & \textbf{\scriptsize{}69.9} & {\scriptsize{}82.5} & {\scriptsize{}33.4} & {\scriptsize{}82.5} & {\scriptsize{}59.5} & {\scriptsize{}65.8} & {\scriptsize{}85.3} & {\scriptsize{}75.6} & {\scriptsize{}86.4} & {\scriptsize{}29.3} & {\scriptsize{}77.1} & {\scriptsize{}60.8} & {\scriptsize{}80.7} & {\scriptsize{}79.0} & {\scriptsize{}80.5} & {\scriptsize{}77.6} & {\scriptsize{}55.9} & {\scriptsize{}78.4} & {\scriptsize{}48.6} & {\scriptsize{}75.2} & {\scriptsize{}61.5}\tabularnewline
\cline{2-24} 
 & \multirow{2}{*}{{\scriptsize{}semi}} & {\scriptsize{}BoxSup \cite{Dai2015Iccv}} & {\scriptsize{}71.0} & {\scriptsize{}86.4} & {\scriptsize{}35.5} & {\scriptsize{}79.7} & {\scriptsize{}65.2} & {\scriptsize{}65.2} & {\scriptsize{}84.3} & {\scriptsize{}78.5} & {\scriptsize{}83.7} & {\scriptsize{}30.5} & {\scriptsize{}76.2} & {\scriptsize{}62.6} & {\scriptsize{}79.3} & {\scriptsize{}76.1} & {\scriptsize{}82.1} & {\scriptsize{}81.3} & {\scriptsize{}57.0} & {\scriptsize{}78.2} & {\scriptsize{}55.0} & {\scriptsize{}72.5} & {\scriptsize{}68.1}\tabularnewline
 &  & {\scriptsize{}$\mathtt{M}\cap\mathtt{G+}$} & \textbf{\scriptsize{}72.8} & {\scriptsize{}87.6} & {\scriptsize{}37.7} & {\scriptsize{}86.7} & {\scriptsize{}65.5} & {\scriptsize{}67.3} & {\scriptsize{}86.8} & {\scriptsize{}81.1} & {\scriptsize{}88.3} & {\scriptsize{}30.7} & {\scriptsize{}77.3} & {\scriptsize{}61.6} & {\scriptsize{}82.7} & {\scriptsize{}79.4} & {\scriptsize{}84.1} & {\scriptsize{}82.0} & {\scriptsize{}60.3} & {\scriptsize{}84.0} & {\scriptsize{}49.4} & {\scriptsize{}77.8} & {\scriptsize{}64.7}\tabularnewline
\cline{2-24} 
 & \multirow{2}{*}{{\scriptsize{}full}} & {\scriptsize{}WSSL \cite{Papandreou2015Iccv}} & {\scriptsize{}72.7 } & {\scriptsize{}89.1} & {\scriptsize{}38.3} & {\scriptsize{}88.1} & {\scriptsize{}63.3} & {\scriptsize{}69.7} & {\scriptsize{}87.1} & {\scriptsize{}83.1} & {\scriptsize{}85.0} & {\scriptsize{}29.3} & {\scriptsize{}76.5} & {\scriptsize{}56.5} & {\scriptsize{}79.8} & {\scriptsize{}77.9} & {\scriptsize{}85.8} & {\scriptsize{}82.4} & {\scriptsize{}57.4} & {\scriptsize{}84.3} & {\scriptsize{}54.9} & {\scriptsize{}80.5} & {\scriptsize{}64.1}\tabularnewline
 &  & {\scriptsize{}\hspace{-0.6em}$\text{DeepLab}_{ours}$\hspace*{0.1em}\cite{Chen2015Iclr}\hspace{-0.6em}} & {\scriptsize{}\uline{73.2}} & {\scriptsize{}88.8} & {\scriptsize{}37.3} & {\scriptsize{}83.8} & {\scriptsize{}66.5} & {\scriptsize{}70.1} & {\scriptsize{}89.0} & {\scriptsize{}81.4} & {\scriptsize{}87.3} & {\scriptsize{}30.2} & {\scriptsize{}78.8} & {\scriptsize{}61.6} & {\scriptsize{}82.4} & {\scriptsize{}82.3} & {\scriptsize{}84.4} & {\scriptsize{}82.2} & {\scriptsize{}59.1} & {\scriptsize{}85.0} & {\scriptsize{}50.8} & {\scriptsize{}79.7} & {\scriptsize{}63.8}\tabularnewline
\end{tabular}\hspace*{\fill}{\scriptsize{}\vspace{0.5em}
}
\par\end{centering}{\scriptsize \par}
{\scriptsize{}\caption{\label{tab:Semantic-labelling-results-boxes-only}Per class semantic
labelling results for methods trained using Pascal VOC12 and COCO.
Test set results. Bold indicates the best performance with the same
supervision and training data. $\mathtt{M}\cap\mathtt{G+}$ denotes
the weakly or semi supervised model trained with $\mbox{MCG}\,\cap\,\mbox{Grabcut+}$.}
\vspace{-1.5em}
}{\scriptsize \par}
\end{table}
\end{landscape}

\captionsetup[subfigure]{labelformat=empty,font=scriptsize}

\begin{figure*}[t]
\vspace{-1em}

\hspace*{\fill}\subfloat[\label{fig:grabcut-input-image-1}Input image]{\centering{}\includegraphics[width=0.18\textwidth,height=0.16\textheight]{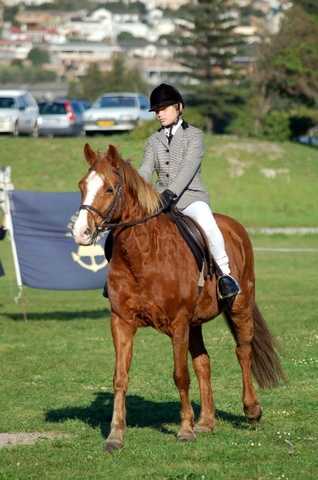}}\quad{}\subfloat[\label{fig:grabcut-gt-image-1}\negthickspace{}Ground~truth]{\centering{}\includegraphics[width=0.18\textwidth,height=0.16\textheight]{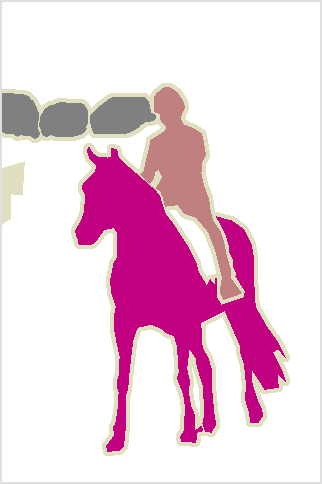}}\quad{}\subfloat[\label{fig:grabcut-variant-rectangle-1}$\mathtt{Box}$]{\centering{}\includegraphics[width=0.18\textwidth,height=0.16\textheight]{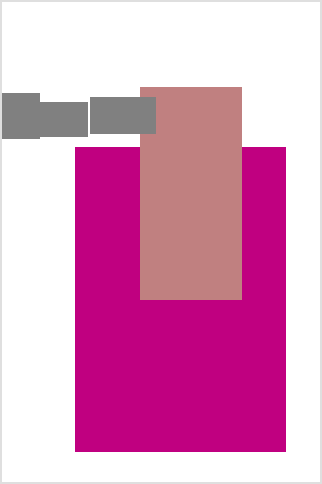}}\quad{}\subfloat[\label{fig:grabcut-variant-densecut-1}$\mathtt{Box^{i}}$]{\centering{}\includegraphics[width=0.18\textwidth,height=0.16\textheight]{figures/weak_annot_sup/2007_005331_img_bb20}}\hspace*{\fill}

\vspace{-1em}

\hspace*{\fill}\subfloat[\label{fig:grabcut-variant-kernel-grabcut-1}\negthickspace{}Bbox-Seg+CRF\negthickspace{}]{\centering{}\includegraphics[width=0.18\textwidth,height=0.16\textheight]{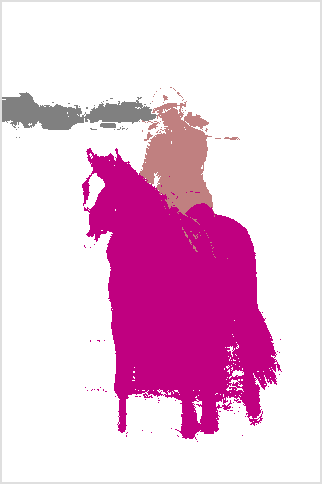}}\quad{}\subfloat[\label{fig:grabcut-variant-MCG-1}MCG]{\centering{}\includegraphics[width=0.18\textwidth,height=0.16\textheight]{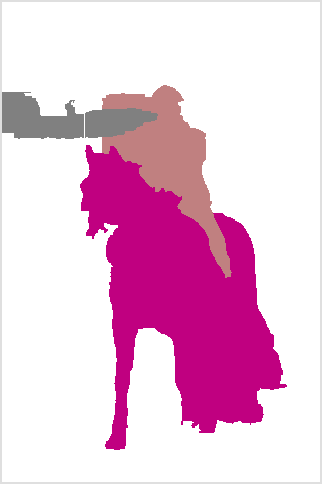}}\quad{}\subfloat[\label{fig:grabcut-variant-densecut}DenseCut]{\centering{}\includegraphics[width=0.18\textwidth,height=0.16\textheight]{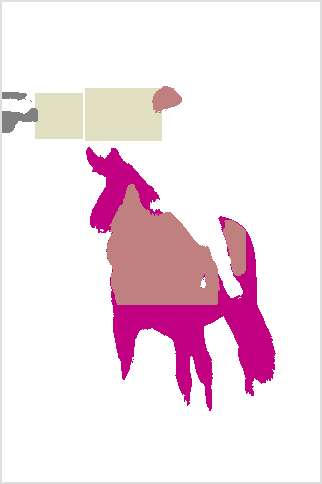}}\quad{}\subfloat[\label{fig:grabcut-variant-grabcut-1}GrabCut]{\centering{}\includegraphics[width=0.18\textwidth,height=0.16\textheight]{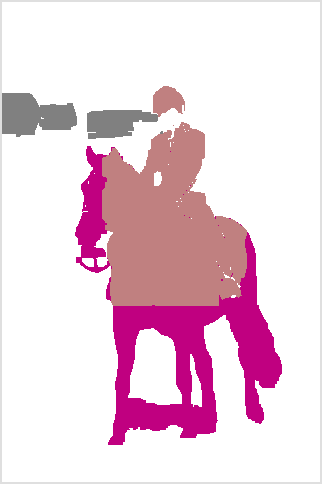}}\hspace*{\fill}

\vspace{-1em}

\hspace*{\fill}\subfloat[\label{fig:grabcut-variant-kernel-grabcut}KGrabCut]{\centering{}\includegraphics[width=0.18\textwidth,height=0.16\textheight]{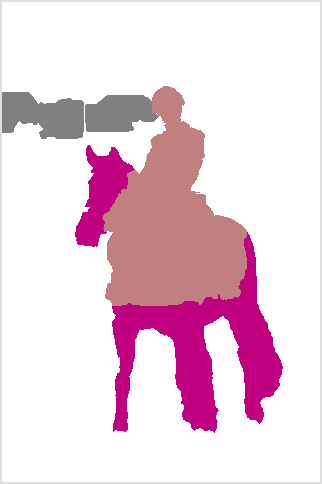}}\quad{}\subfloat[\label{fig:grabcut-variant-grabcut-with-strong-edges-1}GrabCut+]{\centering{}\includegraphics[width=0.18\textwidth,height=0.16\textheight]{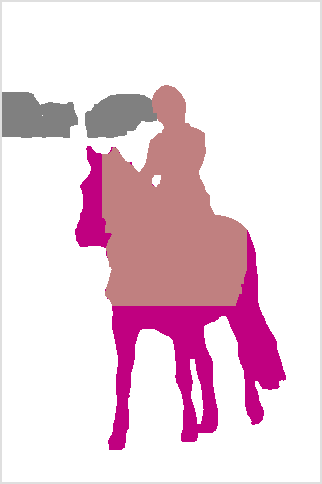}}\quad{}\subfloat[\label{fig:grabcut-variant-grabcut+-perturbed-1}$\mathtt{GrabCut+^{i}}$]{\centering{}\includegraphics[width=0.18\textwidth,height=0.16\textheight]{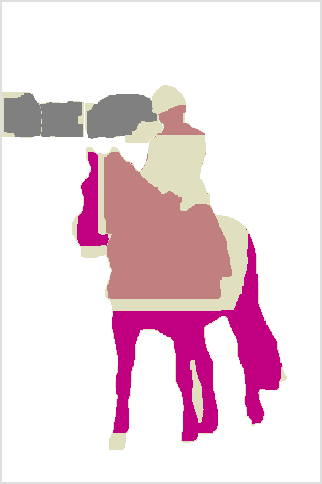}}\quad{}\subfloat[\label{fig:grabcut-variant-M=000026G-1}$\mbox{M}\,\cap\,\mbox{G+}$]{\centering{}\includegraphics[width=0.18\textwidth,height=0.16\textheight]{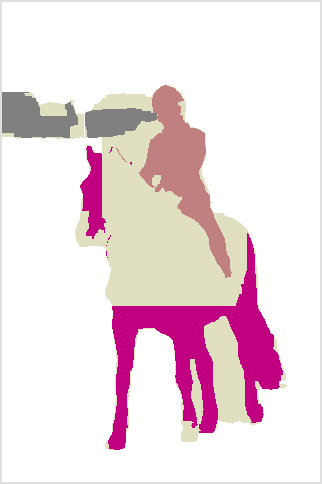}}\hspace*{\fill}

\vspace{1em}

\hspace*{\fill}\subfloat[\label{fig:grabcut-input-image-3}Input image]{\centering{}\includegraphics[width=0.18\textwidth]{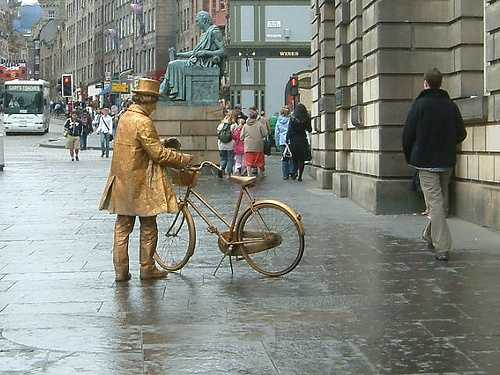}}\quad{}\subfloat[\label{fig:grabcut-gt-image-3}\negthickspace{}Ground~truth]{\centering{}\includegraphics[width=0.18\textwidth]{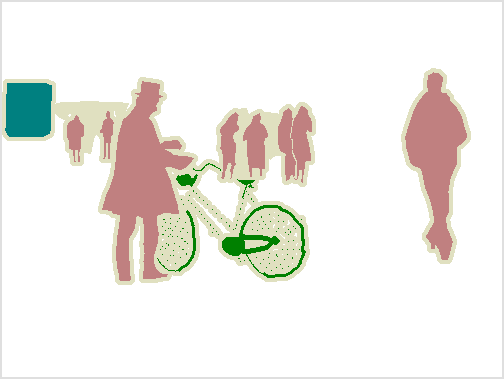}}\quad{}\subfloat[\label{fig:grabcut-variant-rectangle-3}$\mathtt{Box}$]{\centering{}\includegraphics[width=0.18\textwidth]{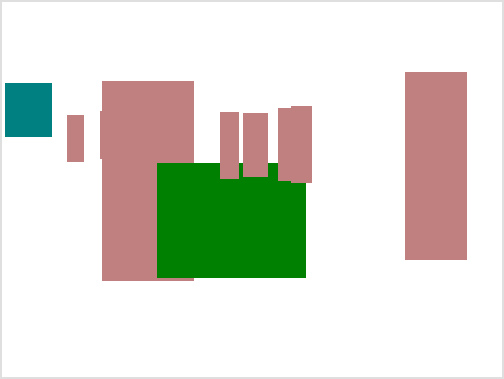}}\quad{}\subfloat[\label{fig:grabcut-variant-densecut-1-3}$\mathtt{Box^{i}}$]{\centering{}\includegraphics[width=0.18\textwidth]{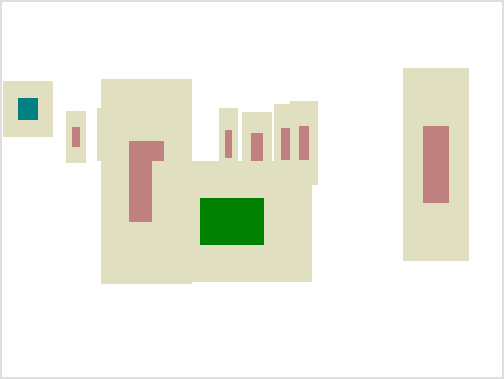}}\hspace*{\fill}

\vspace{-1em}

\hspace*{\fill}\subfloat[\label{fig:grabcut-variant-kernel-grabcut-1-3}\negthickspace{}Bbox-Seg+CRF\negthickspace{}]{\centering{}\includegraphics[width=0.18\textwidth]{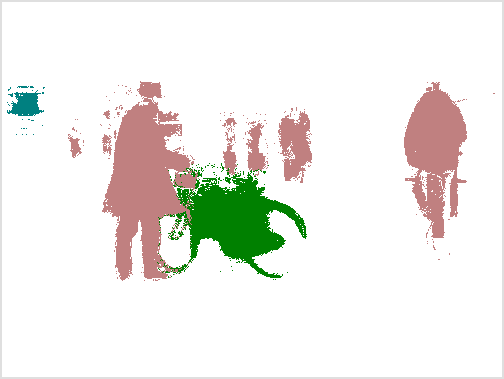}}\quad{}\subfloat[\label{fig:grabcut-variant-MCG-3}MCG]{\centering{}\includegraphics[width=0.18\textwidth]{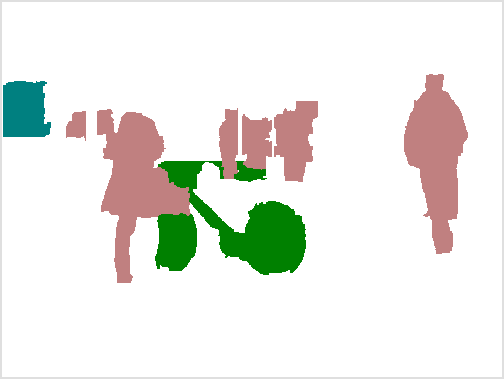}}\quad{}\subfloat[\label{fig:grabcut-variant-densecut-4}DenseCut]{\centering{}\includegraphics[width=0.18\textwidth]{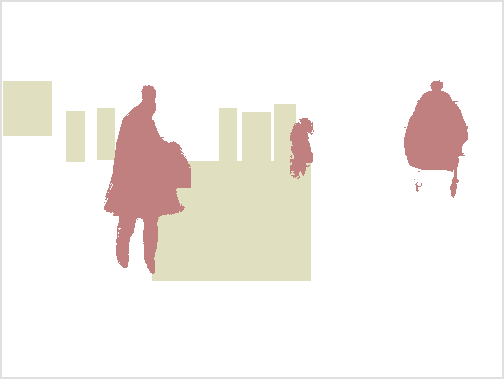}}\quad{}\subfloat[\label{fig:grabcut-variant-grabcut-3}GrabCut]{\centering{}\includegraphics[width=0.18\textwidth]{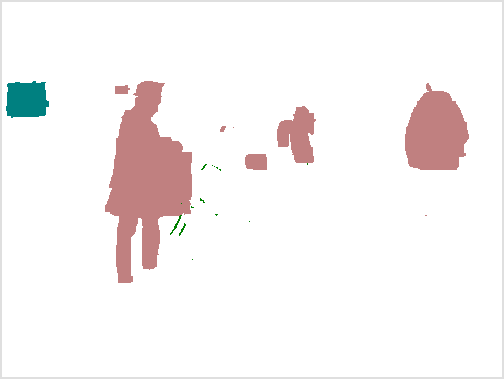}}\hspace*{\fill}

\vspace{-1em}

\hspace*{\fill}\subfloat[\label{fig:grabcut-variant-kernel-grabcut-4}KGrabCut]{\centering{}\includegraphics[width=0.18\textwidth]{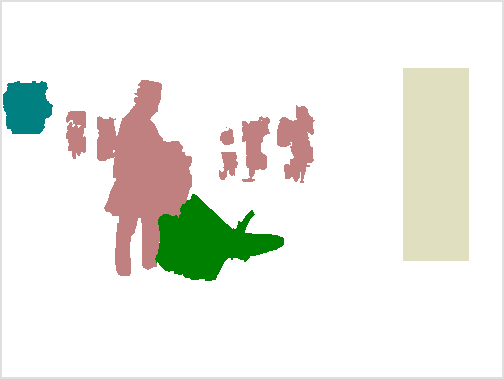}}\quad{}\subfloat[\label{fig:grabcut-variant-grabcut-with-strong-edges-3}GrabCut+]{\centering{}\includegraphics[width=0.18\textwidth]{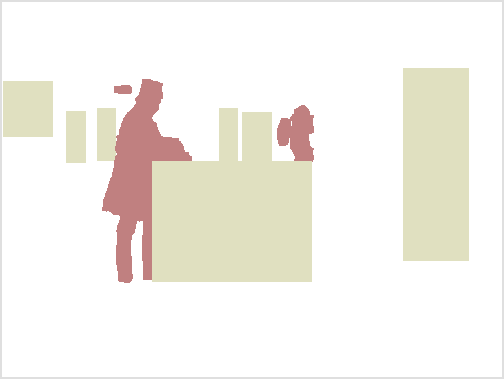}}\quad{}\subfloat[\label{fig:grabcut-variant-grabcut+-perturbed-3}$\mathtt{GrabCut+^{i}}$]{\centering{}\includegraphics[width=0.18\textwidth]{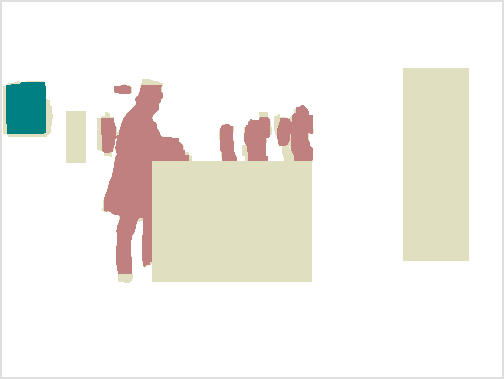}}\quad{}\subfloat[\label{fig:grabcut-variant-M=000026G-3}$\mbox{M}\,\cap\,\mbox{G+}$]{\centering{}\includegraphics[width=0.18\textwidth]{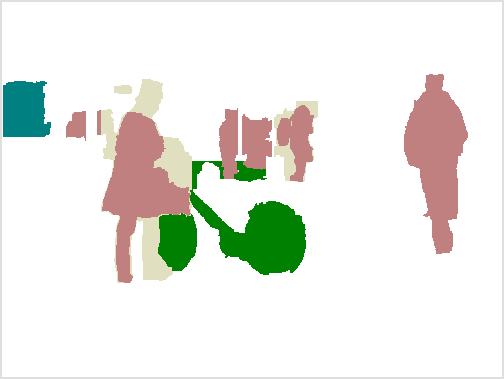}}\hspace*{\fill}

\vspace{-0.5em}

\caption{\label{fig:grabcut-variants-examples-1}Different segmentations obtained
starting from a bounding box. White is background and ignore regions
are beige. $\mbox{M}\,\cap\,\mbox{G+}$ denotes $\mbox{MCG}\,\cap\,\mbox{Grabcut+}$.}
\vspace{-1em}
\end{figure*}

\setcounter{figure}{1}

\begin{figure*}[t]
\vspace{-1em}

\hspace*{\fill}\subfloat[\label{fig:grabcut-input-image-4}Input image]{\centering{}\includegraphics[width=0.19\textwidth,height=0.15\textheight]{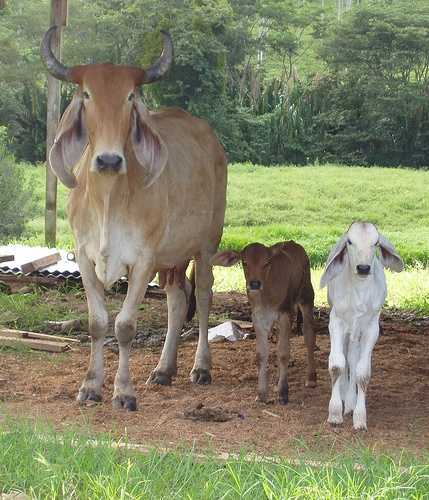}}\quad{}\subfloat[\label{fig:grabcut-gt-image-4}\negthickspace{}Ground~truth]{\centering{}\includegraphics[width=0.19\textwidth,height=0.15\textheight]{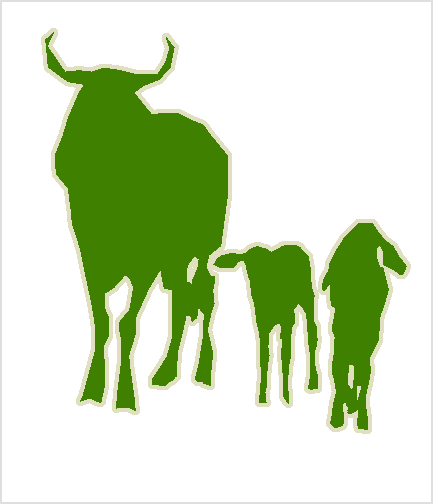}}\quad{}\subfloat[\label{fig:grabcut-variant-rectangle-4}$\mathtt{Box}$]{\centering{}\includegraphics[width=0.19\textwidth,height=0.15\textheight]{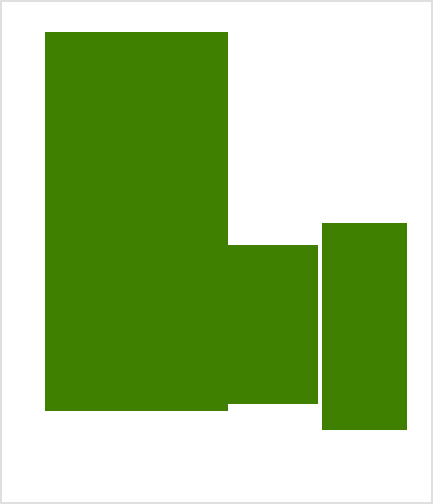}}\quad{}\subfloat[\label{fig:grabcut-variant-densecut-1-4}$\mathtt{Box^{i}}$]{\centering{}\includegraphics[width=0.19\textwidth,height=0.15\textheight]{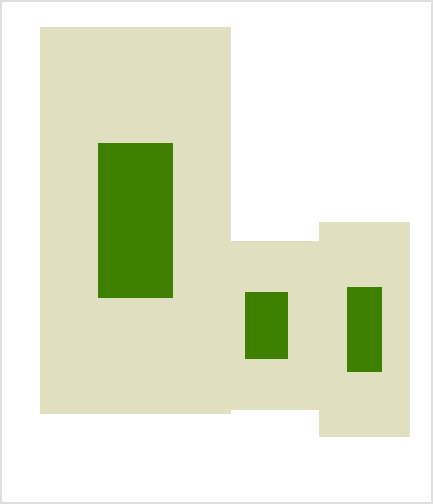}}\hspace*{\fill}

\vspace{-1em}

\hspace*{\fill}\subfloat[\label{fig:grabcut-variant-kernel-grabcut-1-4}\negthickspace{}Bbox-Seg+CRF\negthickspace{}]{\centering{}\includegraphics[width=0.19\textwidth,height=0.15\textheight]{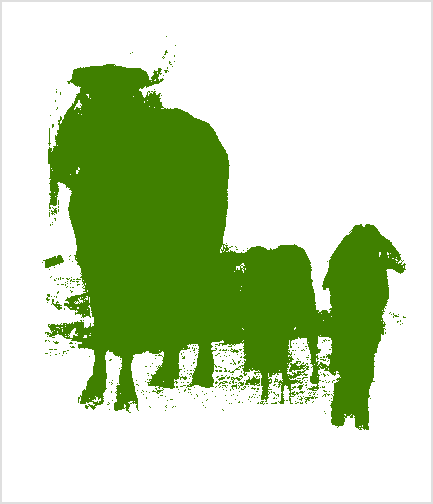}}\quad{}\subfloat[\label{fig:grabcut-variant-MCG-4}MCG]{\centering{}\includegraphics[width=0.19\textwidth,height=0.15\textheight]{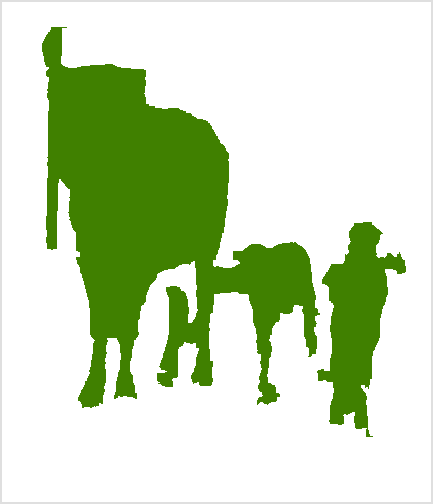}}\quad{}\subfloat[\label{fig:grabcut-variant-densecut-5}DenseCut]{\centering{}\includegraphics[width=0.19\textwidth,height=0.15\textheight]{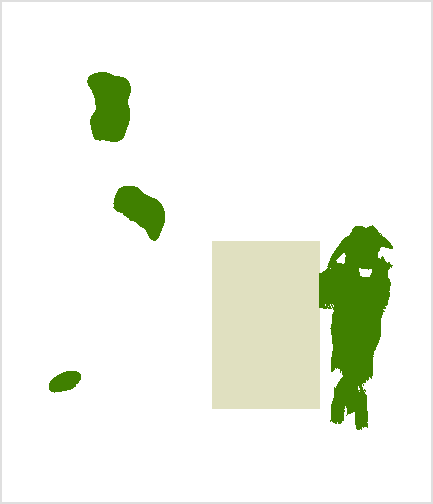}}\quad{}\subfloat[\label{fig:grabcut-variant-grabcut-4}GrabCut]{\centering{}\includegraphics[width=0.19\textwidth,height=0.15\textheight]{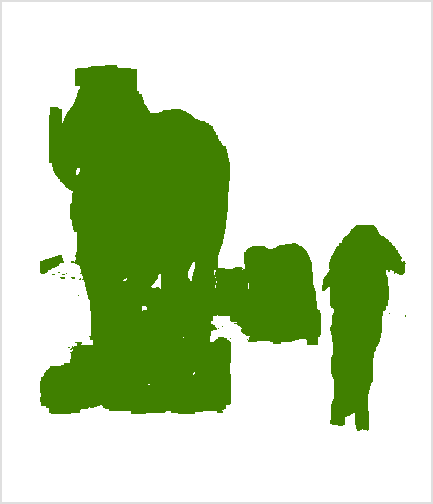}}\hspace*{\fill}

\vspace{-1em}

\hspace*{\fill}\subfloat[\label{fig:grabcut-variant-kernel-grabcut-5}KGrabCut]{\centering{}\includegraphics[width=0.19\textwidth,height=0.15\textheight]{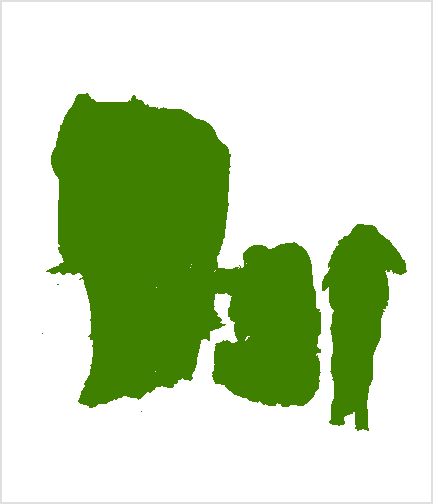}}\quad{}\subfloat[\label{fig:grabcut-variant-grabcut-with-strong-edges-4}GrabCut+]{\centering{}\includegraphics[width=0.19\textwidth,height=0.15\textheight]{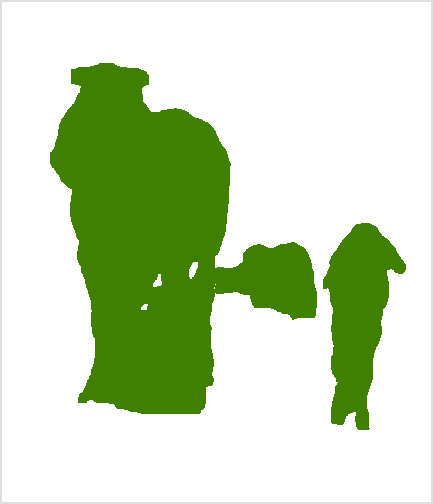}}\quad{}\subfloat[\label{fig:grabcut-variant-grabcut+-perturbed-4}$\mathtt{GrabCut+^{i}}$]{\centering{}\includegraphics[width=0.19\textwidth,height=0.15\textheight]{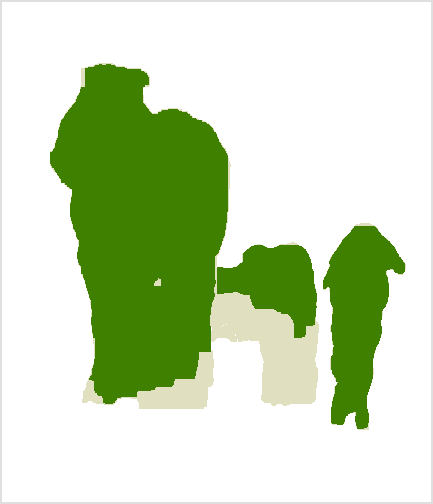}}\quad{}\subfloat[\label{fig:grabcut-variant-M=000026G-4}$\mbox{M}\,\cap\,\mbox{G+}$]{\centering{}\includegraphics[width=0.19\textwidth,height=0.15\textheight]{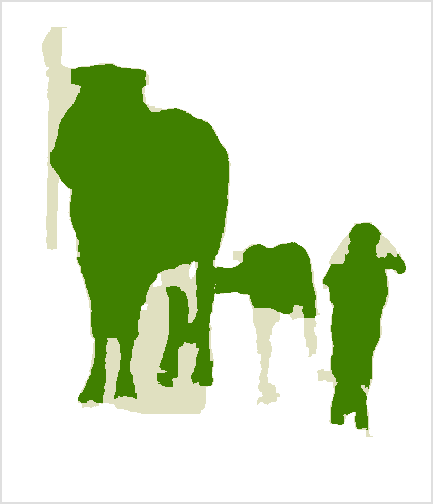}}\hspace*{\fill}

\vspace{1em}

\hspace*{\fill}\subfloat[\label{fig:grabcut-input-image-3-1}Input image]{\centering{}\includegraphics[width=0.19\textwidth]{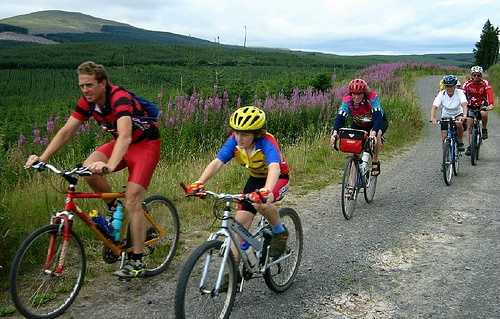}}\quad{}\subfloat[\label{fig:grabcut-gt-image-3-1}\negthickspace{}Ground~truth]{\centering{}\includegraphics[width=0.19\textwidth]{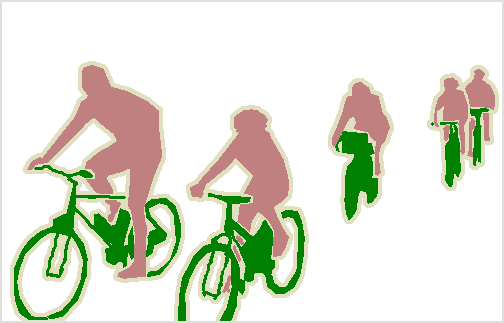}}\quad{}\subfloat[\label{fig:grabcut-variant-rectangle-3-1}$\mathtt{Box}$]{\centering{}\includegraphics[width=0.19\textwidth]{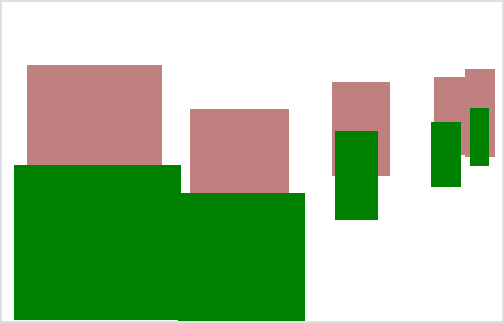}}\quad{}\subfloat[\label{fig:grabcut-variant-densecut-1-3-1}$\mathtt{Box^{i}}$]{\centering{}\includegraphics[width=0.19\textwidth]{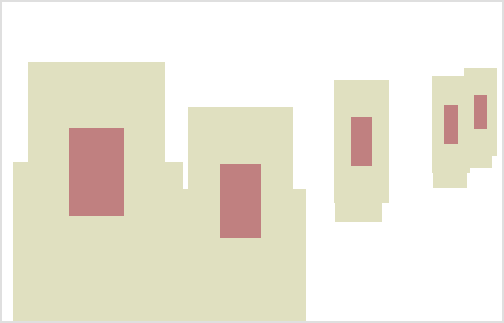}}\hspace*{\fill}

\vspace{-1em}

\hspace*{\fill}\subfloat[\label{fig:grabcut-variant-kernel-grabcut-1-3-1}\negthickspace{}Bbox-Seg+CRF\negthickspace{}]{\centering{}\includegraphics[width=0.19\textwidth]{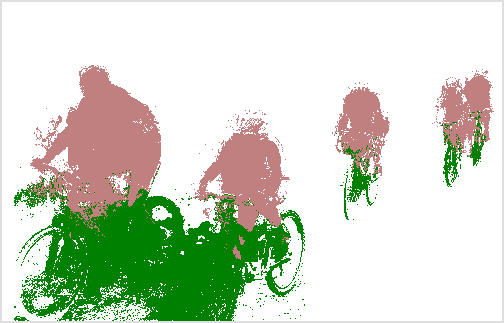}}\quad{}\subfloat[\label{fig:grabcut-variant-MCG-3-1}MCG]{\centering{}\includegraphics[width=0.19\textwidth]{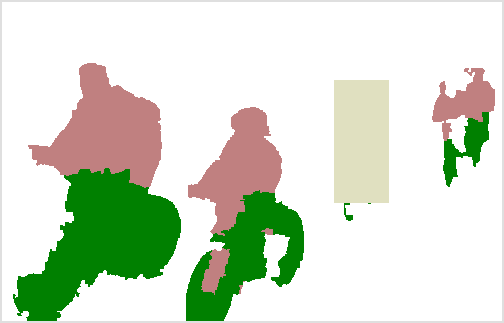}}\quad{}\subfloat[\label{fig:grabcut-variant-densecut-4-1}DenseCut]{\centering{}\includegraphics[width=0.19\textwidth]{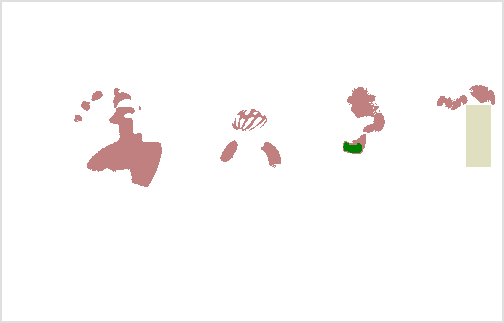}}\quad{}\subfloat[\label{fig:grabcut-variant-grabcut-3-1}GrabCut]{\centering{}\includegraphics[width=0.19\textwidth]{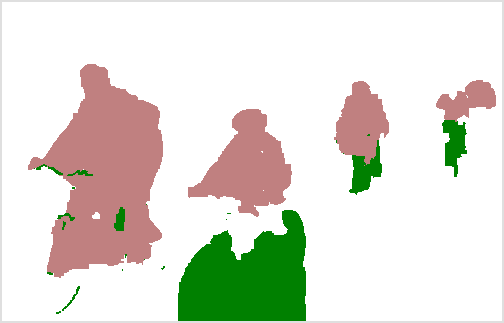}}\hspace*{\fill}

\vspace{-1em}

\hspace*{\fill}\subfloat[\label{fig:grabcut-variant-kernel-grabcut-4-1}KGrabCut]{\centering{}\includegraphics[width=0.19\textwidth]{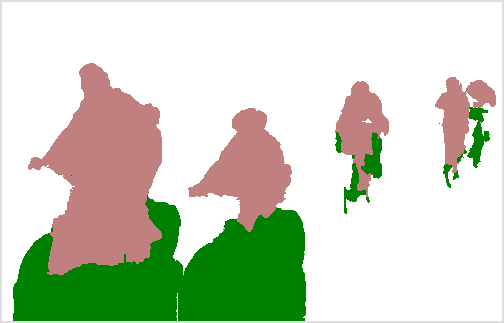}}\quad{}\subfloat[\label{fig:grabcut-variant-grabcut-with-strong-edges-3-1}GrabCut+]{\centering{}\includegraphics[width=0.19\textwidth]{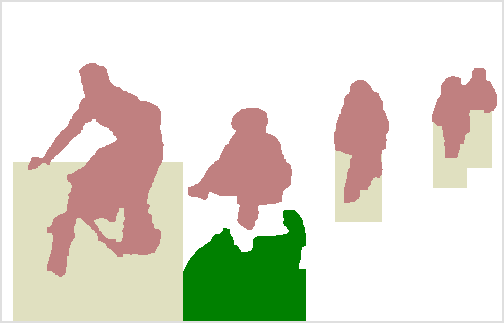}}\quad{}\subfloat[\label{fig:grabcut-variant-grabcut+-perturbed-3-1}$\mathtt{GrabCut+^{i}}$]{\centering{}\includegraphics[width=0.19\textwidth]{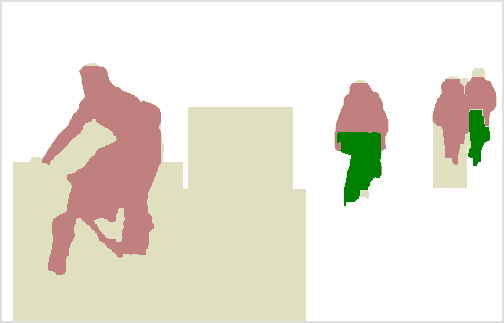}}\quad{}\subfloat[\label{fig:grabcut-variant-M=000026G-3-1}$\mbox{M}\,\cap\,\mbox{G+}$]{\centering{}\includegraphics[width=0.19\textwidth]{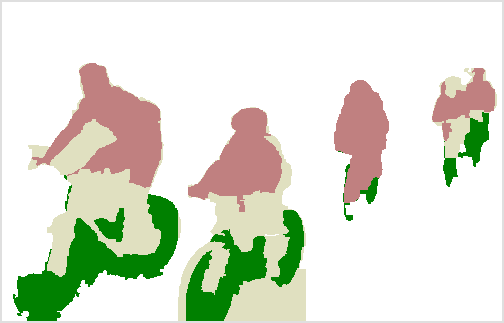}}\hspace*{\fill}

\vspace{-0.5em}

\caption{\label{fig:grabcut-variants-examples-1}Different segmentations obtained
starting from a bounding box. White is background and ignore regions
are beige. $\mbox{M}\,\cap\,\mbox{G+}$ denotes $\mbox{MCG}\,\cap\,\mbox{Grabcut+}$.}
\vspace{-1em}
\end{figure*}

\setcounter{figure}{1}

\begin{figure*}[t]
{\scriptsize{}\vspace{-3em}
}{\scriptsize \par}

{\scriptsize{}\hspace*{\fill}}\subfloat[\label{fig:grabcut-input-image-2}Input image]{\centering{}{\scriptsize{}\includegraphics[width=0.18\textwidth,height=0.19\textheight]{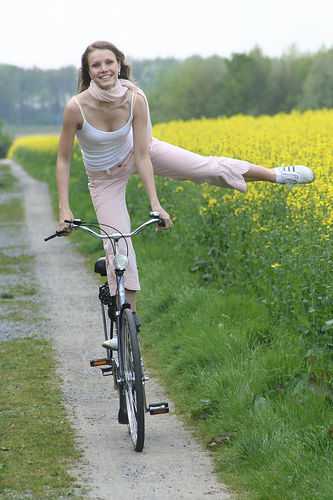}}}{\scriptsize{}\quad{}}\subfloat[\label{fig:grabcut-gt-image-2}\negthickspace{}Ground~truth]{\centering{}{\scriptsize{}\includegraphics[width=0.18\textwidth,height=0.19\textheight]{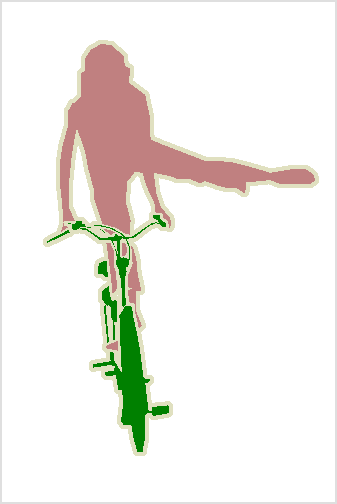}}}{\scriptsize{}\quad{}}\subfloat[\label{fig:grabcut-variant-rectangle-2}$\mathtt{Box}$]{\centering{}{\scriptsize{}\includegraphics[width=0.18\textwidth,height=0.19\textheight]{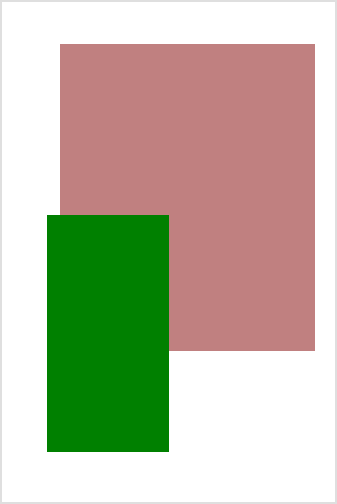}}}{\scriptsize{}\quad{}}\subfloat[\label{fig:grabcut-variant-densecut-1-2}$\mathtt{Box^{i}}$]{\centering{}{\scriptsize{}\includegraphics[width=0.18\textwidth,height=0.19\textheight]{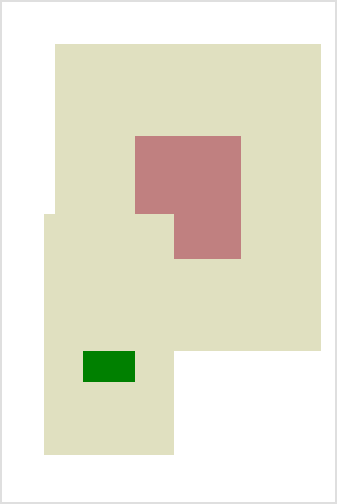}}}{\scriptsize{}\hspace*{\fill}}{\scriptsize \par}

{\scriptsize{}\vspace{-1em}
}{\scriptsize \par}

{\scriptsize{}\hspace*{\fill}}\subfloat[\label{fig:grabcut-variant-kernel-grabcut-1-2}\negthickspace{}\negthickspace{}{\scriptsize{}Bbox-Seg+CRF}\negthickspace{}\negthickspace{}]{\centering{}{\scriptsize{}\includegraphics[width=0.18\textwidth,height=0.19\textheight]{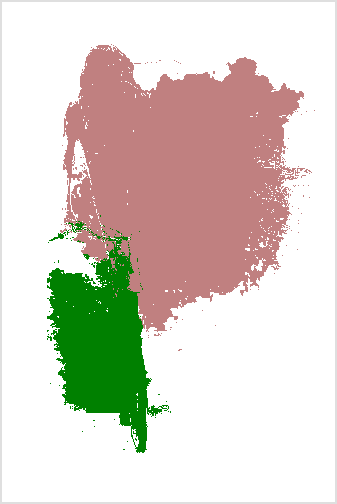}}}{\scriptsize{}\quad{}}\subfloat[\label{fig:grabcut-variant-MCG-2}MCG]{\centering{}{\scriptsize{}\includegraphics[width=0.18\textwidth,height=0.19\textheight]{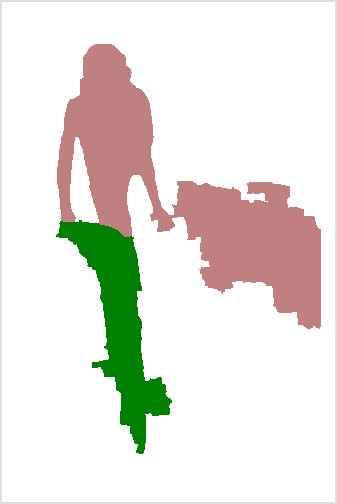}}}{\scriptsize{}\quad{}}\subfloat[\label{fig:grabcut-variant-densecut-3}DenseCut]{\centering{}{\scriptsize{}\includegraphics[width=0.18\textwidth,height=0.19\textheight]{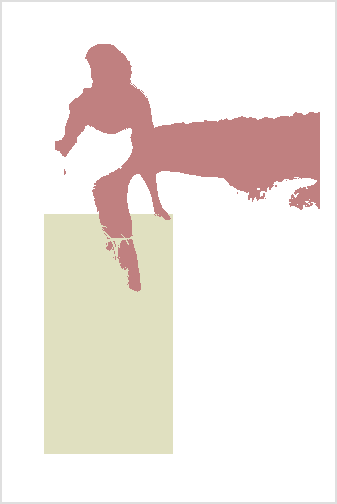}}}{\scriptsize{}\quad{}}\subfloat[\label{fig:grabcut-variant-grabcut-2}GrabCut]{\centering{}{\scriptsize{}\includegraphics[width=0.18\textwidth,height=0.19\textheight]{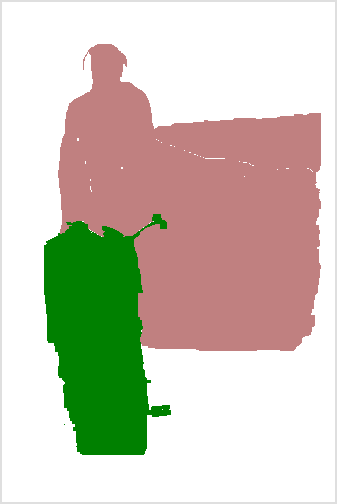}}}{\scriptsize{}\hspace*{\fill}}{\scriptsize \par}

{\scriptsize{}\vspace{-1em}
}{\scriptsize \par}

{\scriptsize{}\hspace*{\fill}}\subfloat[\label{fig:grabcut-variant-kernel-grabcut-3}KGrabCut]{\centering{}{\scriptsize{}\includegraphics[width=0.18\textwidth,height=0.19\textheight]{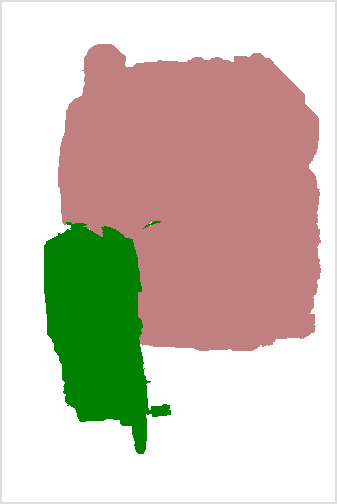}}}{\scriptsize{}\quad{}}\subfloat[\label{fig:grabcut-variant-grabcut-with-strong-edges-2}GrabCut+]{\centering{}{\scriptsize{}\includegraphics[width=0.18\textwidth,height=0.19\textheight]{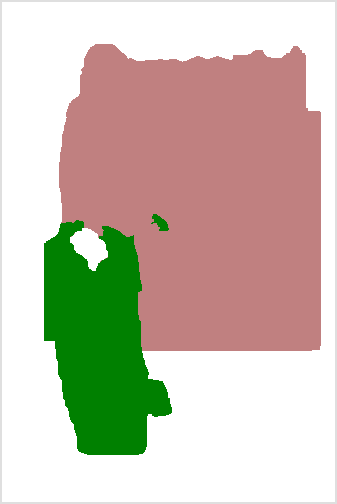}}}{\scriptsize{}\quad{}}\subfloat[\label{fig:grabcut-variant-grabcut+-perturbed-2}$\mathtt{GrabCut+^{i}}$]{\centering{}{\scriptsize{}\includegraphics[width=0.18\textwidth,height=0.19\textheight]{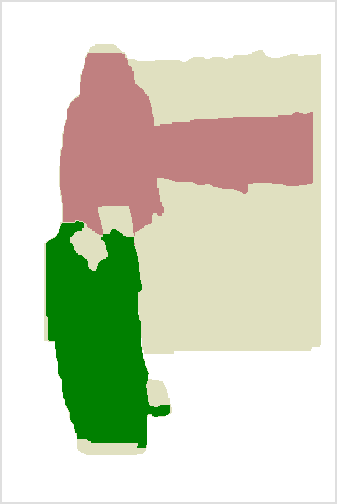}}}{\scriptsize{}\quad{}}\subfloat[\label{fig:grabcut-variant-M=000026G-2}$\mbox{M}\,\cap\,\mbox{G+}$]{\centering{}{\scriptsize{}\includegraphics[width=0.18\textwidth,height=0.19\textheight]{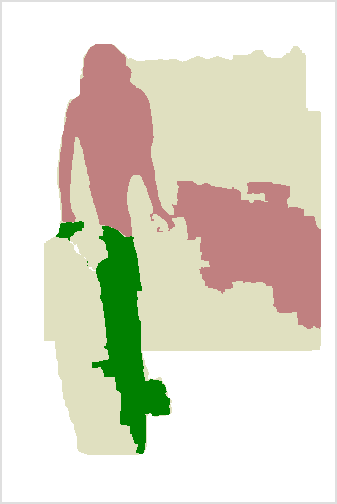}}}{\scriptsize{}\hspace*{\fill}}{\scriptsize \par}

\vspace{0.5em}

\hspace*{\fill}\subfloat[\label{fig:grabcut-input-image-1}Input image]{\centering{}\includegraphics[width=0.18\textwidth]{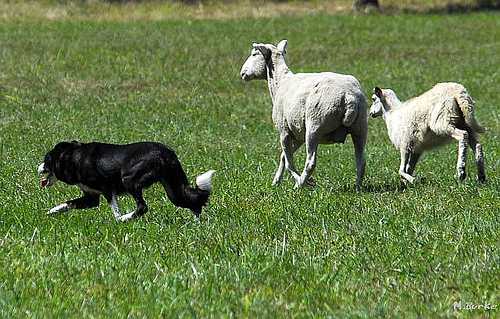}}\quad{}\subfloat[\label{fig:grabcut-gt-image-1}\negthickspace{}Ground~truth]{\centering{}\includegraphics[width=0.18\textwidth]{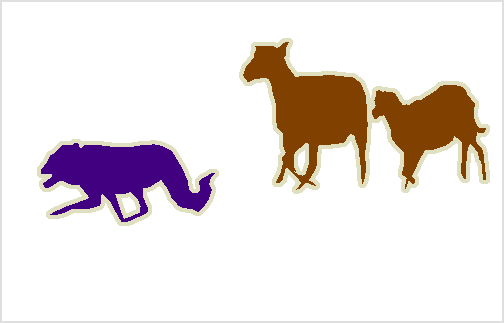}}\quad{}\subfloat[\label{fig:grabcut-variant-rectangle-1}$\mathtt{Box}$]{\centering{}\includegraphics[width=0.18\textwidth]{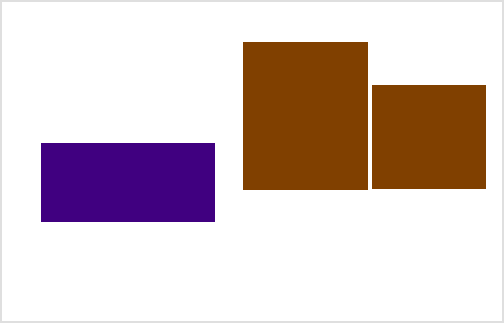}}\quad{}\subfloat[\label{fig:grabcut-variant-densecut-1-1}$\mathtt{Box^{i}}$]{\centering{}\includegraphics[width=0.18\textwidth]{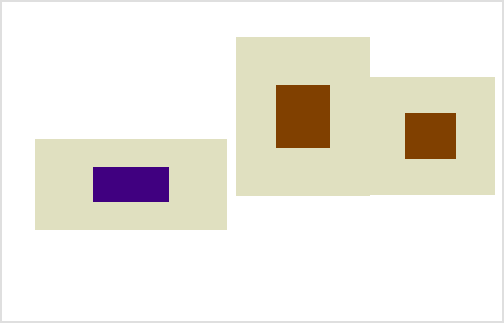}}\hspace*{\fill}

\vspace{-1em}

\hspace*{\fill}\subfloat[\label{fig:grabcut-variant-kernel-grabcut-1-1}\negthickspace{}Bbox-Seg+CRF\negthickspace{}]{\centering{}\includegraphics[width=0.18\textwidth]{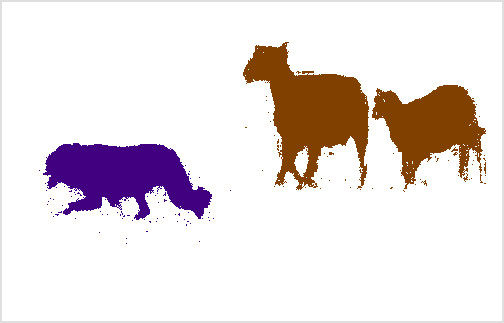}}\quad{}\subfloat[\label{fig:grabcut-variant-MCG-1}MCG]{\centering{}\includegraphics[width=0.18\textwidth]{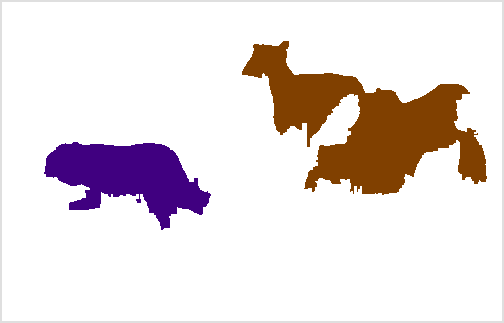}}\quad{}\subfloat[\label{fig:grabcut-variant-densecut-2}DenseCut]{\centering{}\includegraphics[width=0.18\textwidth]{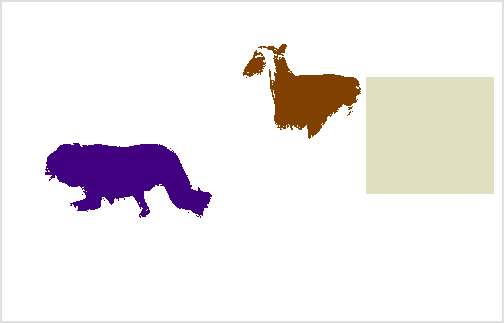}}\quad{}\subfloat[\label{fig:grabcut-variant-grabcut-1}GrabCut]{\centering{}\includegraphics[width=0.18\textwidth]{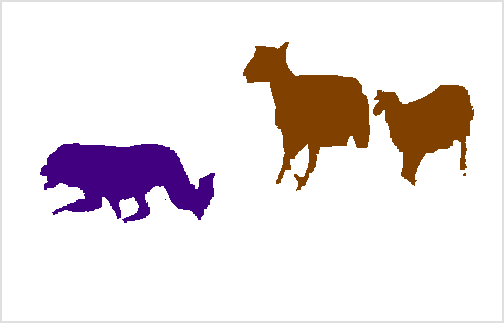}}\hspace*{\fill}

\vspace{-1em}

\hspace*{\fill}\subfloat[\label{fig:grabcut-variant-kernel-grabcut-2}KGrabCut]{\centering{}\includegraphics[width=0.18\textwidth]{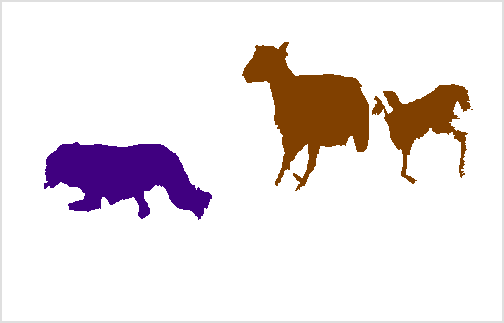}}\quad{}\subfloat[\label{fig:grabcut-variant-grabcut-with-strong-edges-1}GrabCut+]{\centering{}\includegraphics[width=0.18\textwidth]{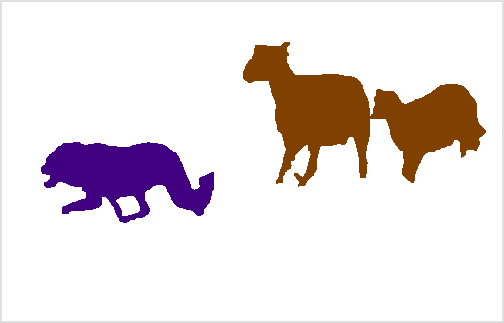}}\quad{}\subfloat[\label{fig:grabcut-variant-grabcut+-perturbed-1}$\mathtt{GrabCut+^{i}}$]{\centering{}\includegraphics[width=0.18\textwidth]{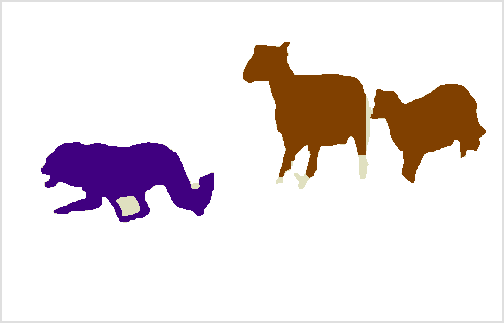}}\quad{}\subfloat[\label{fig:grabcut-variant-M=000026G-1}$\mbox{M}\,\cap\,\mbox{G+}$]{\centering{}\includegraphics[width=0.18\textwidth]{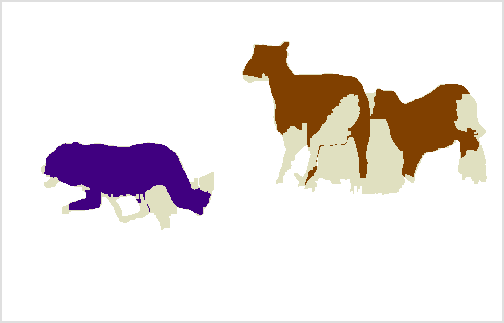}}\hspace*{\fill}

{\scriptsize{}\vspace{-0.5em}
}{\scriptsize \par}

\caption{\label{fig:grabcut-variants-examples-2}Different segmentations obtained
starting from a bounding box. White is background and ignore regions
are beige. $\mbox{M}\,\cap\,\mbox{G+}$ denotes $\mbox{MCG}\,\cap\,\mbox{Grabcut+}$.}
\vspace{-1em}
\end{figure*}

\setcounter{figure}{1}

\begin{figure*}[t]
{\scriptsize{}\vspace{-1em}
}{\scriptsize \par}

{\scriptsize{}\hspace*{\fill}}\subfloat[\label{fig:grabcut-input-image-2-1}Input image]{\centering{}{\scriptsize{}\includegraphics[width=0.18\textwidth,height=0.08\textheight]{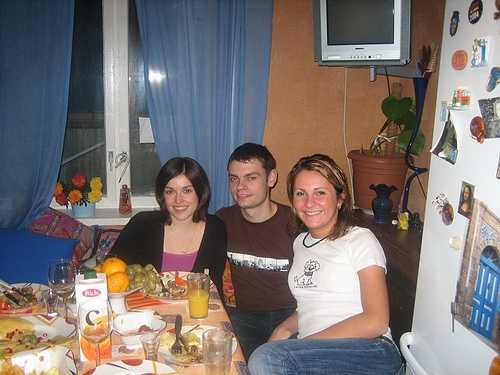}}}{\scriptsize{}\quad{}}\subfloat[\label{fig:grabcut-gt-image-2-1}\negthickspace{}Ground~truth]{\centering{}{\scriptsize{}\includegraphics[width=0.18\textwidth,height=0.08\textheight]{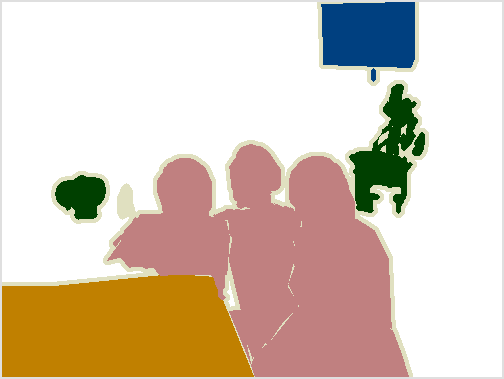}}}{\scriptsize{}\quad{}}\subfloat[\label{fig:grabcut-variant-rectangle-2-1}$\mathtt{Box}$]{\centering{}{\scriptsize{}\includegraphics[width=0.18\textwidth,height=0.08\textheight]{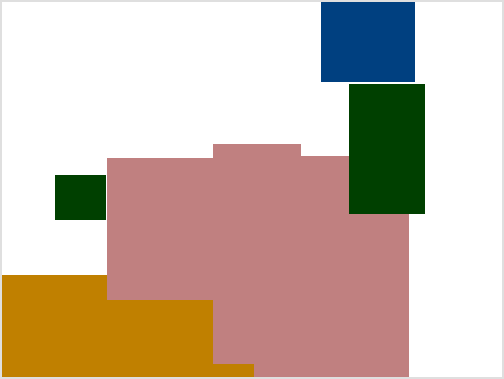}}}{\scriptsize{}\quad{}}\subfloat[\label{fig:grabcut-variant-densecut-1-2-1}$\mathtt{Box^{i}}$]{\centering{}{\scriptsize{}\includegraphics[width=0.18\textwidth,height=0.08\textheight]{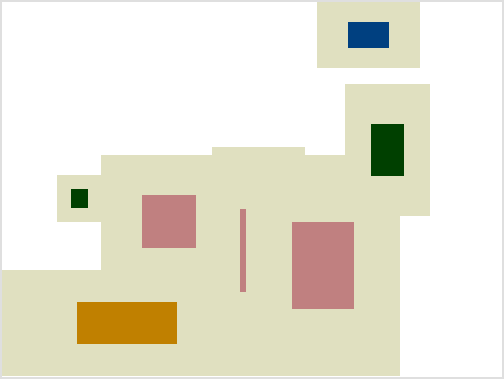}}}{\scriptsize{}\hspace*{\fill}}{\scriptsize \par}

{\scriptsize{}\vspace{-1em}
}{\scriptsize \par}

{\scriptsize{}\hspace*{\fill}}\subfloat[\label{fig:grabcut-variant-kernel-grabcut-1-2-1}\negthickspace{}\negthickspace{}{\scriptsize{}Bbox-Seg+CRF}\negthickspace{}\negthickspace{}]{\centering{}{\scriptsize{}\includegraphics[width=0.18\textwidth,height=0.08\textheight]{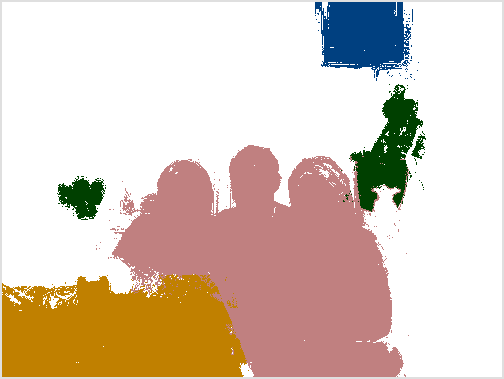}}}{\scriptsize{}\quad{}}\subfloat[\label{fig:grabcut-variant-MCG-2-1}MCG]{\centering{}{\scriptsize{}\includegraphics[width=0.18\textwidth,height=0.08\textheight]{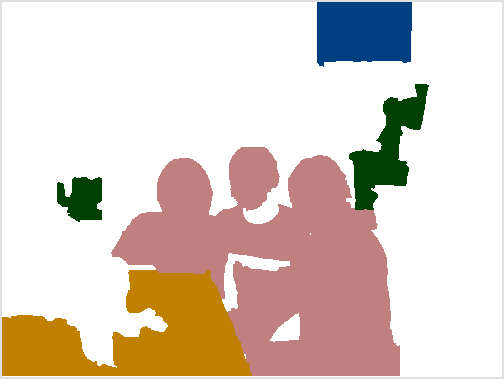}}}{\scriptsize{}\quad{}}\subfloat[\label{fig:grabcut-variant-densecut-3-1}DenseCut]{\centering{}{\scriptsize{}\includegraphics[width=0.18\textwidth,height=0.08\textheight]{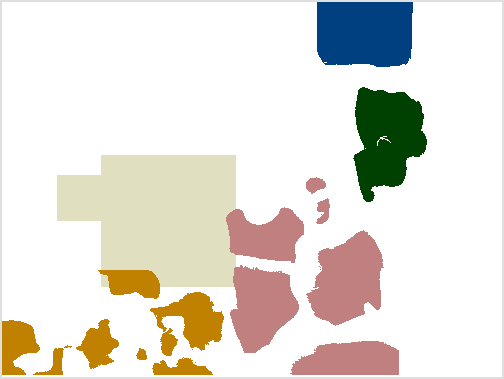}}}{\scriptsize{}\quad{}}\subfloat[\label{fig:grabcut-variant-grabcut-2-1}GrabCut]{\centering{}{\scriptsize{}\includegraphics[width=0.18\textwidth,height=0.08\textheight]{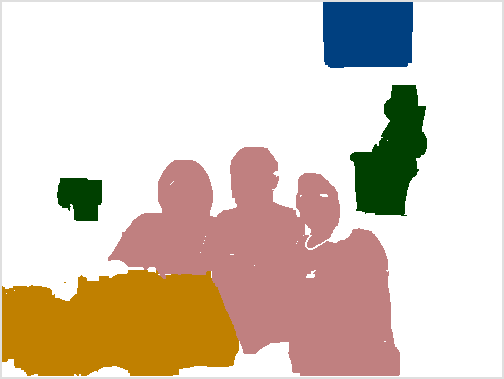}}}{\scriptsize{}\hspace*{\fill}}{\scriptsize \par}

{\scriptsize{}\vspace{-1em}
}{\scriptsize \par}

{\scriptsize{}\hspace*{\fill}}\subfloat[\label{fig:grabcut-variant-kernel-grabcut-3-1}KGrabCut]{\centering{}{\scriptsize{}\includegraphics[width=0.18\textwidth,height=0.08\textheight]{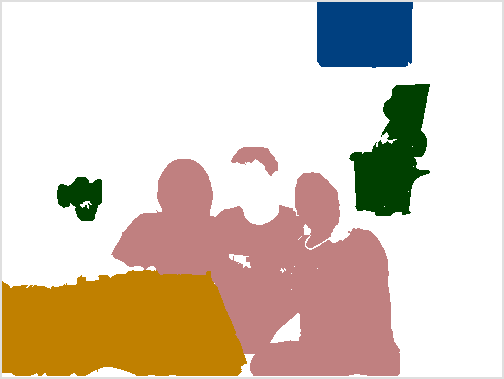}}}{\scriptsize{}\quad{}}\subfloat[\label{fig:grabcut-variant-grabcut-with-strong-edges-2-1}GrabCut+]{\centering{}{\scriptsize{}\includegraphics[width=0.18\textwidth,height=0.08\textheight]{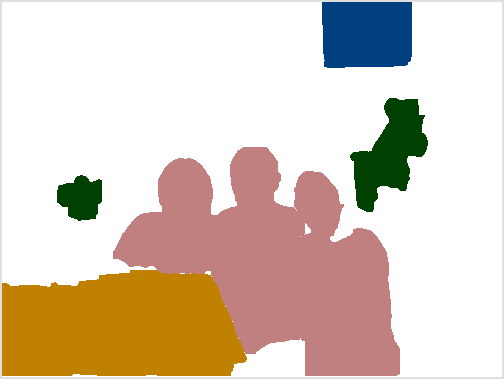}}}{\scriptsize{}\quad{}}\subfloat[\label{fig:grabcut-variant-grabcut+-perturbed-2-1}$\mathtt{GrabCut+^{i}}$]{\centering{}{\scriptsize{}\includegraphics[width=0.18\textwidth,height=0.08\textheight]{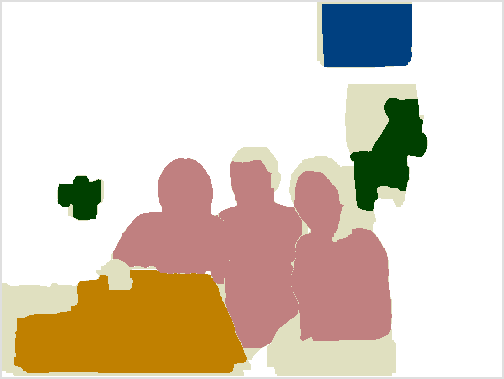}}}{\scriptsize{}\quad{}}\subfloat[\label{fig:grabcut-variant-M=000026G-2-1}$\mbox{M}\,\cap\,\mbox{G+}$]{\centering{}{\scriptsize{}\includegraphics[width=0.18\textwidth,height=0.08\textheight]{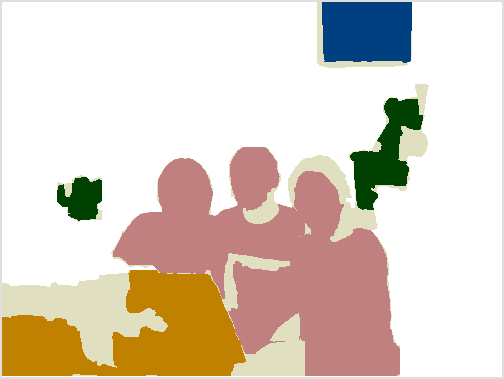}}}{\scriptsize{}\hspace*{\fill}}{\scriptsize \par}

\vspace{0.5em}

\hspace*{\fill}\subfloat[\label{fig:grabcut-input-image-1-1}Input image]{\centering{}\includegraphics[width=0.18\textwidth,height=0.08\textheight]{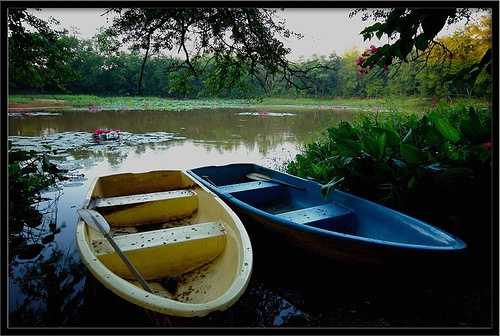}}\quad{}\subfloat[\label{fig:grabcut-gt-image-1-1}\negthickspace{}Ground~truth]{\centering{}\includegraphics[width=0.18\textwidth,height=0.08\textheight]{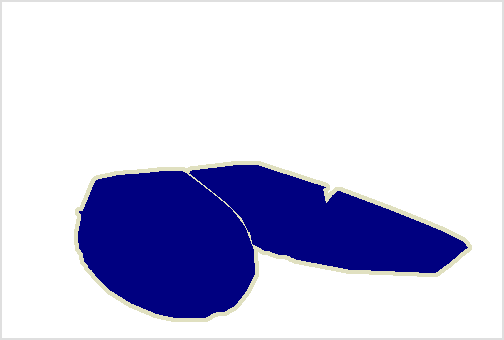}}\quad{}\subfloat[\label{fig:grabcut-variant-rectangle-1-1}$\mathtt{Box}$]{\centering{}\includegraphics[width=0.18\textwidth,height=0.08\textheight]{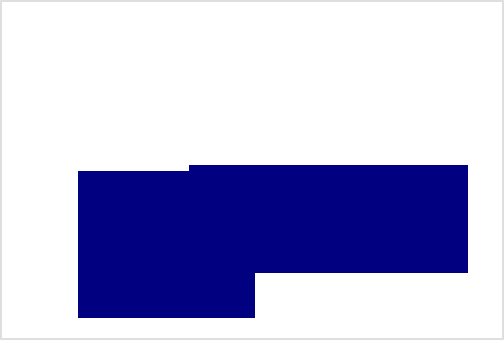}}\quad{}\subfloat[\label{fig:grabcut-variant-densecut-1-1-1}$\mathtt{Box^{i}}$]{\centering{}\includegraphics[width=0.18\textwidth,height=0.08\textheight]{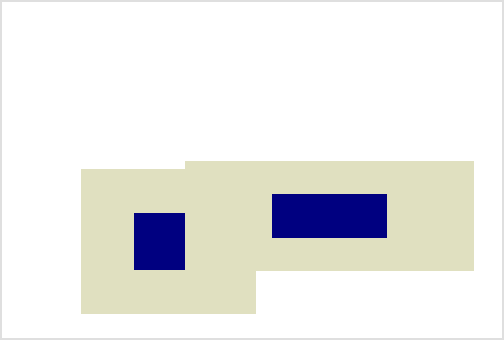}}\hspace*{\fill}

\vspace{-1em}

\hspace*{\fill}\subfloat[\label{fig:grabcut-variant-kernel-grabcut-1-1-1}\negthickspace{}Bbox-Seg+CRF\negthickspace{}]{\centering{}\includegraphics[width=0.18\textwidth,height=0.08\textheight]{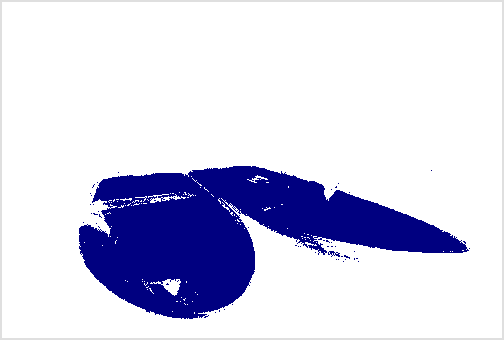}}\quad{}\subfloat[\label{fig:grabcut-variant-MCG-1-1}MCG]{\centering{}\includegraphics[width=0.18\textwidth,height=0.08\textheight]{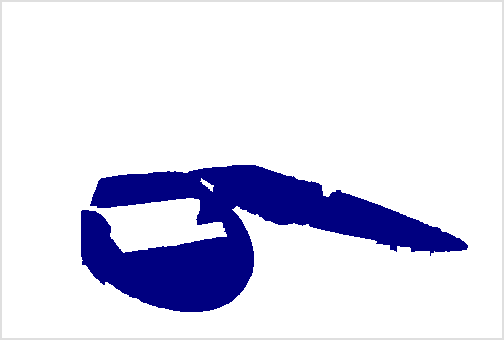}}\quad{}\subfloat[\label{fig:grabcut-variant-densecut-2-1}DenseCut]{\centering{}\includegraphics[width=0.18\textwidth,height=0.08\textheight]{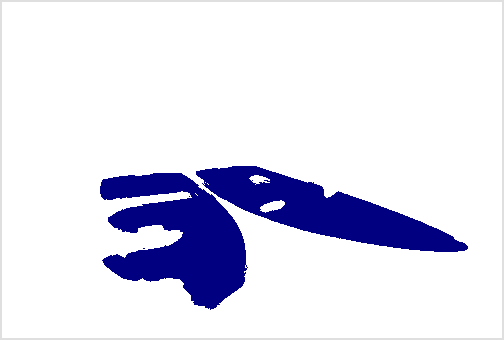}}\quad{}\subfloat[\label{fig:grabcut-variant-grabcut-1-1}GrabCut]{\centering{}\includegraphics[width=0.18\textwidth,height=0.08\textheight]{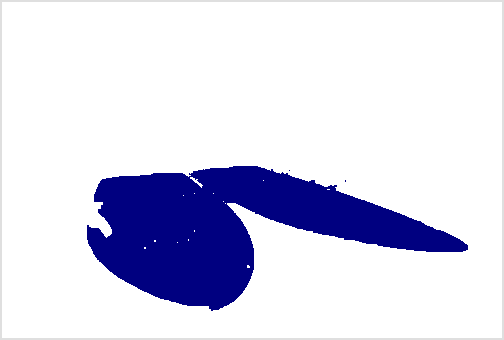}}\hspace*{\fill}

\vspace{-1em}

\hspace*{\fill}\subfloat[\label{fig:grabcut-variant-kernel-grabcut-2-1}KGrabCut]{\centering{}\includegraphics[width=0.18\textwidth,height=0.08\textheight]{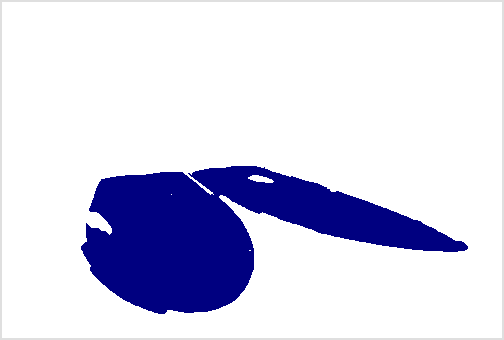}}\quad{}\subfloat[\label{fig:grabcut-variant-grabcut-with-strong-edges-1-1}GrabCut+]{\centering{}\includegraphics[width=0.18\textwidth,height=0.08\textheight]{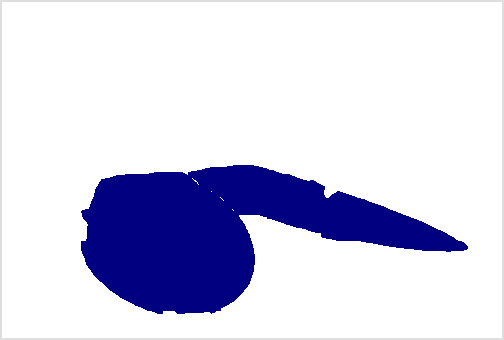}}\quad{}\subfloat[\label{fig:grabcut-variant-grabcut+-perturbed-1-1}$\mathtt{GrabCut+^{i}}$]{\centering{}\includegraphics[width=0.18\textwidth,height=0.08\textheight]{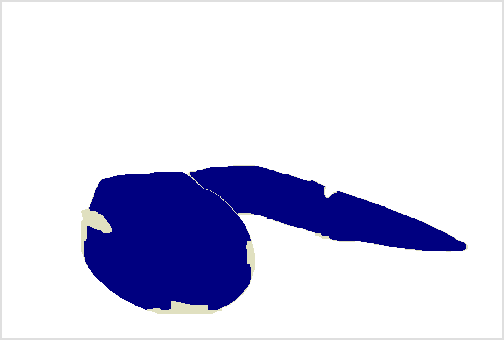}}\quad{}\subfloat[\label{fig:grabcut-variant-M=000026G-1-1}$\mbox{M}\,\cap\,\mbox{G+}$]{\centering{}\includegraphics[width=0.18\textwidth,height=0.08\textheight]{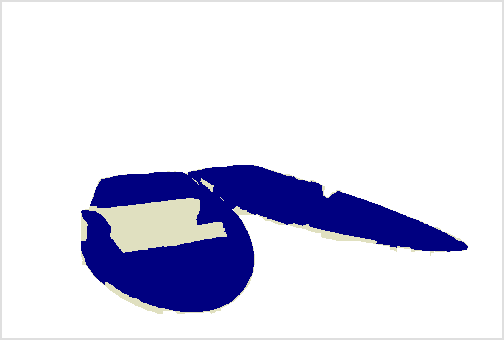}}\hspace*{\fill}

\vspace{0.5em}

\hspace*{\fill}\subfloat[\label{fig:grabcut-input-image-1-1-1}Input image]{\centering{}\includegraphics[width=0.18\textwidth,height=0.08\textheight]{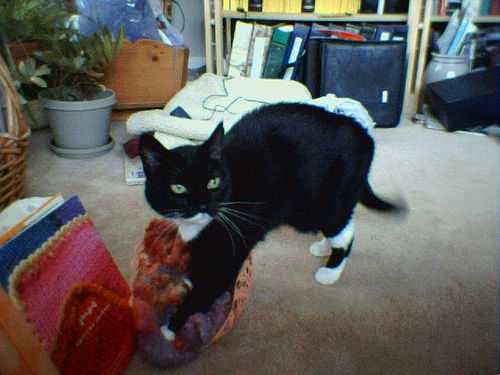}}\quad{}\subfloat[\label{fig:grabcut-gt-image-1-1-1}\negthickspace{}Ground~truth]{\centering{}\includegraphics[width=0.18\textwidth,height=0.08\textheight]{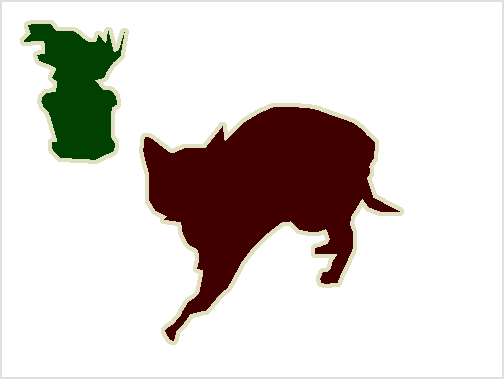}}\quad{}\subfloat[\label{fig:grabcut-variant-rectangle-1-1-1}$\mathtt{Box}$]{\centering{}\includegraphics[width=0.18\textwidth,height=0.08\textheight]{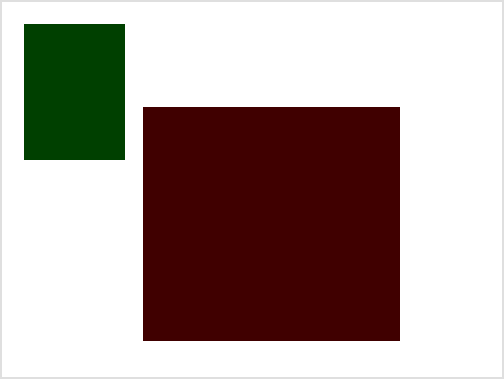}}\quad{}\subfloat[\label{fig:grabcut-variant-densecut-1-1-1-1}$\mathtt{Box^{i}}$]{\centering{}\includegraphics[width=0.18\textwidth,height=0.08\textheight]{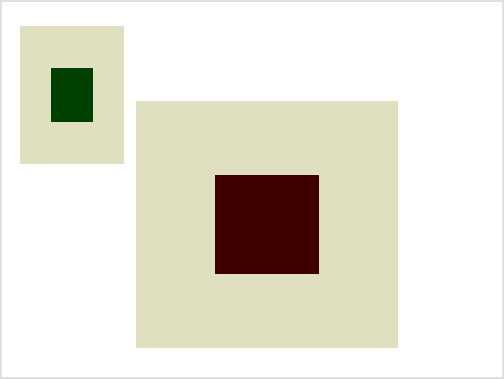}}\hspace*{\fill}

\vspace{-1em}

\hspace*{\fill}\subfloat[\label{fig:grabcut-variant-kernel-grabcut-1-1-1-1}\negthickspace{}Bbox-Seg+CRF\negthickspace{}]{\centering{}\includegraphics[width=0.18\textwidth,height=0.08\textheight]{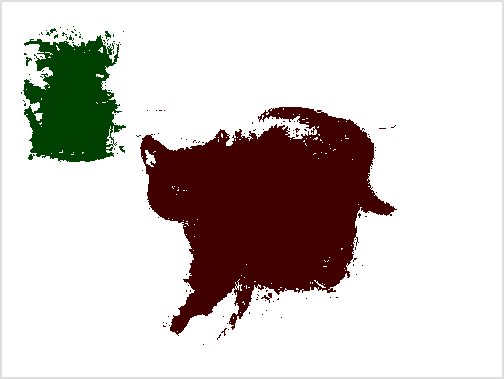}}\quad{}\subfloat[\label{fig:grabcut-variant-MCG-1-1-1}MCG]{\centering{}\includegraphics[width=0.18\textwidth,height=0.08\textheight]{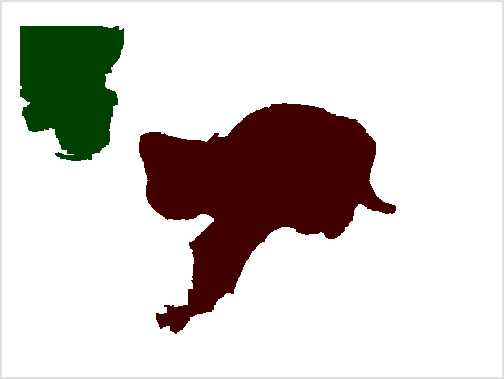}}\quad{}\subfloat[\label{fig:grabcut-variant-densecut-2-1-1}DenseCut]{\centering{}\includegraphics[width=0.18\textwidth,height=0.08\textheight]{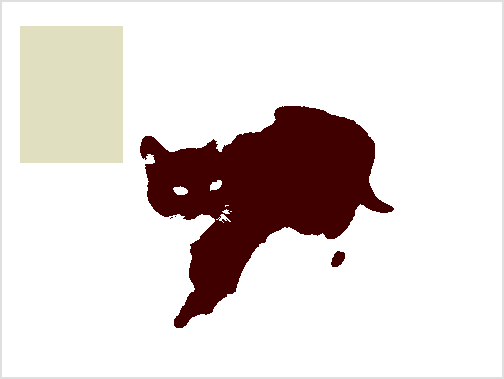}}\quad{}\subfloat[\label{fig:grabcut-variant-grabcut-1-1-1}GrabCut]{\centering{}\includegraphics[width=0.18\textwidth,height=0.08\textheight]{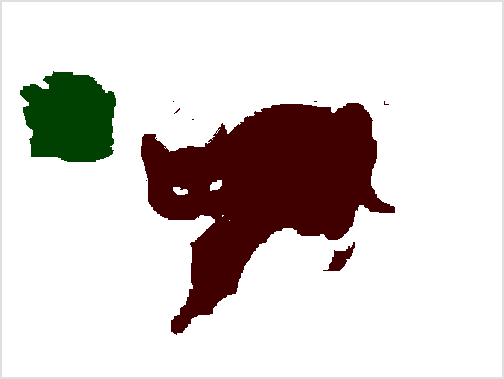}}\hspace*{\fill}

\vspace{-1em}

\hspace*{\fill}\subfloat[\label{fig:grabcut-variant-kernel-grabcut-2-1-1}KGrabCut]{\centering{}\includegraphics[width=0.18\textwidth,height=0.08\textheight]{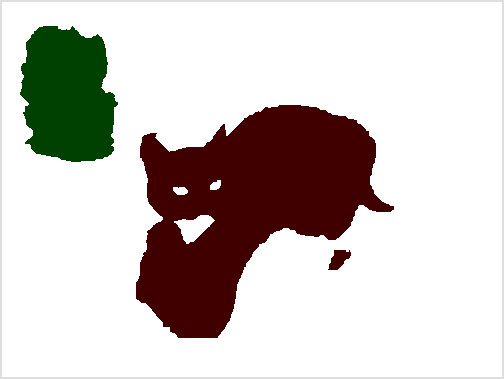}}\quad{}\subfloat[\label{fig:grabcut-variant-grabcut-with-strong-edges-1-1-1}GrabCut+]{\centering{}\includegraphics[width=0.18\textwidth,height=0.08\textheight]{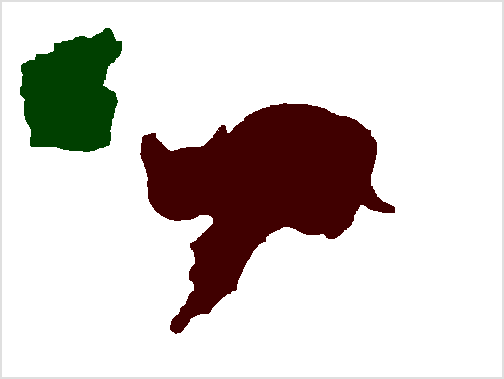}}\quad{}\subfloat[\label{fig:grabcut-variant-grabcut+-perturbed-1-1-1}$\mathtt{GrabCut+^{i}}$]{\centering{}\includegraphics[width=0.18\textwidth,height=0.08\textheight]{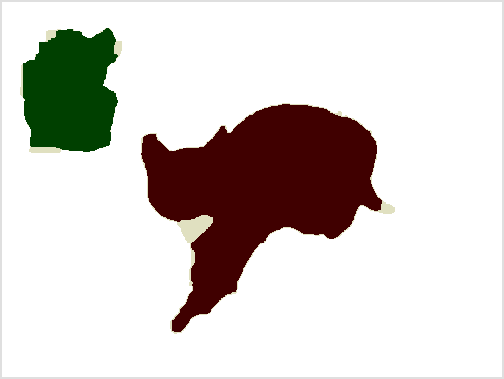}}\quad{}\subfloat[\label{fig:grabcut-variant-M=000026G-1-1-1}$\mbox{M}\,\cap\,\mbox{G+}$]{\centering{}\includegraphics[width=0.18\textwidth,height=0.08\textheight]{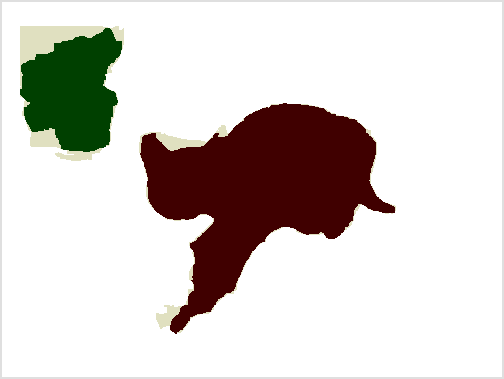}}\hspace*{\fill}

{\scriptsize{}\vspace{-0.5em}
}{\scriptsize \par}

\caption{\label{fig:grabcut-variants-examples-2-1}Different segmentations
obtained starting from a bounding box. White is background and ignore
regions are beige. $\mbox{M}\,\cap\,\mbox{G+}$ denotes $\mbox{MCG}\,\cap\,\mbox{Grabcut+}$.}
\vspace{-1em}
\end{figure*}

\begin{figure*}[t]
\begingroup{
\setlength{\tabcolsep}{0pt} 
\renewcommand{\arraystretch}{0.2}\vspace{-1em}

\hspace*{\fill}%
\begin{tabular}[b]{ccccccccccccc}
\includegraphics[width=0.14\textwidth,height=0.07\textheight]{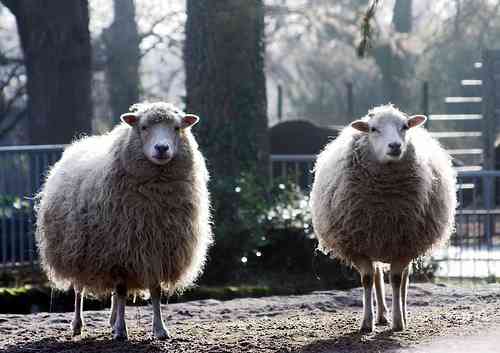} & \hspace*{0.1em} & \includegraphics[width=0.14\textwidth,height=0.07\textheight]{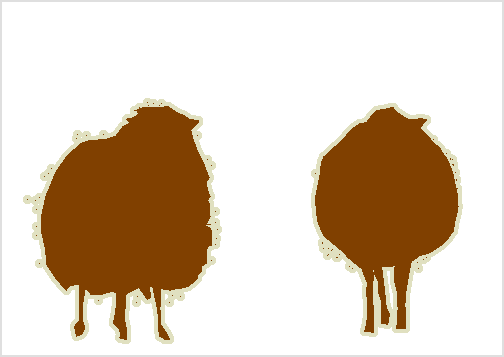} & \hspace*{0.1em} & \includegraphics[width=0.14\textwidth,height=0.07\textheight]{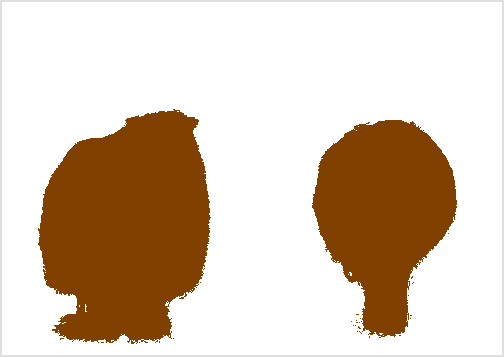} & \hspace*{0.1em} & \includegraphics[width=0.14\textwidth,height=0.07\textheight]{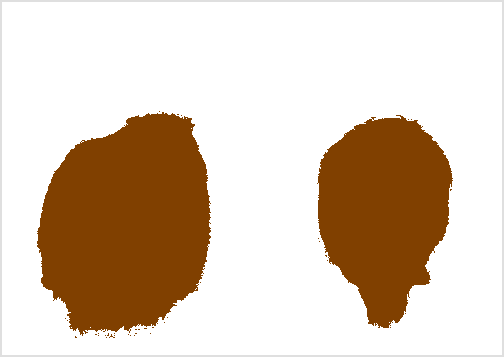} & \hspace*{0.1em} & \includegraphics[width=0.14\textwidth,height=0.07\textheight]{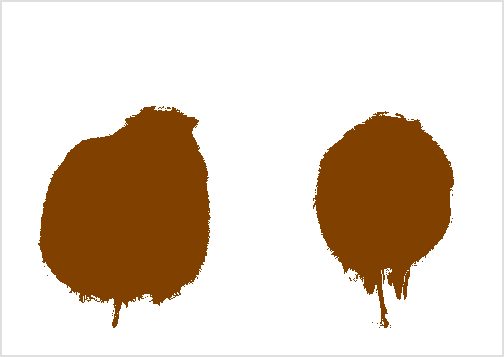} & \hspace*{0.1em} & \includegraphics[width=0.14\textwidth,height=0.07\textheight]{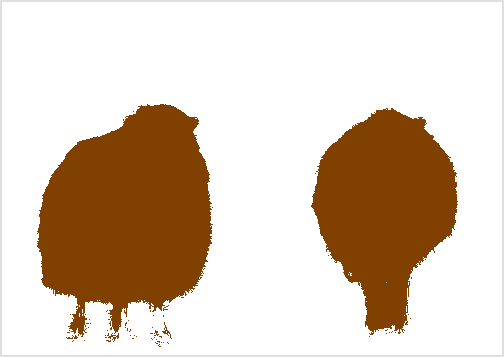} & \hspace*{0.1em} & \includegraphics[width=0.14\textwidth,height=0.07\textheight]{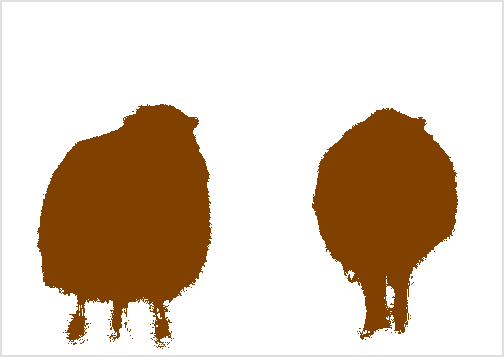}\tabularnewline
\includegraphics[width=0.14\textwidth,height=0.12\textheight]{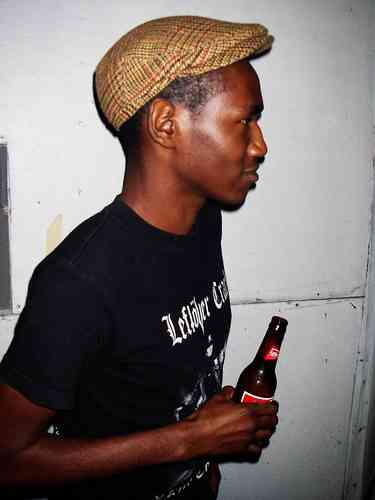} & \hspace*{0.1em} & \includegraphics[width=0.14\textwidth,height=0.12\textheight]{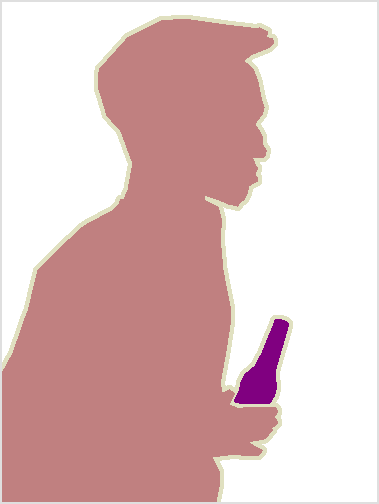} & \hspace*{0.1em} & \includegraphics[width=0.14\textwidth,height=0.12\textheight]{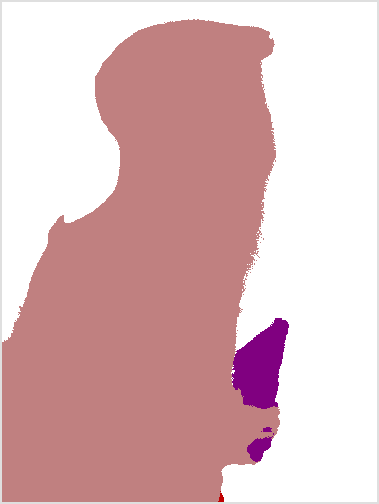} & \hspace*{0.1em} & \includegraphics[width=0.14\textwidth,height=0.12\textheight]{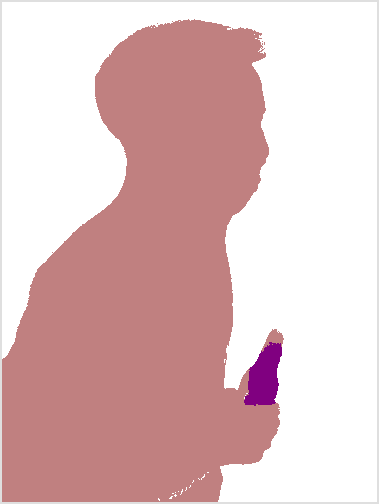} & \hspace*{0.1em} & \includegraphics[width=0.14\textwidth,height=0.12\textheight]{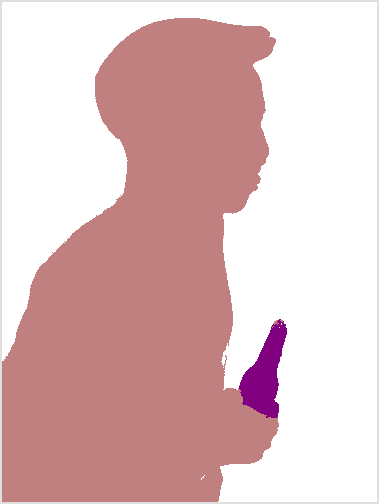} & \hspace*{0.1em} & \includegraphics[width=0.14\textwidth,height=0.12\textheight]{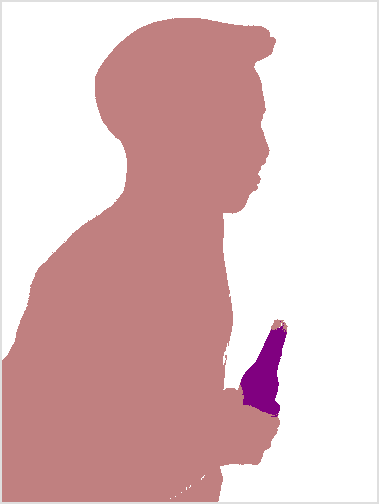} & \hspace*{0.1em} & \includegraphics[width=0.14\textwidth,height=0.12\textheight]{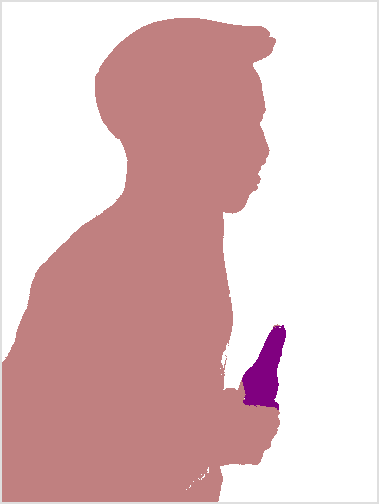}\tabularnewline
\includegraphics[width=0.14\textwidth]{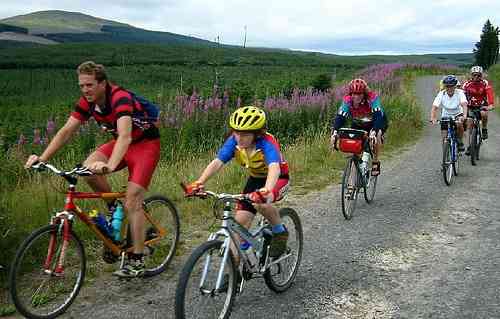} & \hspace*{0.1em} & \includegraphics[width=0.14\textwidth]{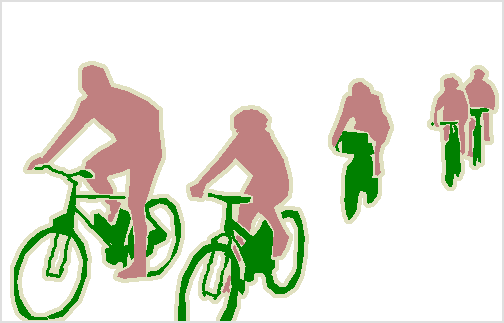} & \hspace*{0.1em} & \includegraphics[width=0.14\textwidth]{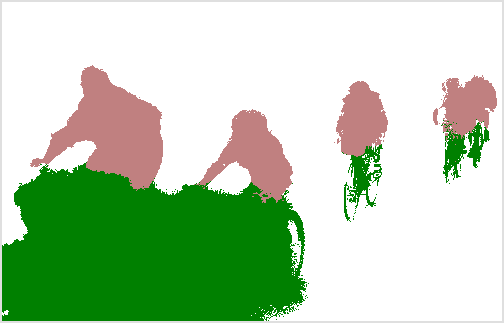} & \hspace*{0.1em} & \includegraphics[width=0.14\textwidth]{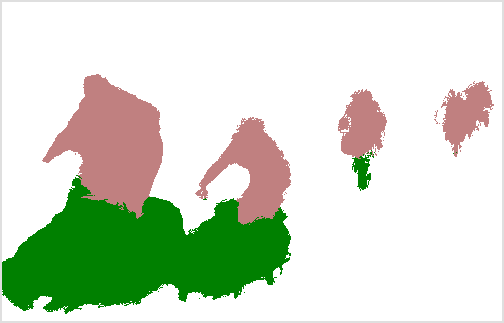} & \hspace*{0.1em} & \includegraphics[width=0.14\textwidth]{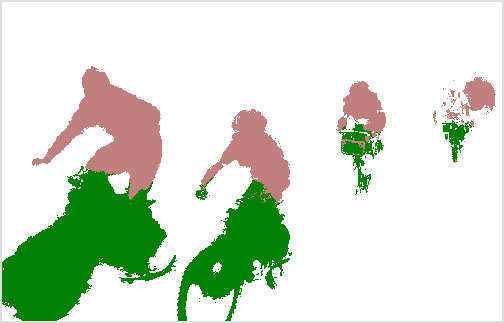} & \hspace*{0.1em} & \includegraphics[width=0.14\textwidth]{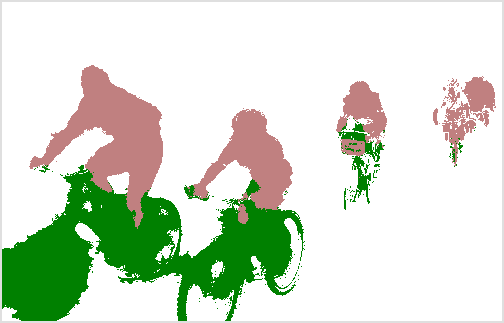} & \hspace*{0.1em} & \includegraphics[width=0.14\textwidth]{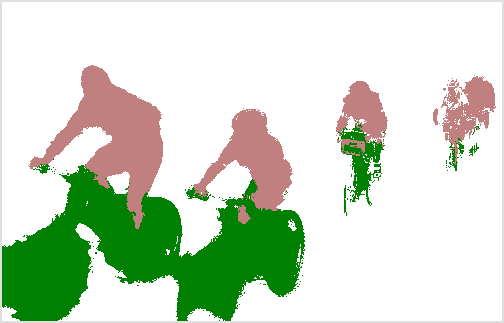}\tabularnewline
\includegraphics[width=0.14\textwidth]{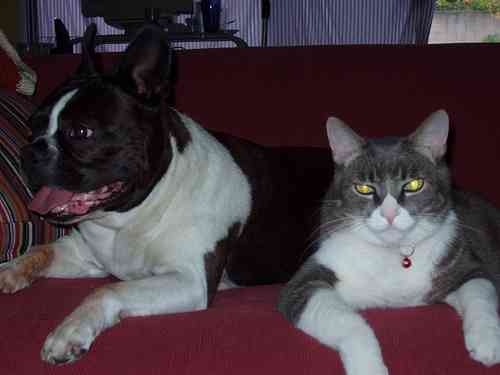} & \hspace*{0.1em} & \includegraphics[width=0.14\textwidth]{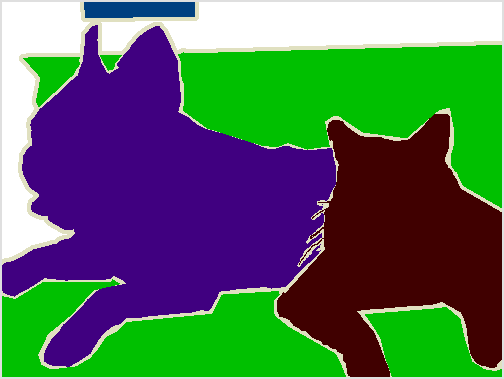} & \hspace*{0.1em} & \includegraphics[width=0.14\textwidth]{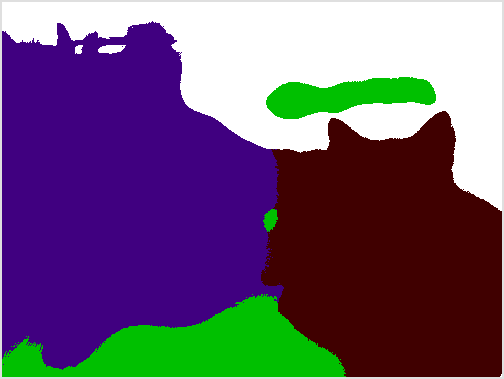} & \hspace*{0.1em} & \includegraphics[width=0.14\textwidth]{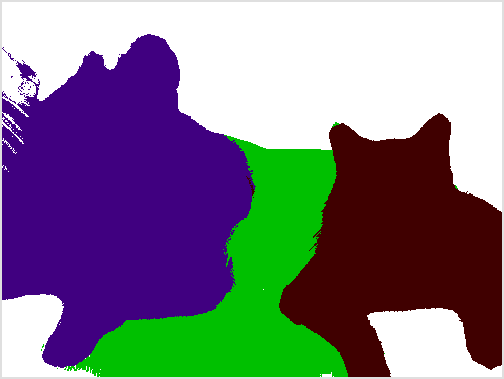} & \hspace*{0.1em} & \includegraphics[width=0.14\textwidth]{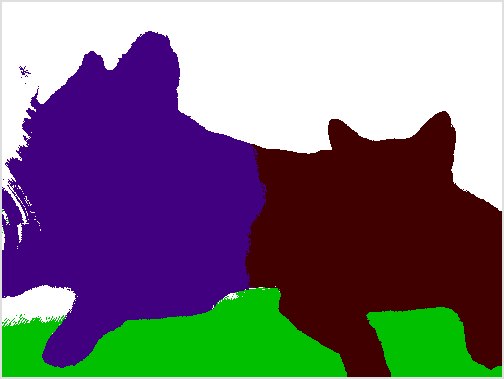} & \hspace*{0.1em} & \includegraphics[width=0.14\textwidth]{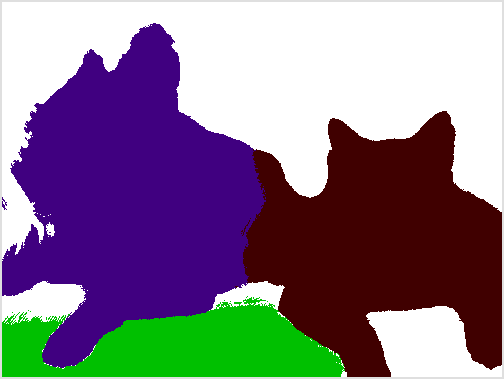} & \hspace*{0.1em} & \includegraphics[width=0.14\textwidth]{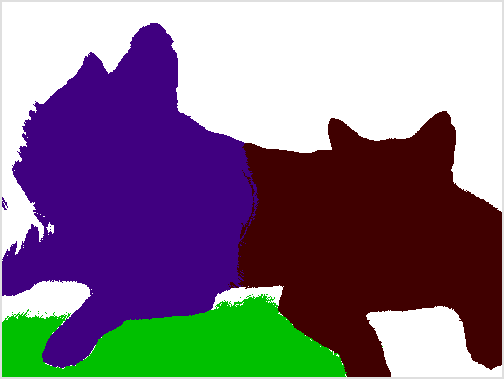}\tabularnewline
\includegraphics[width=0.14\textwidth]{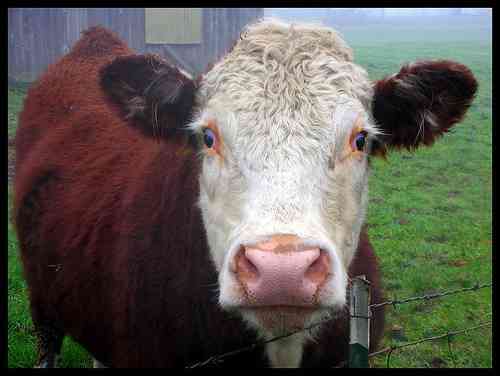} & \hspace*{0.1em} & \includegraphics[width=0.14\textwidth]{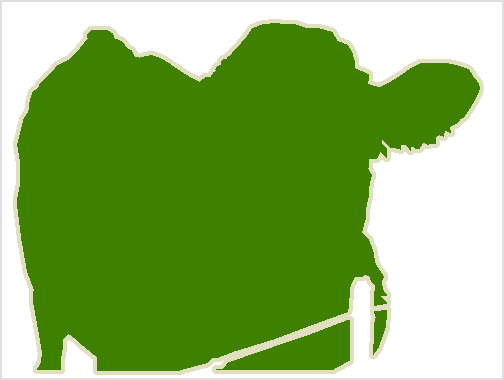} & \hspace*{0.1em} & \includegraphics[width=0.14\textwidth]{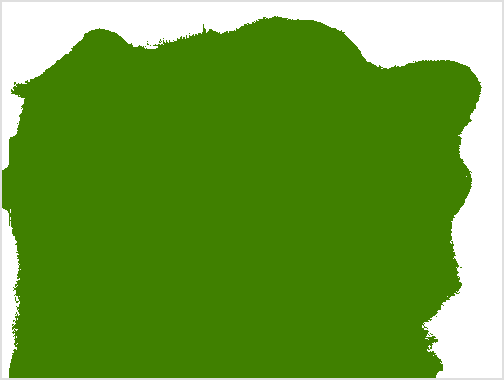} & \hspace*{0.1em} & \includegraphics[width=0.14\textwidth]{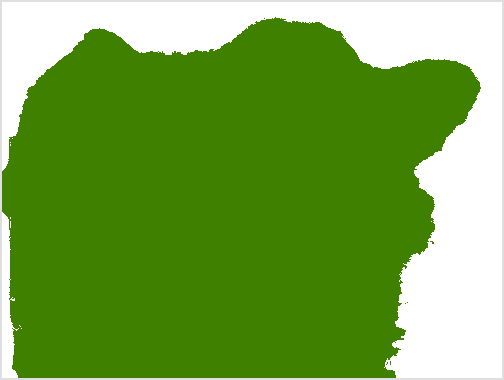} & \hspace*{0.1em} & \includegraphics[width=0.14\textwidth]{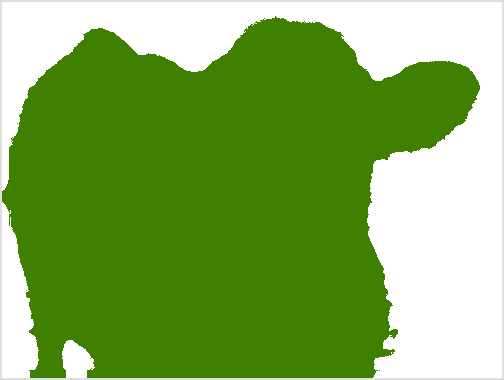} & \hspace*{0.1em} & \includegraphics[width=0.14\textwidth]{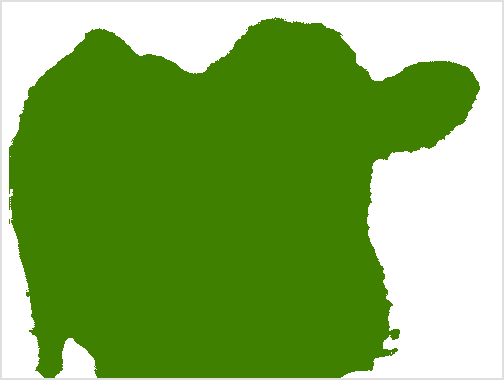} & \hspace*{0.1em} & \includegraphics[width=0.14\textwidth]{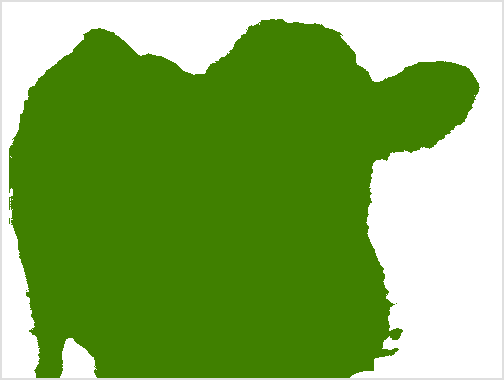}\tabularnewline
\includegraphics[width=0.14\textwidth]{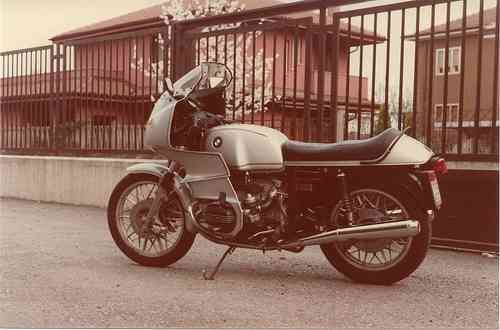} & \hspace*{0.1em} & \includegraphics[width=0.14\textwidth]{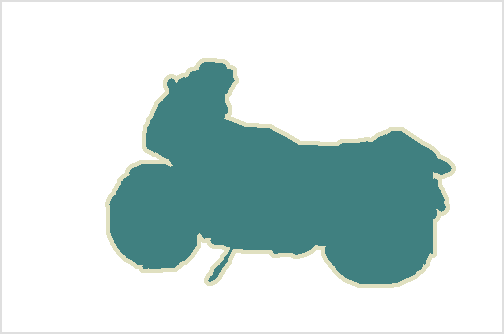} & \hspace*{0.1em} & \includegraphics[width=0.14\textwidth]{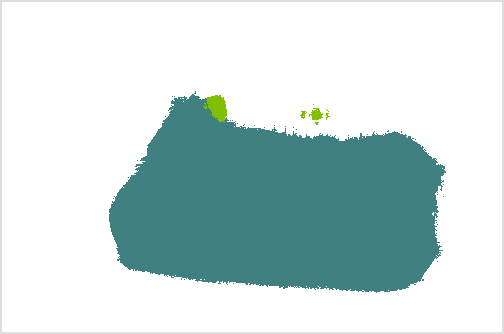} & \hspace*{0.1em} & \includegraphics[width=0.14\textwidth]{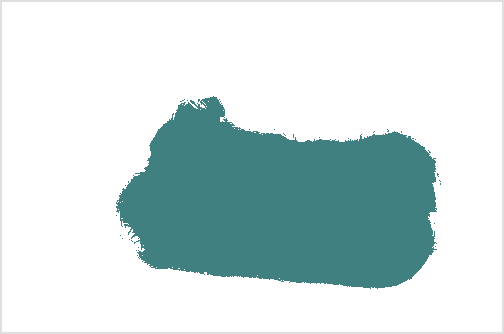} & \hspace*{0.1em} & \includegraphics[width=0.14\textwidth]{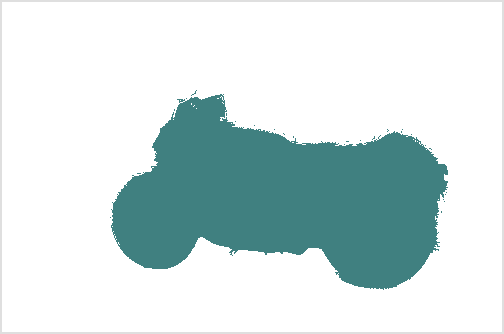} & \hspace*{0.1em} & \includegraphics[width=0.14\textwidth]{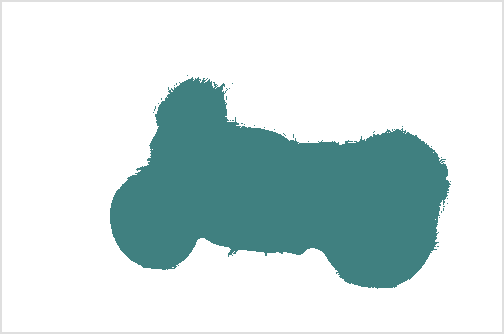} & \hspace*{0.1em} & \includegraphics[width=0.14\textwidth]{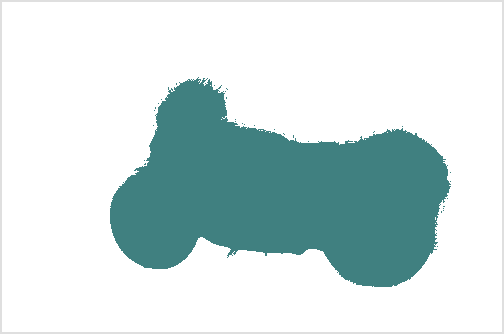}\tabularnewline
\includegraphics[width=0.14\textwidth]{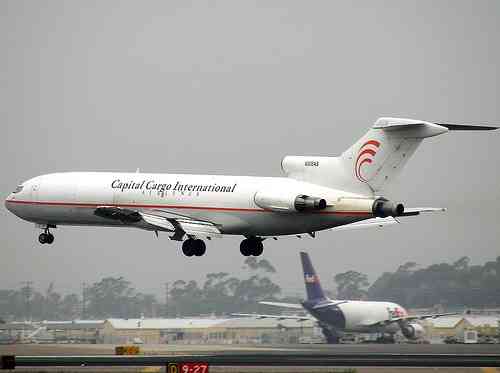} & \hspace*{0.1em} & \includegraphics[width=0.14\textwidth]{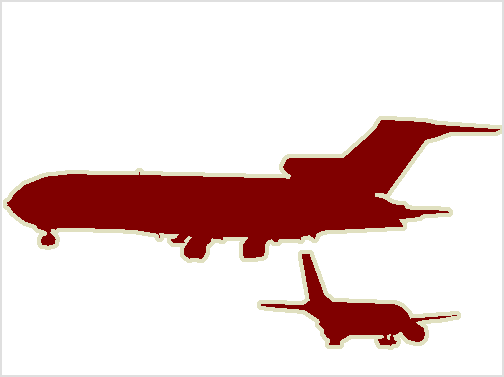} & \hspace*{0.1em} & \includegraphics[width=0.14\textwidth]{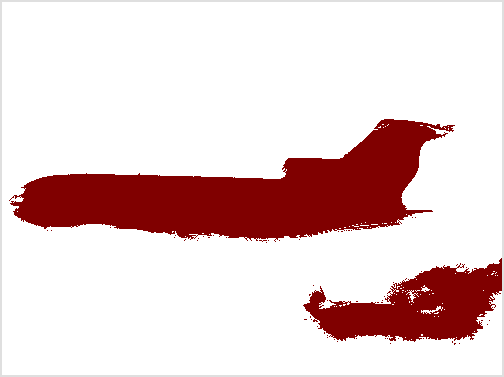} & \hspace*{0.1em} & \includegraphics[width=0.14\textwidth]{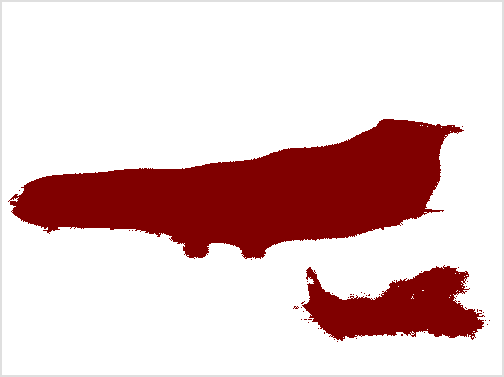} & \hspace*{0.1em} & \includegraphics[width=0.14\textwidth]{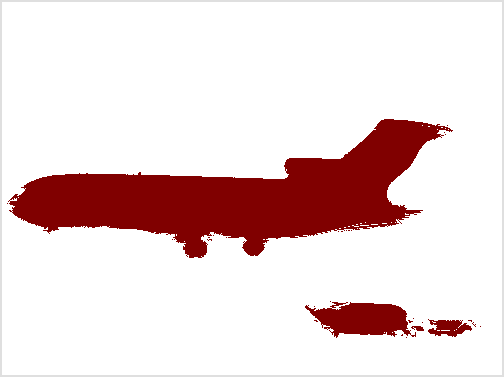} & \hspace*{0.1em} & \includegraphics[width=0.14\textwidth]{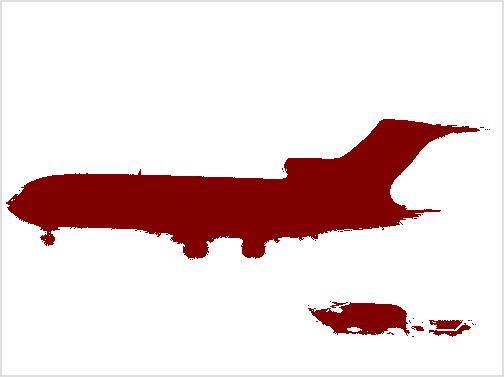} & \hspace*{0.1em} & \includegraphics[width=0.14\textwidth]{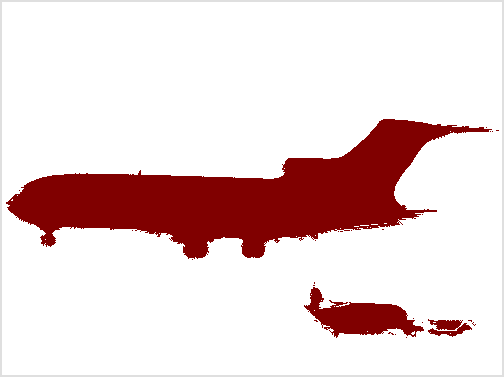}\tabularnewline
\includegraphics[width=0.14\textwidth]{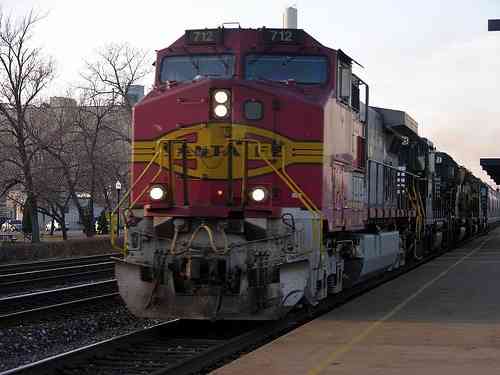} & \hspace*{0.1em} & \includegraphics[width=0.14\textwidth]{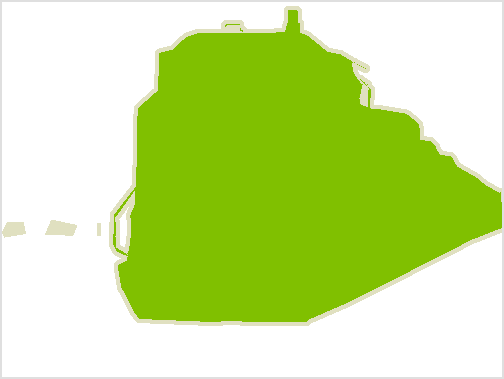} & \hspace*{0.1em} & \includegraphics[width=0.14\textwidth]{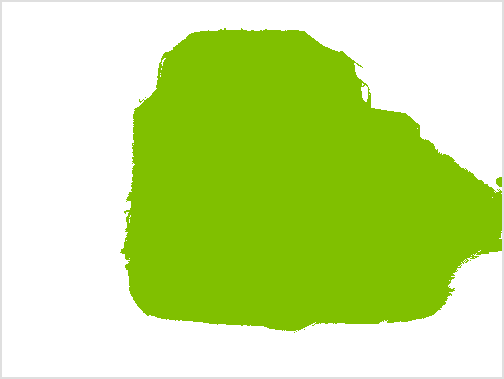} & \hspace*{0.1em} & \includegraphics[width=0.14\textwidth]{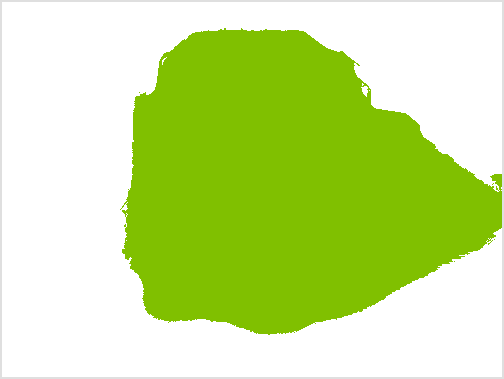} & \hspace*{0.1em} & \includegraphics[width=0.14\textwidth]{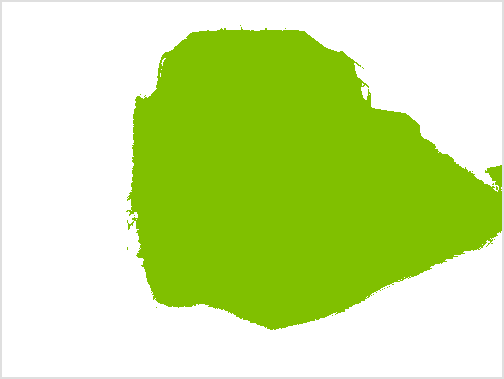} & \hspace*{0.1em} & \includegraphics[width=0.14\textwidth]{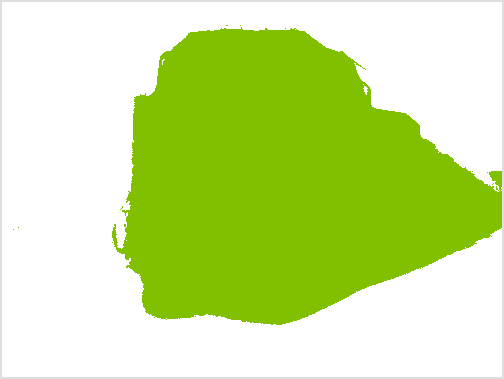} & \hspace*{0.1em} & \includegraphics[width=0.14\textwidth]{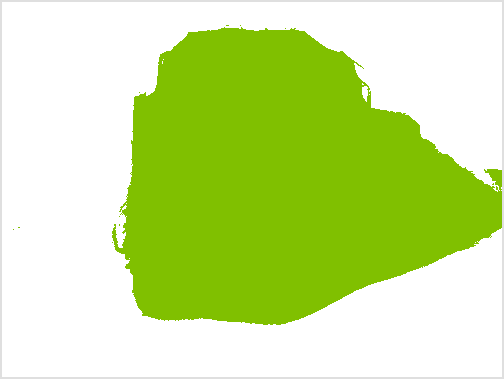}\tabularnewline
\includegraphics[width=0.14\textwidth]{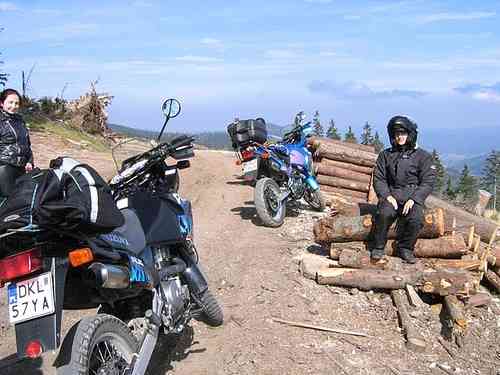} & \hspace*{0.1em} & \includegraphics[width=0.14\textwidth]{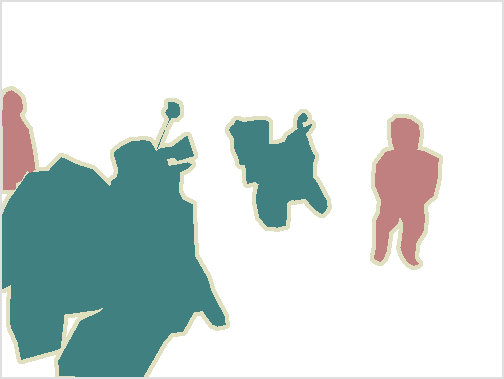} & \hspace*{0.1em} & \includegraphics[width=0.14\textwidth]{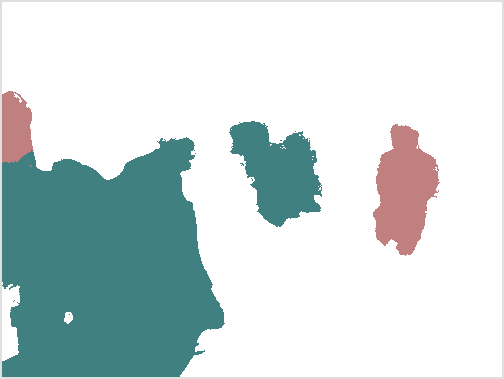} & \hspace*{0.1em} & \includegraphics[width=0.14\textwidth]{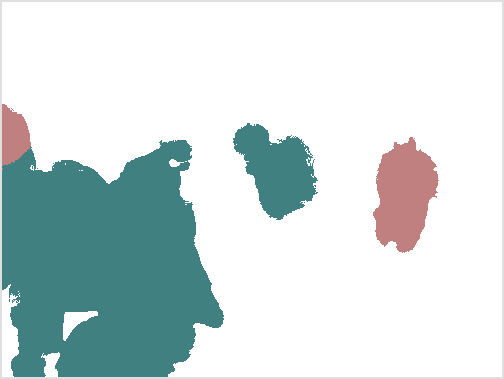} & \hspace*{0.1em} & \includegraphics[width=0.14\textwidth]{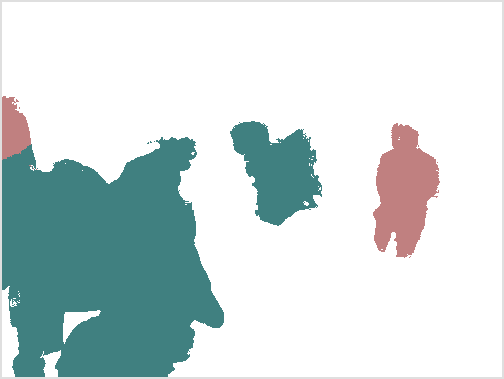} & \hspace*{0.1em} & \includegraphics[width=0.14\textwidth]{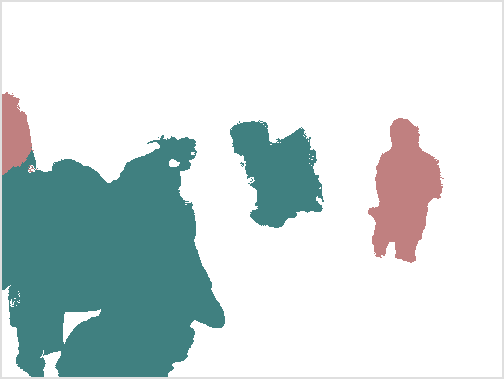} & \hspace*{0.1em} & \includegraphics[width=0.14\textwidth]{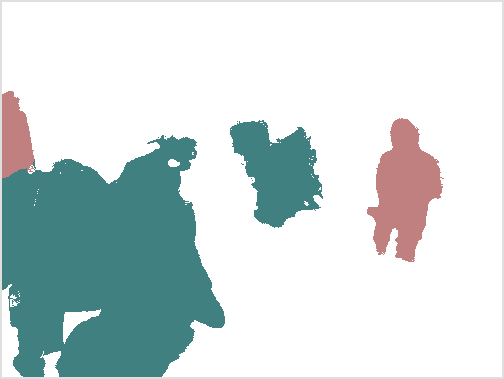}\tabularnewline
\includegraphics[width=0.14\textwidth,height=0.07\textheight]{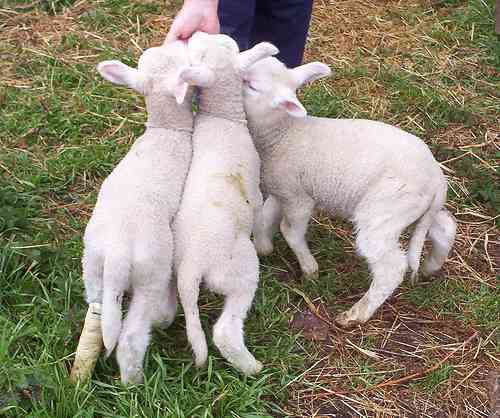} & \hspace*{0.1em} & \includegraphics[width=0.14\textwidth,height=0.07\textheight]{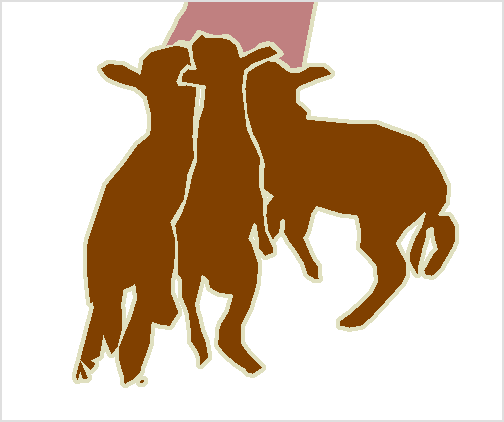} & \hspace*{0.1em} & \includegraphics[width=0.14\textwidth,height=0.07\textheight]{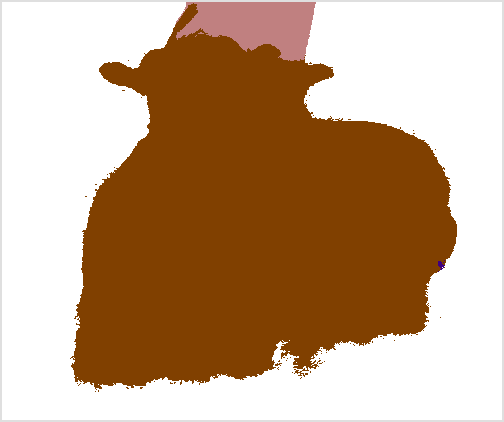} & \hspace*{0.1em} & \includegraphics[width=0.14\textwidth,height=0.07\textheight]{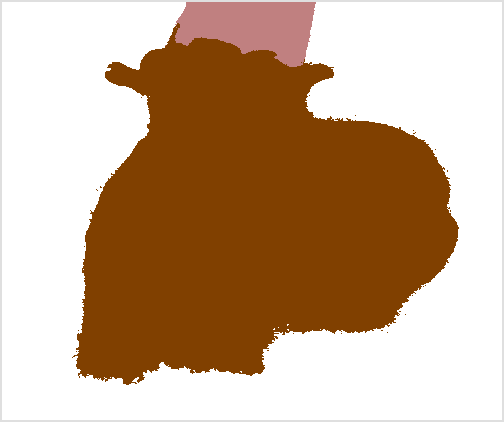} & \hspace*{0.1em} & \includegraphics[width=0.14\textwidth,height=0.07\textheight]{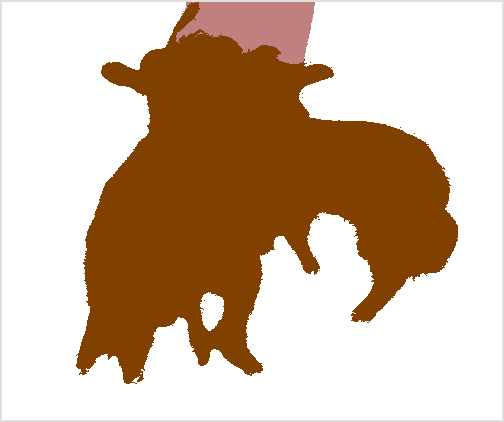} & \hspace*{0.1em} & \includegraphics[width=0.14\textwidth,height=0.07\textheight]{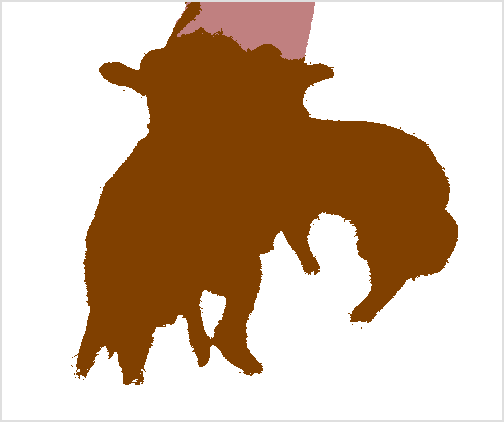} & \hspace*{0.1em} & \includegraphics[width=0.14\textwidth,height=0.07\textheight]{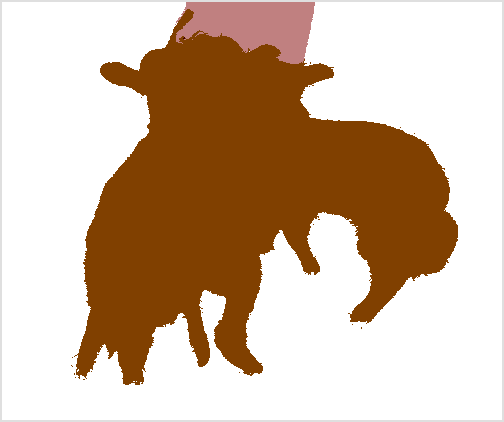}\tabularnewline
\includegraphics[width=0.14\textwidth]{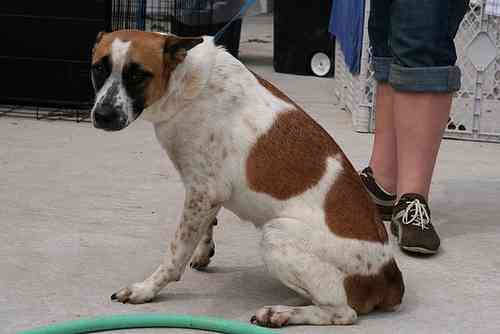} & \hspace*{0.1em} & \includegraphics[width=0.14\textwidth]{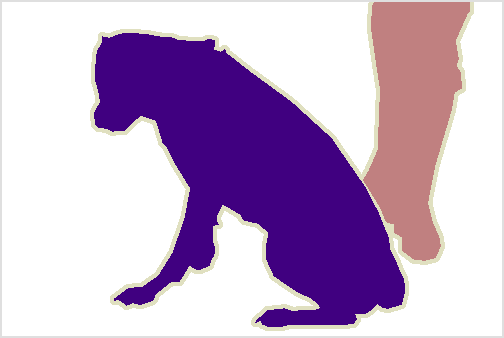} & \hspace*{0.1em} & \includegraphics[width=0.14\textwidth]{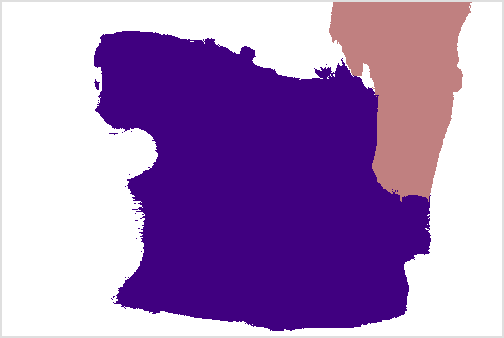} & \hspace*{0.1em} & \includegraphics[width=0.14\textwidth]{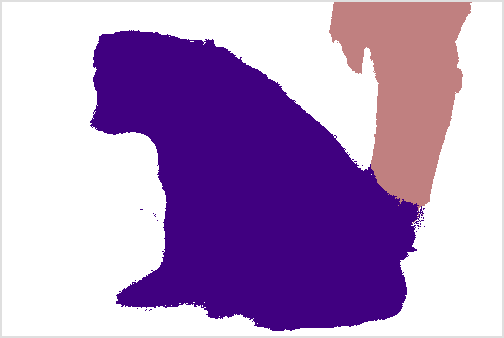} & \hspace*{0.1em} & \includegraphics[width=0.14\textwidth]{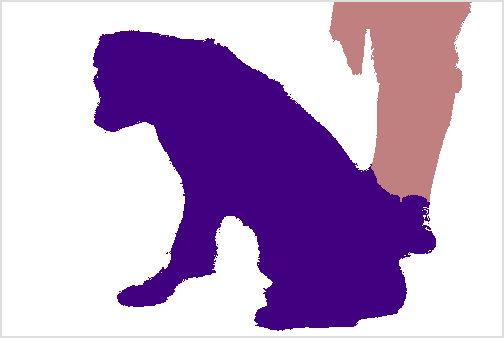} & \hspace*{0.1em} & \includegraphics[width=0.14\textwidth]{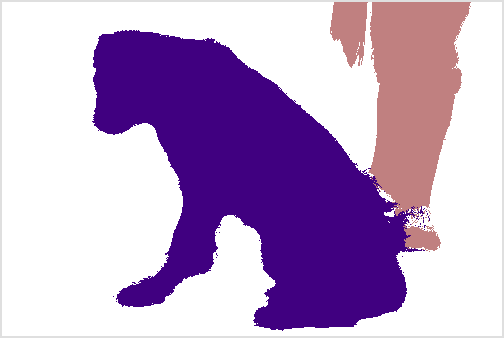} & \hspace*{0.1em} & \includegraphics[width=0.14\textwidth]{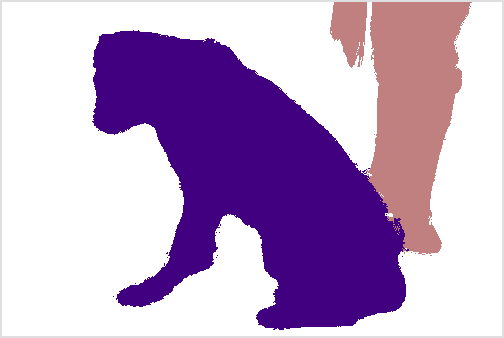}\tabularnewline
\includegraphics[width=0.14\textwidth]{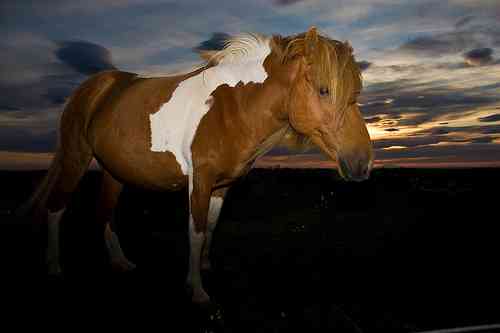} & \hspace*{0.1em} & \includegraphics[width=0.14\textwidth]{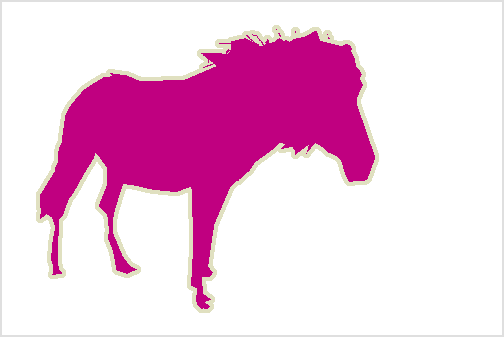} & \hspace*{0.1em} & \includegraphics[width=0.14\textwidth]{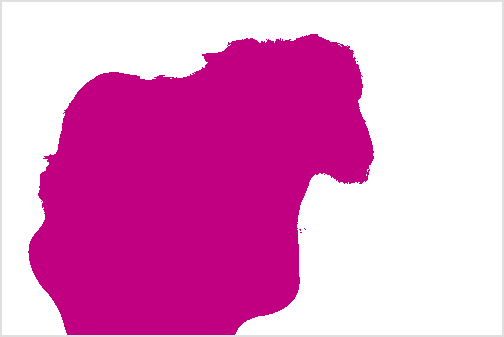} & \hspace*{0.1em} & \includegraphics[width=0.14\textwidth]{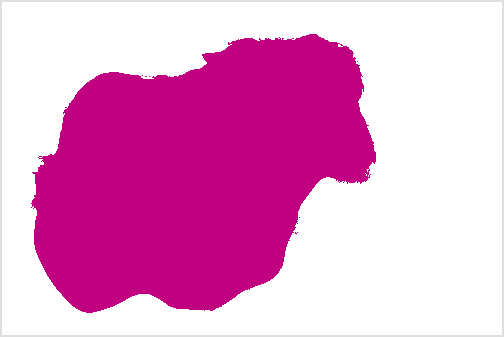} & \hspace*{0.1em} & \includegraphics[width=0.14\textwidth]{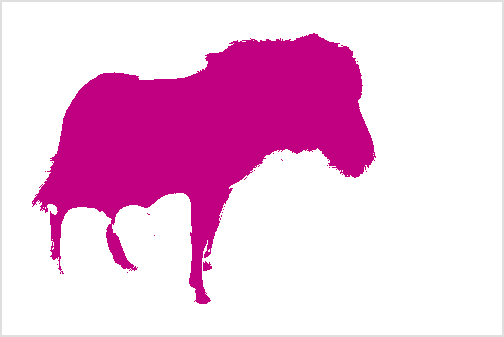} & \hspace*{0.1em} & \includegraphics[width=0.14\textwidth]{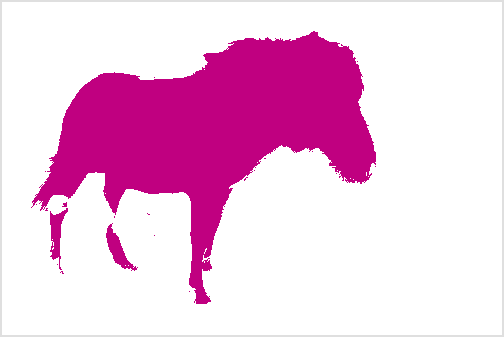} & \hspace*{0.1em} & \includegraphics[width=0.14\textwidth]{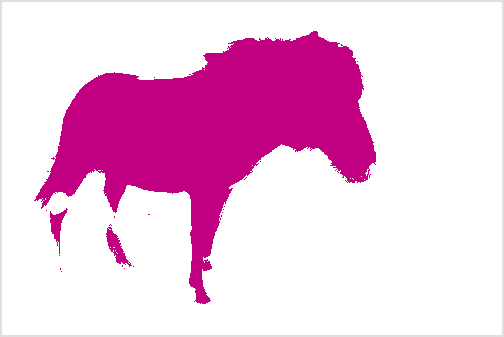}\tabularnewline
Image &  & %
\begin{tabular}{c}
Ground\tabularnewline
truth\tabularnewline
\end{tabular}  &  & $\mathtt{Box}$ &  & $\mathtt{Box^{i}}$ &  & $\mbox{M}\,\cap\,\mbox{G+}$ &  & %
\begin{tabular}{c}
Semi\tabularnewline
supervised\tabularnewline
$\mbox{M}\,\cap\,\mbox{G+}$\tabularnewline
\end{tabular} &  & %
\begin{tabular}{c}
Fully\tabularnewline
supervised\tabularnewline
\end{tabular}\tabularnewline
\end{tabular}\hspace*{\fill}\vspace{-1em}

\caption{\label{fig:Qualitative-results-1}Qualitative results on VOC12. $\mbox{M}\,\cap\,\mbox{G+}$
denotes the weakly supervised model trained on $\mbox{MCG}\,\cap\,\mbox{Grabcut+}$.}
\vspace{-1em}
}\endgroup
\end{figure*}

\begin{figure*}[t]
\begingroup{
\setlength{\tabcolsep}{0pt} 
\renewcommand{\arraystretch}{0.2}
\captionsetup[subfigure]{labelformat=empty,font=scriptsize}

\hspace*{\fill}%
\begin{tabular}[b]{ccccccccc}
\includegraphics[width=0.19\textwidth,height=0.09\textheight]{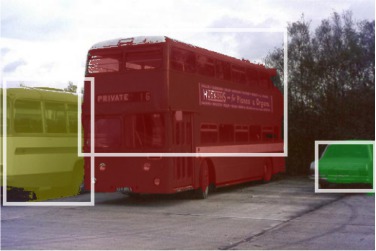} & \hspace*{0.1em} & \includegraphics[width=0.19\textwidth,height=0.09\textheight]{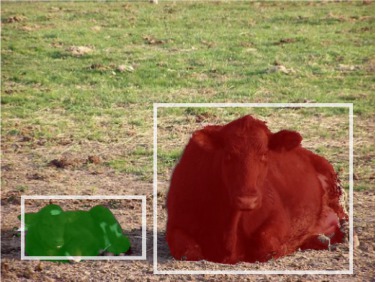} & \hspace*{0.1em} & \includegraphics[width=0.19\textwidth,height=0.09\textheight]{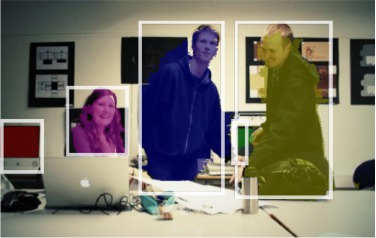} & \hspace*{0.1em} & \includegraphics[width=0.19\textwidth,height=0.09\textheight]{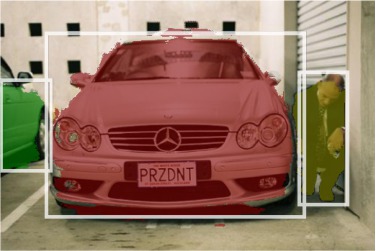} & \hspace*{0.1em} & \includegraphics[width=0.19\textwidth,height=0.09\textheight]{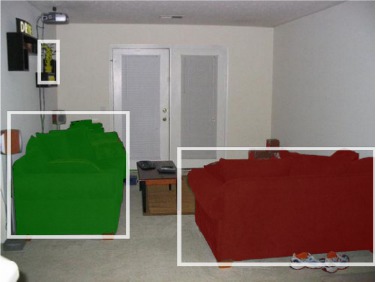}\tabularnewline
\includegraphics[width=0.19\textwidth,height=0.09\textheight]{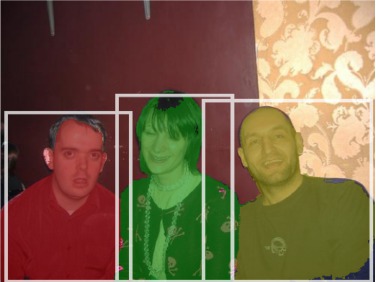} & \hspace*{0.1em} & \includegraphics[width=0.19\textwidth,height=0.09\textheight]{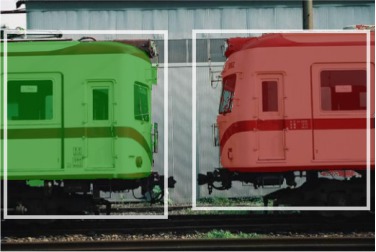} & \hspace*{0.1em} & \includegraphics[width=0.19\textwidth,height=0.09\textheight]{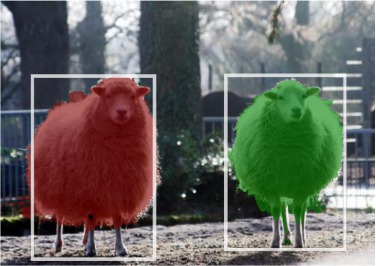} & \hspace*{0.1em} & \includegraphics[width=0.19\textwidth,height=0.09\textheight]{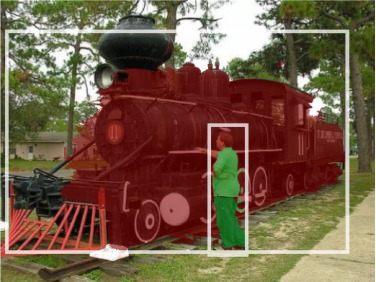} & \hspace*{0.1em} & \includegraphics[width=0.19\textwidth,height=0.09\textheight]{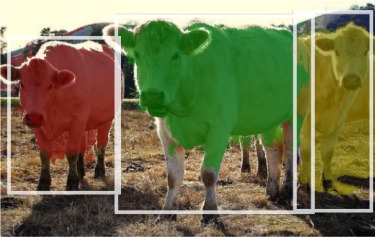}\tabularnewline
\includegraphics[width=0.19\textwidth,height=0.09\textheight]{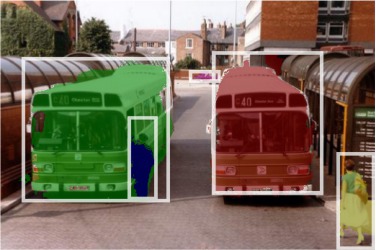} & \hspace*{0.1em} & \includegraphics[width=0.19\textwidth,height=0.09\textheight]{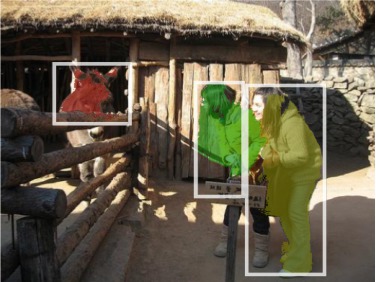} & \hspace*{0.1em} & \includegraphics[width=0.19\textwidth,height=0.09\textheight]{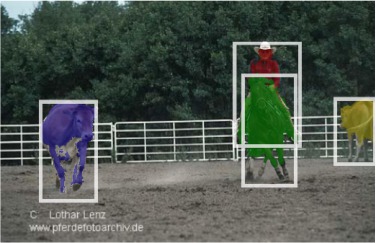} & \hspace*{0.1em} & \includegraphics[width=0.19\textwidth,height=0.09\textheight]{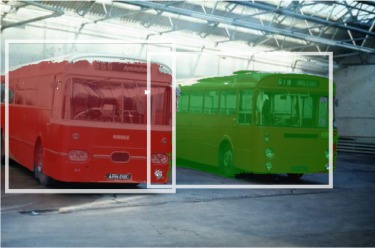} & \hspace*{0.1em} & \includegraphics[width=0.19\textwidth,height=0.09\textheight]{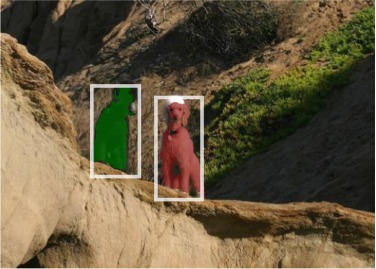}\tabularnewline
\includegraphics[width=0.19\textwidth,height=0.15\textheight]{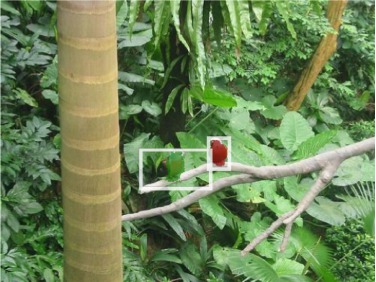} & \hspace*{0.1em} & \includegraphics[width=0.19\textwidth,height=0.15\textheight]{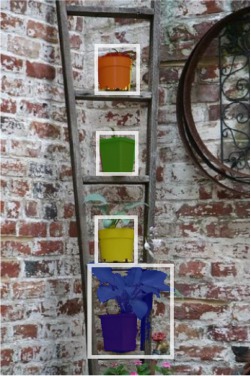} & \hspace*{0.1em} & \includegraphics[width=0.19\textwidth,height=0.15\textheight]{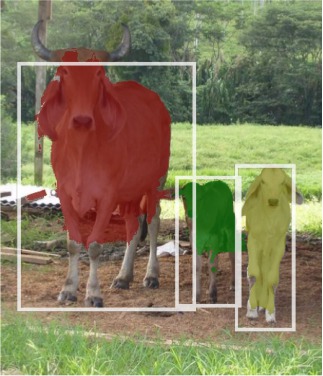} & \hspace*{0.1em} & \includegraphics[width=0.19\textwidth,height=0.15\textheight]{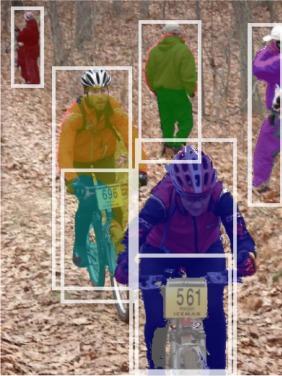} & \hspace*{0.1em} & \includegraphics[width=0.19\textwidth,height=0.15\textheight]{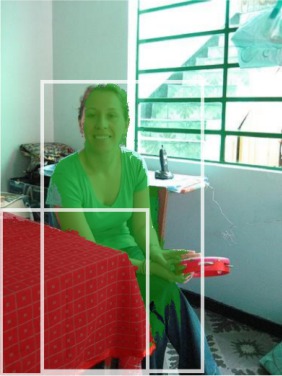}\tabularnewline
\vspace{0.1em}
 &  &  &  &  &  &  &  & \tabularnewline
\multicolumn{9}{c}{$\mathrm{DeepMask}$}\tabularnewline
\vspace{1em}
 &  &  &  &  &  &  &  & \tabularnewline
\includegraphics[width=0.19\textwidth,height=0.09\textheight]{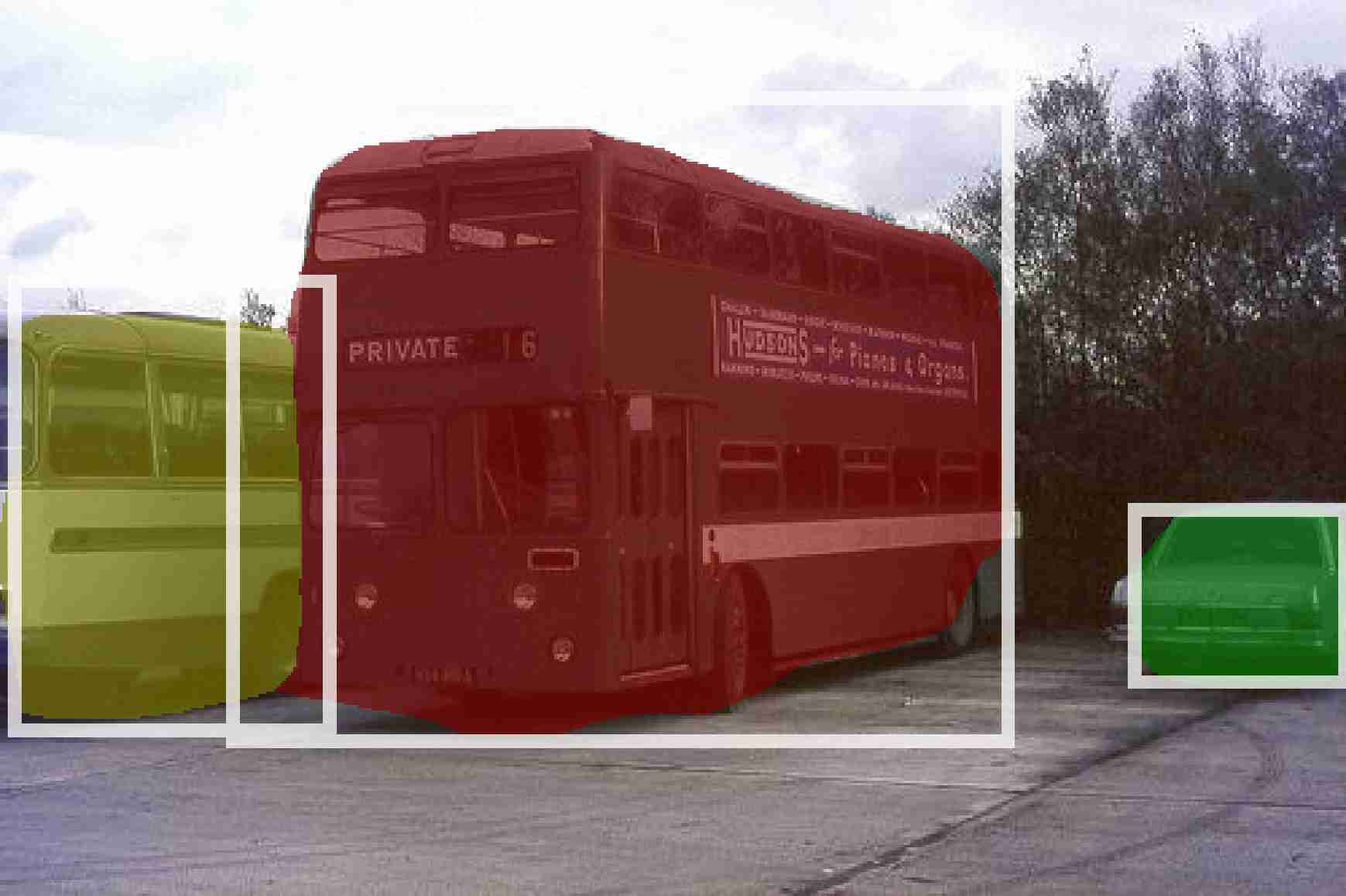} & \hspace*{0.1em} & \includegraphics[width=0.19\textwidth,height=0.09\textheight]{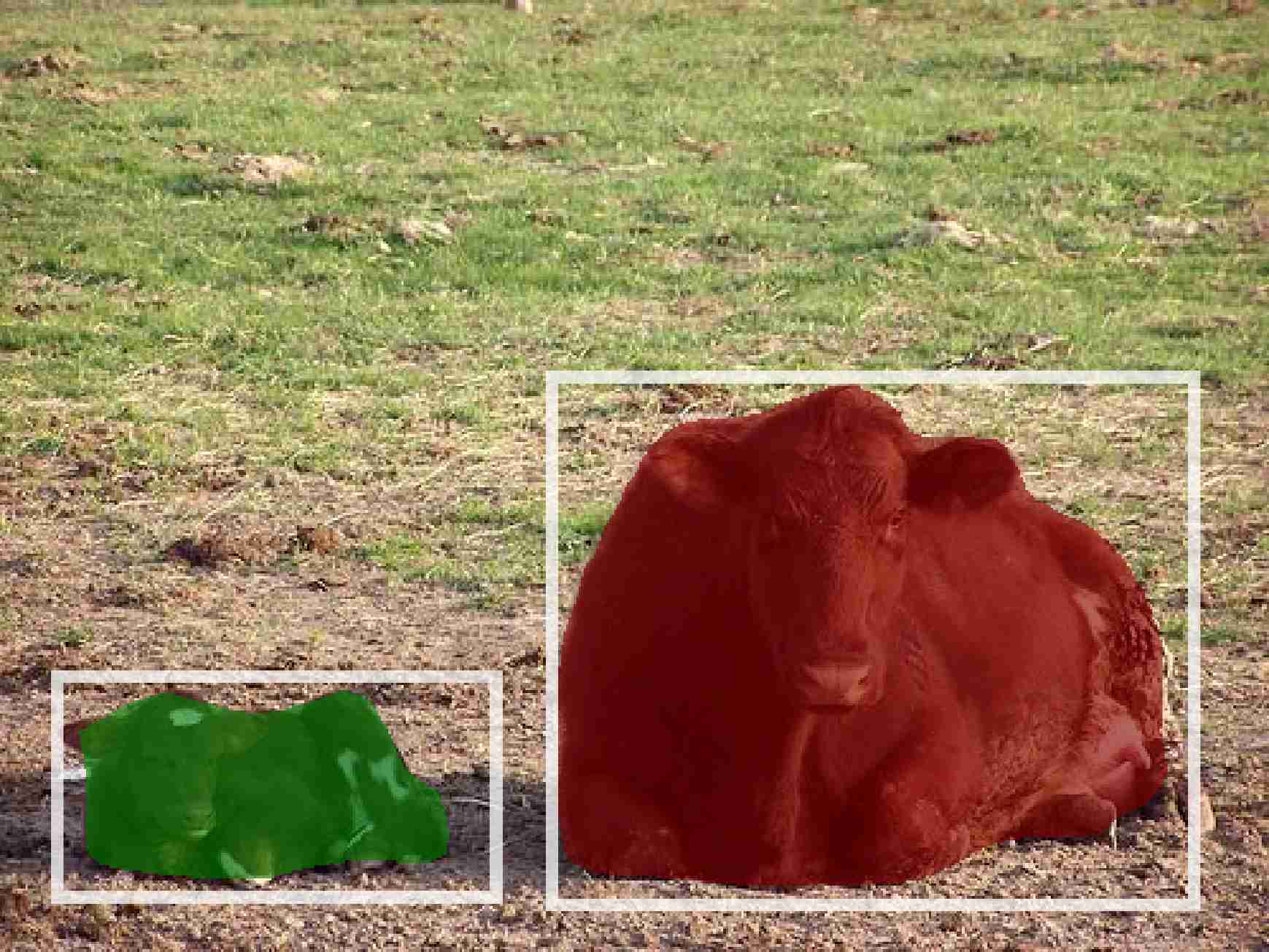} & \hspace*{0.1em} & \includegraphics[width=0.19\textwidth,height=0.09\textheight]{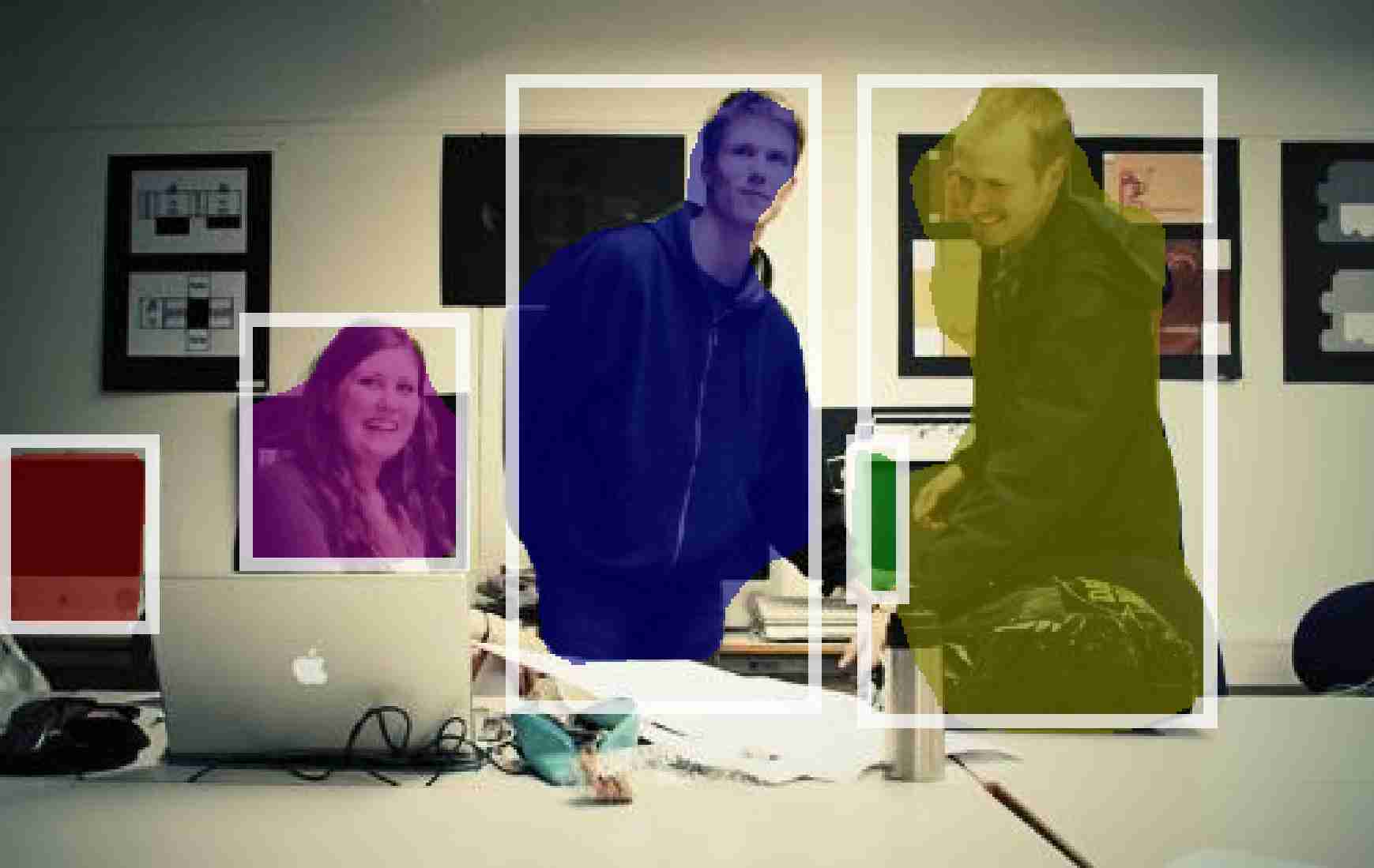} & \hspace*{0.1em} & \includegraphics[width=0.19\textwidth,height=0.09\textheight]{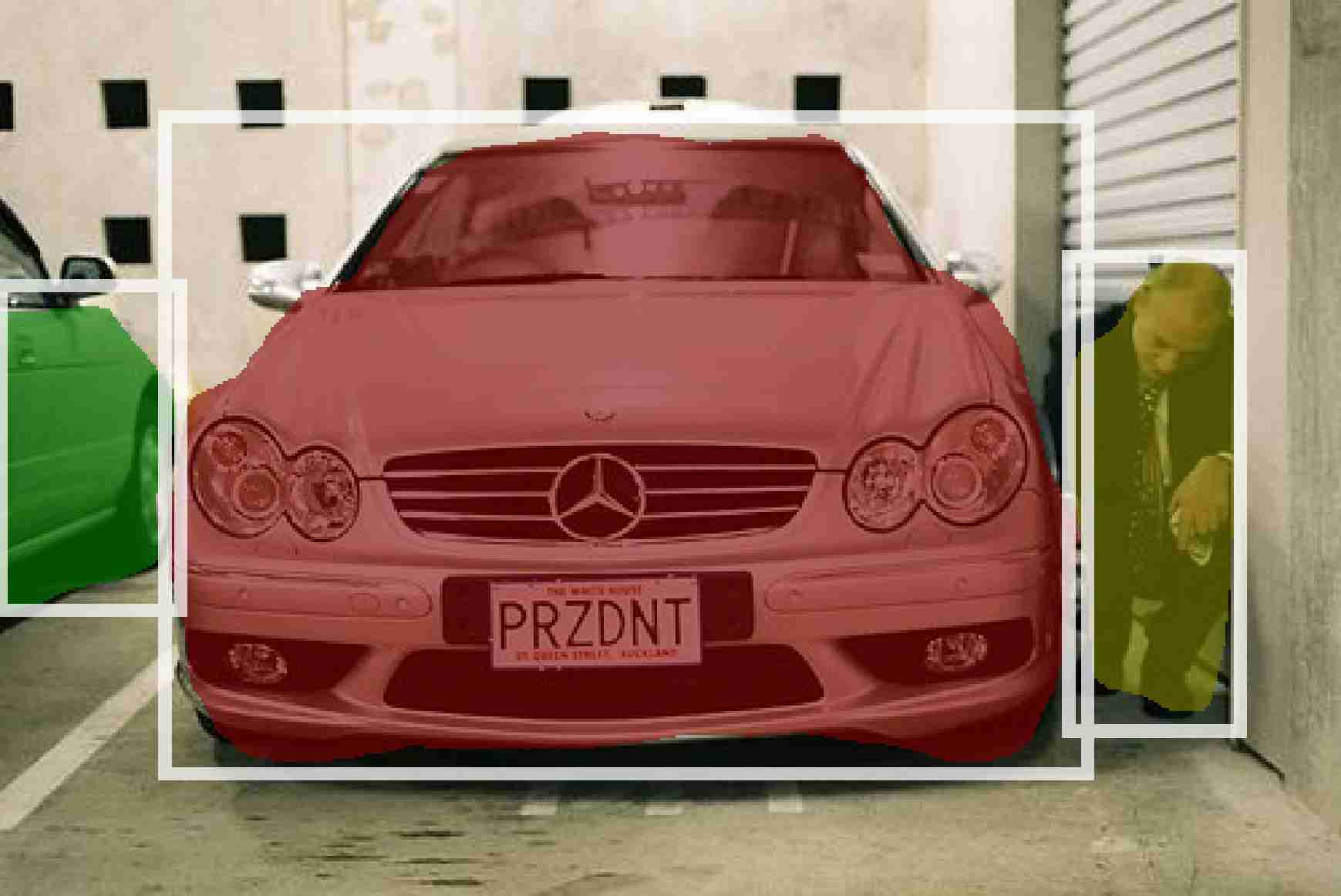} & \hspace*{0.1em} & \includegraphics[width=0.19\textwidth,height=0.09\textheight]{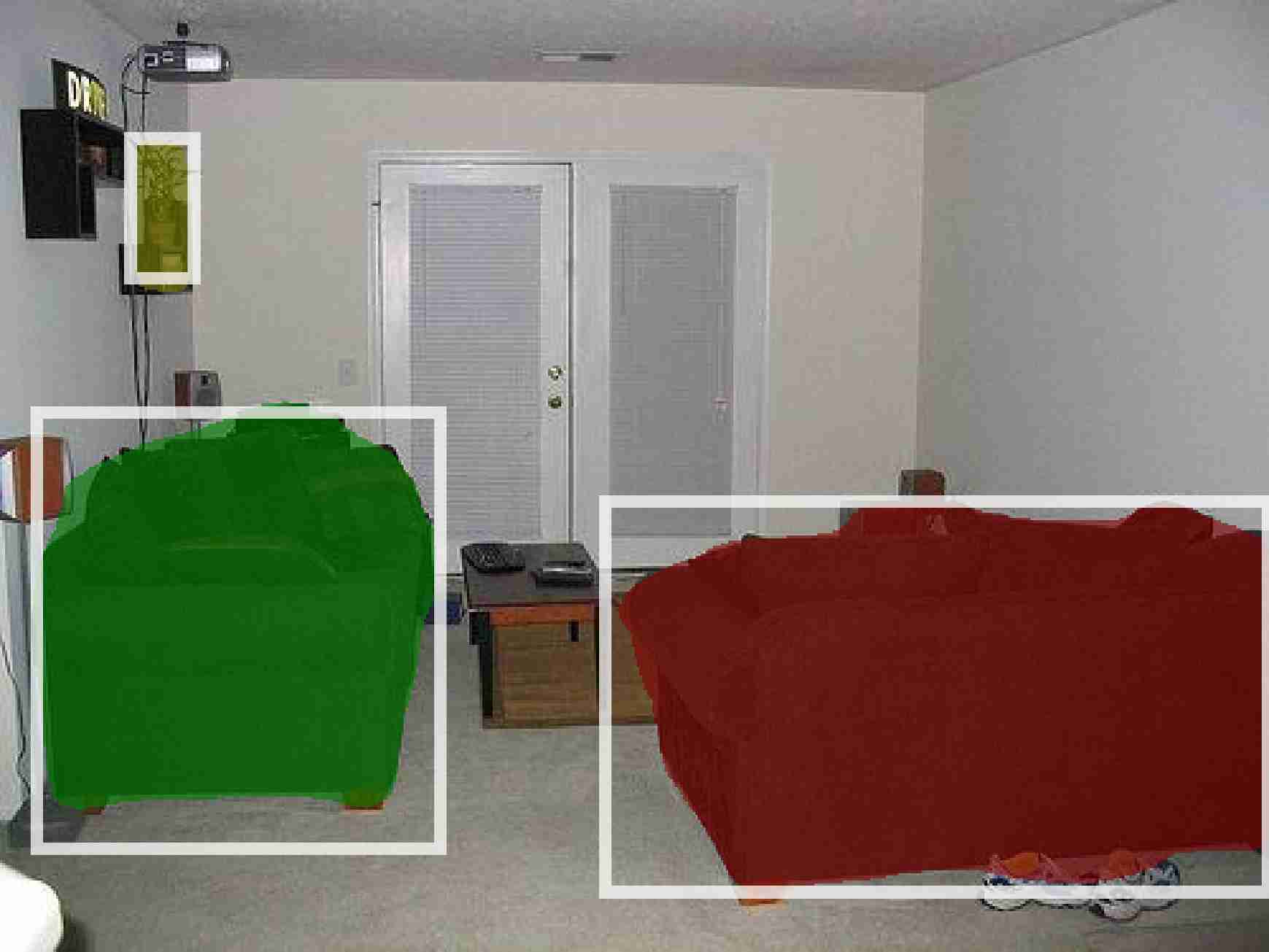}\tabularnewline
\includegraphics[width=0.19\textwidth,height=0.09\textheight]{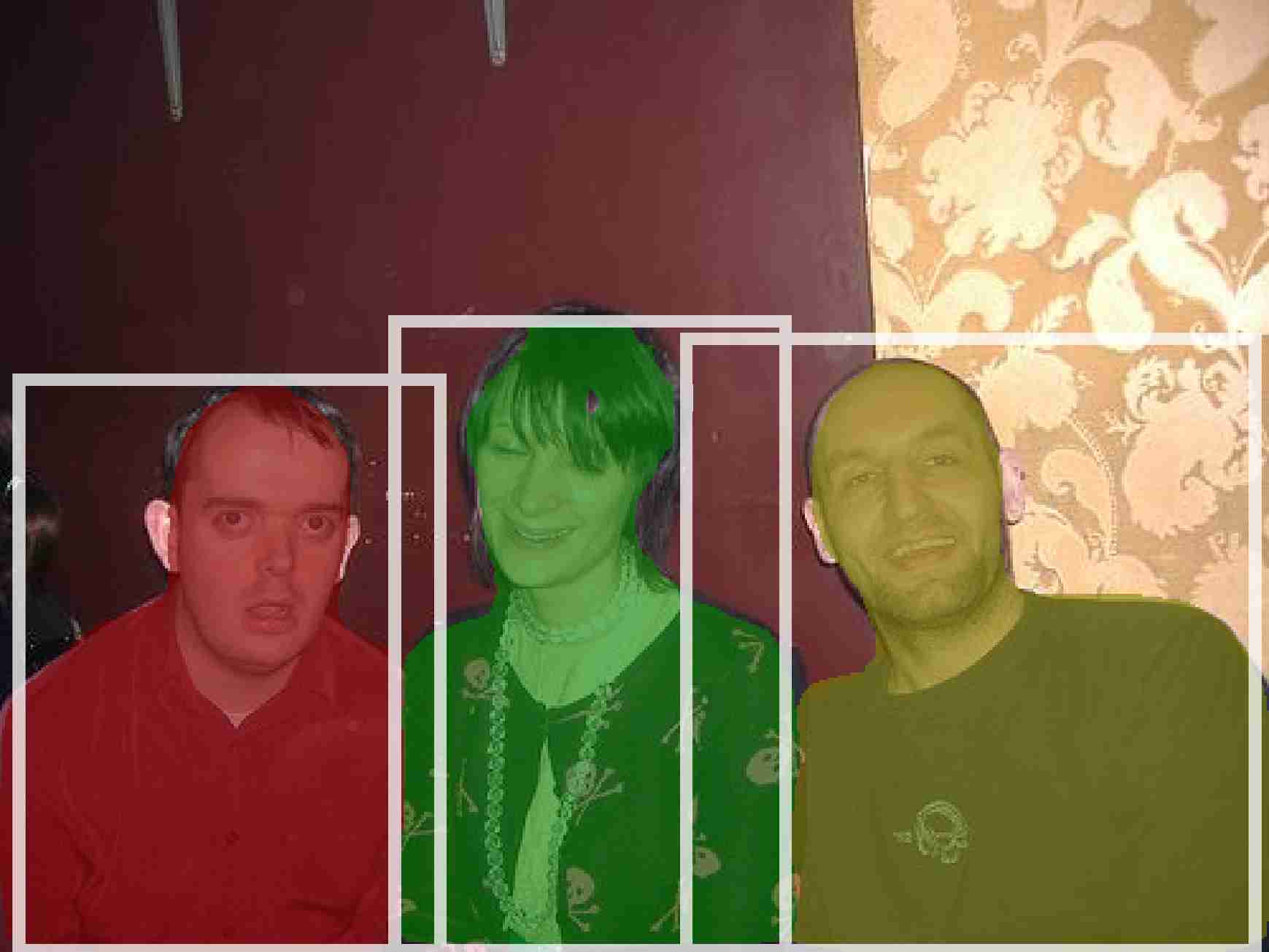} & \hspace*{0.1em} & \includegraphics[width=0.19\textwidth,height=0.09\textheight]{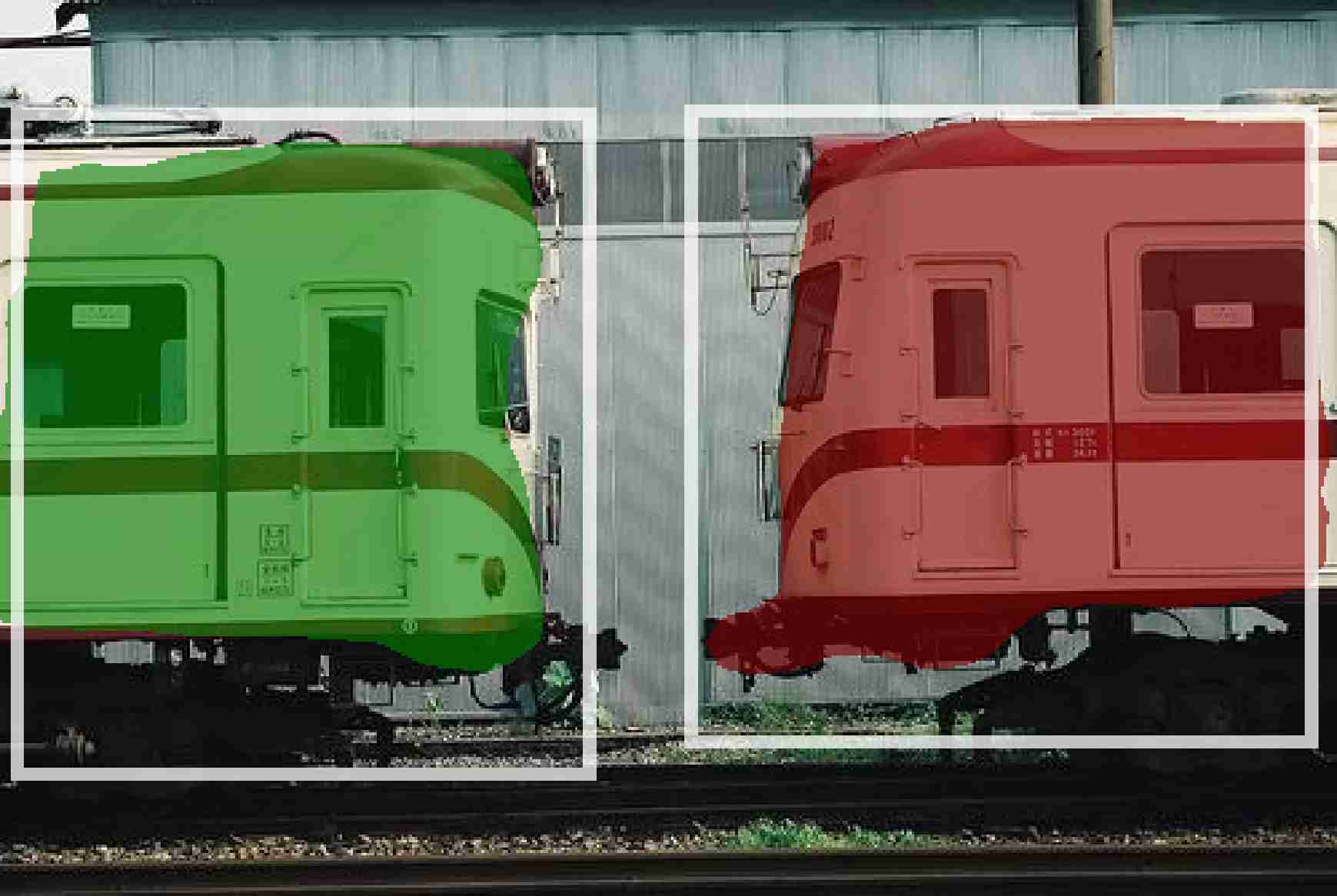} & \hspace*{0.1em} & \includegraphics[width=0.19\textwidth,height=0.09\textheight]{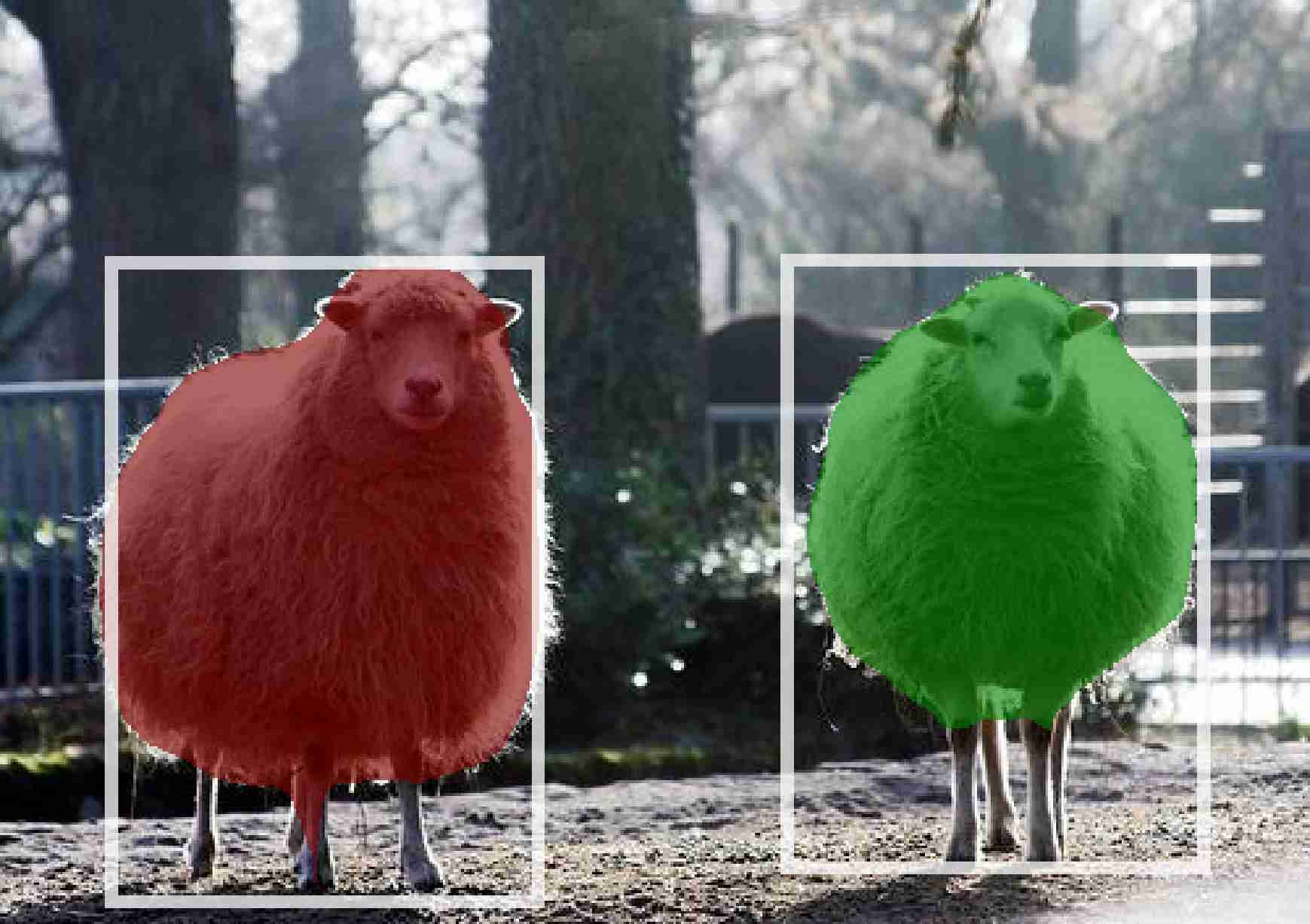} & \hspace*{0.1em} & \includegraphics[width=0.19\textwidth,height=0.09\textheight]{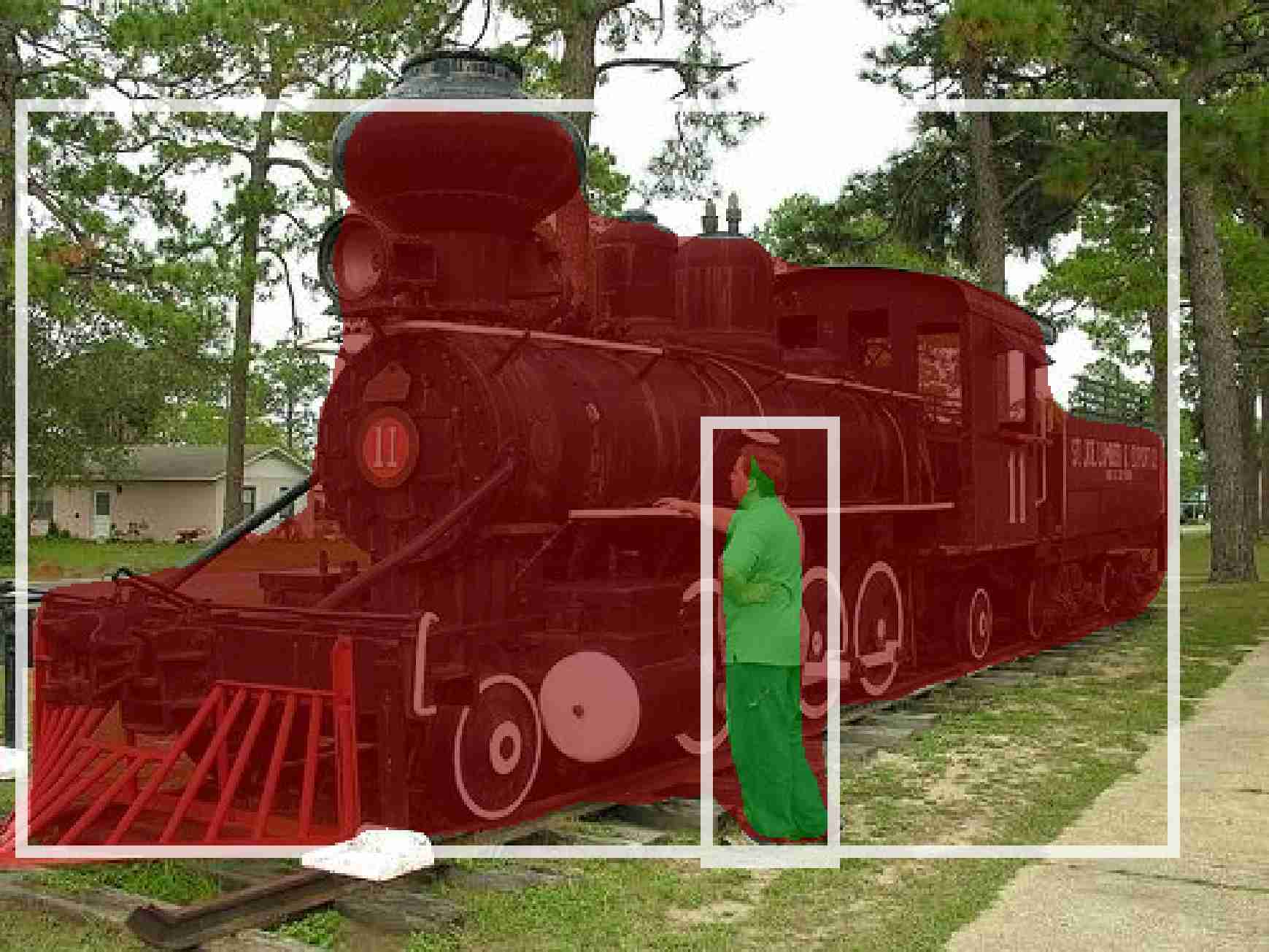} & \hspace*{0.1em} & \includegraphics[width=0.19\textwidth,height=0.09\textheight]{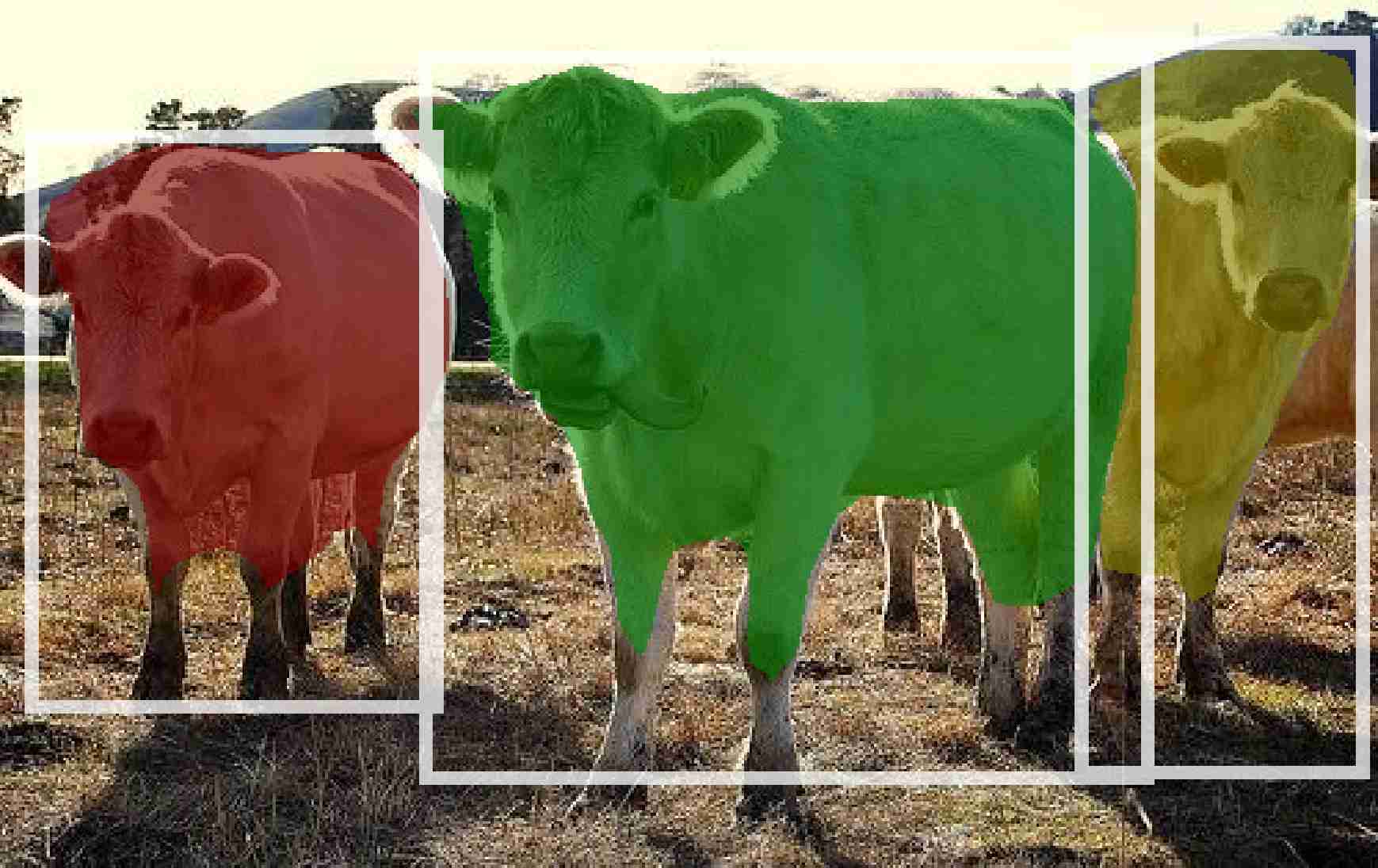}\tabularnewline
\includegraphics[width=0.19\textwidth,height=0.09\textheight]{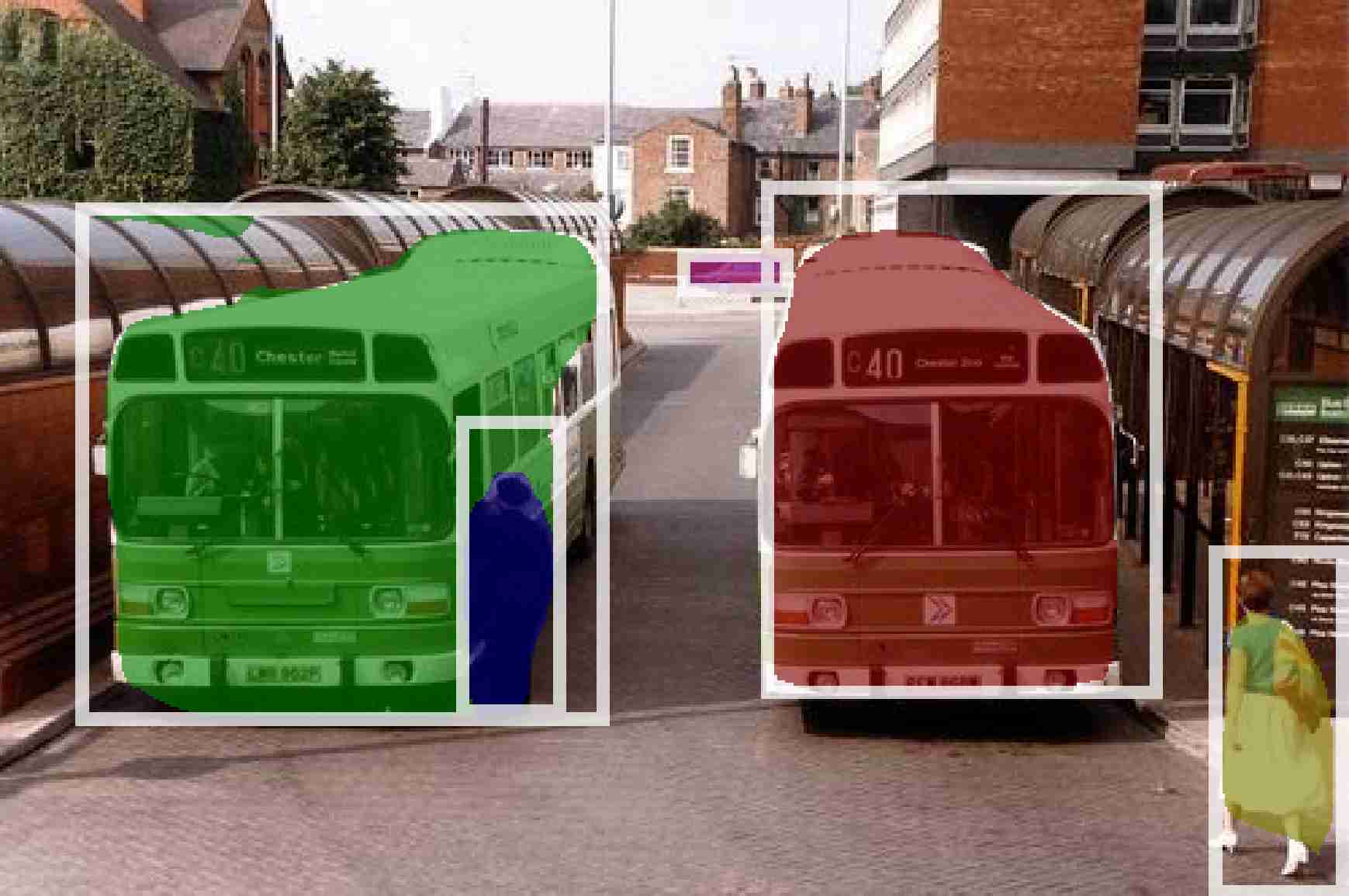} & \hspace*{0.1em} & \includegraphics[width=0.19\textwidth,height=0.09\textheight]{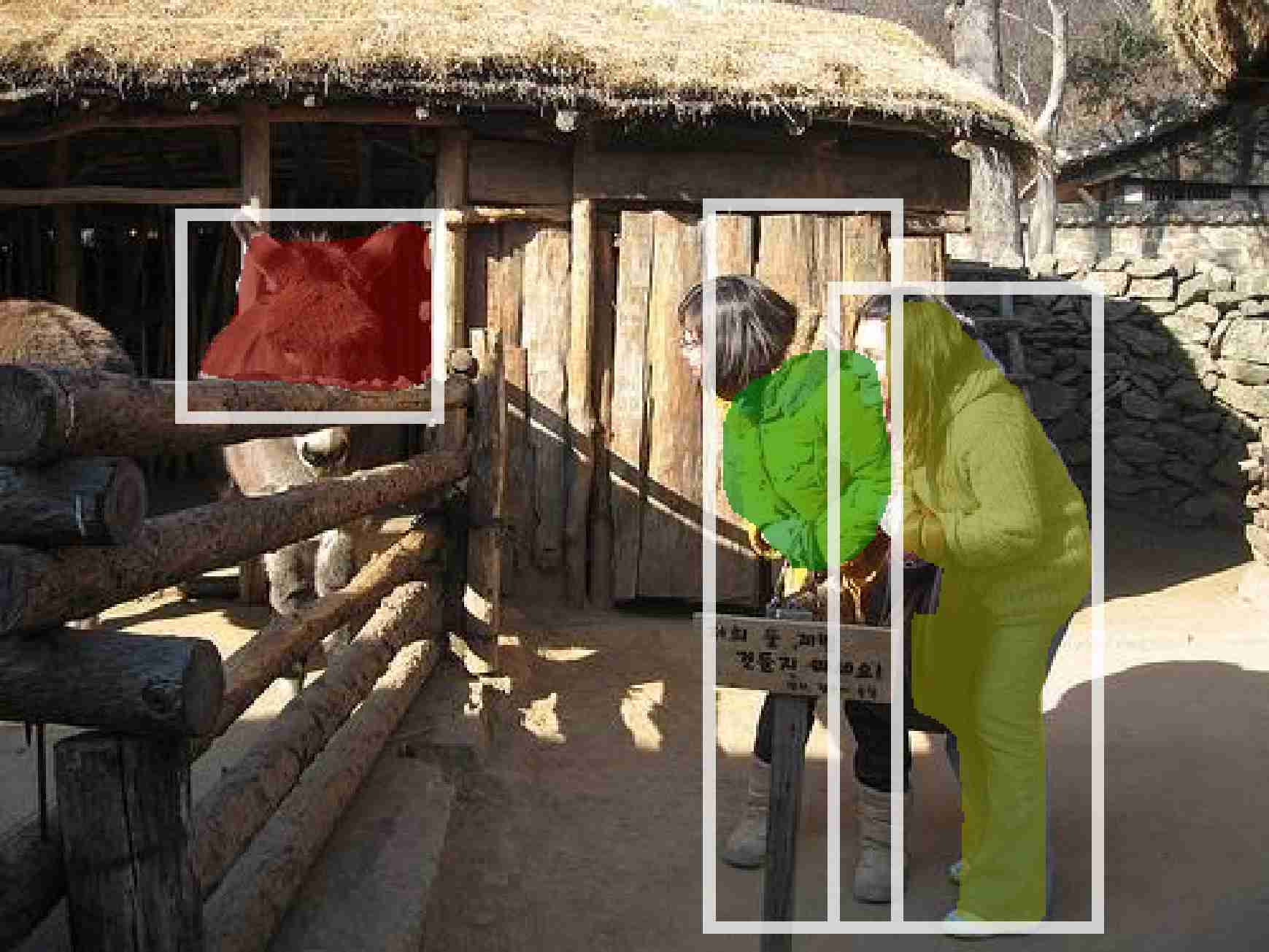} & \hspace*{0.1em} & \includegraphics[width=0.19\textwidth,height=0.09\textheight]{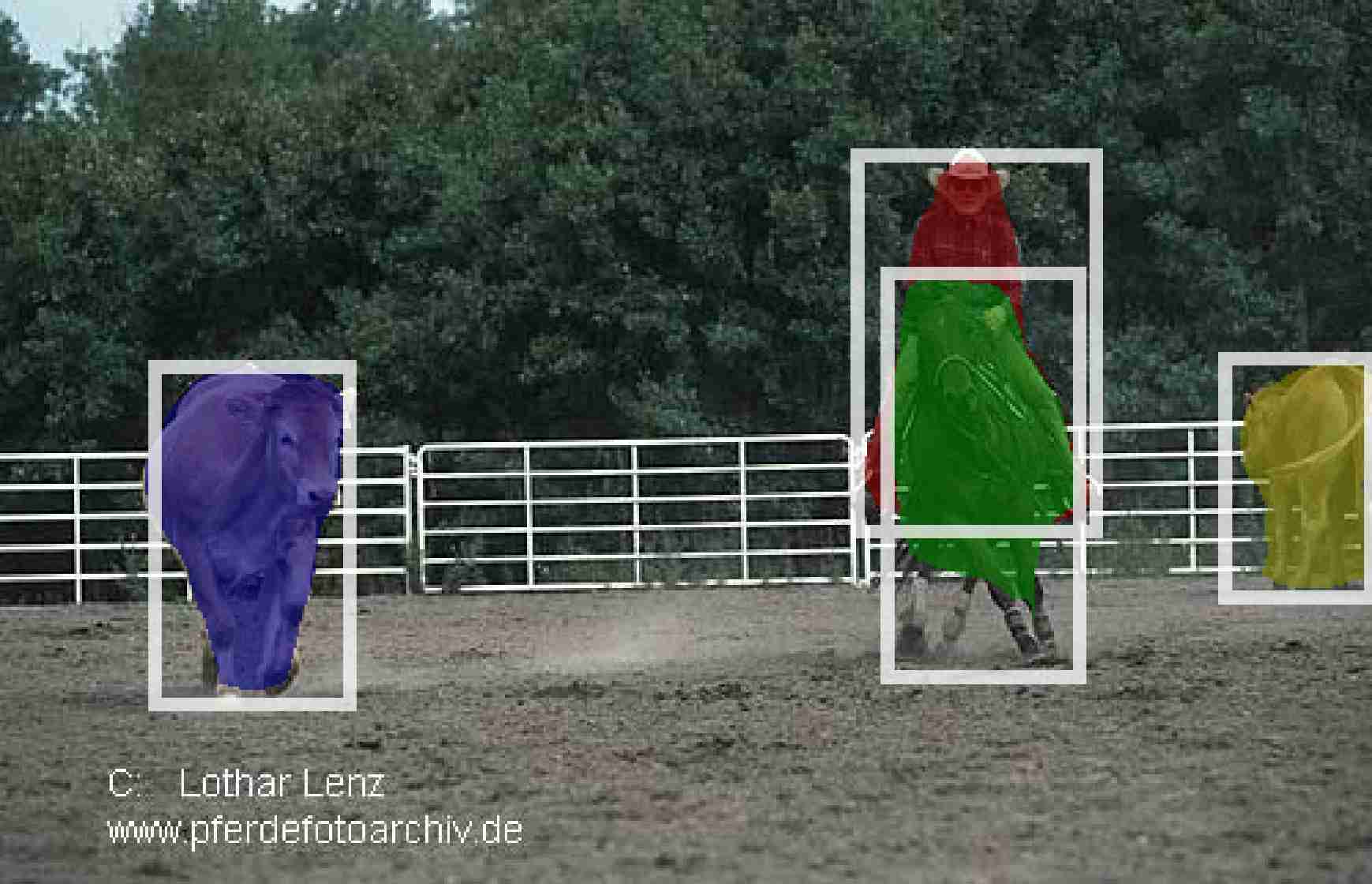} & \hspace*{0.1em} & \includegraphics[width=0.19\textwidth,height=0.09\textheight]{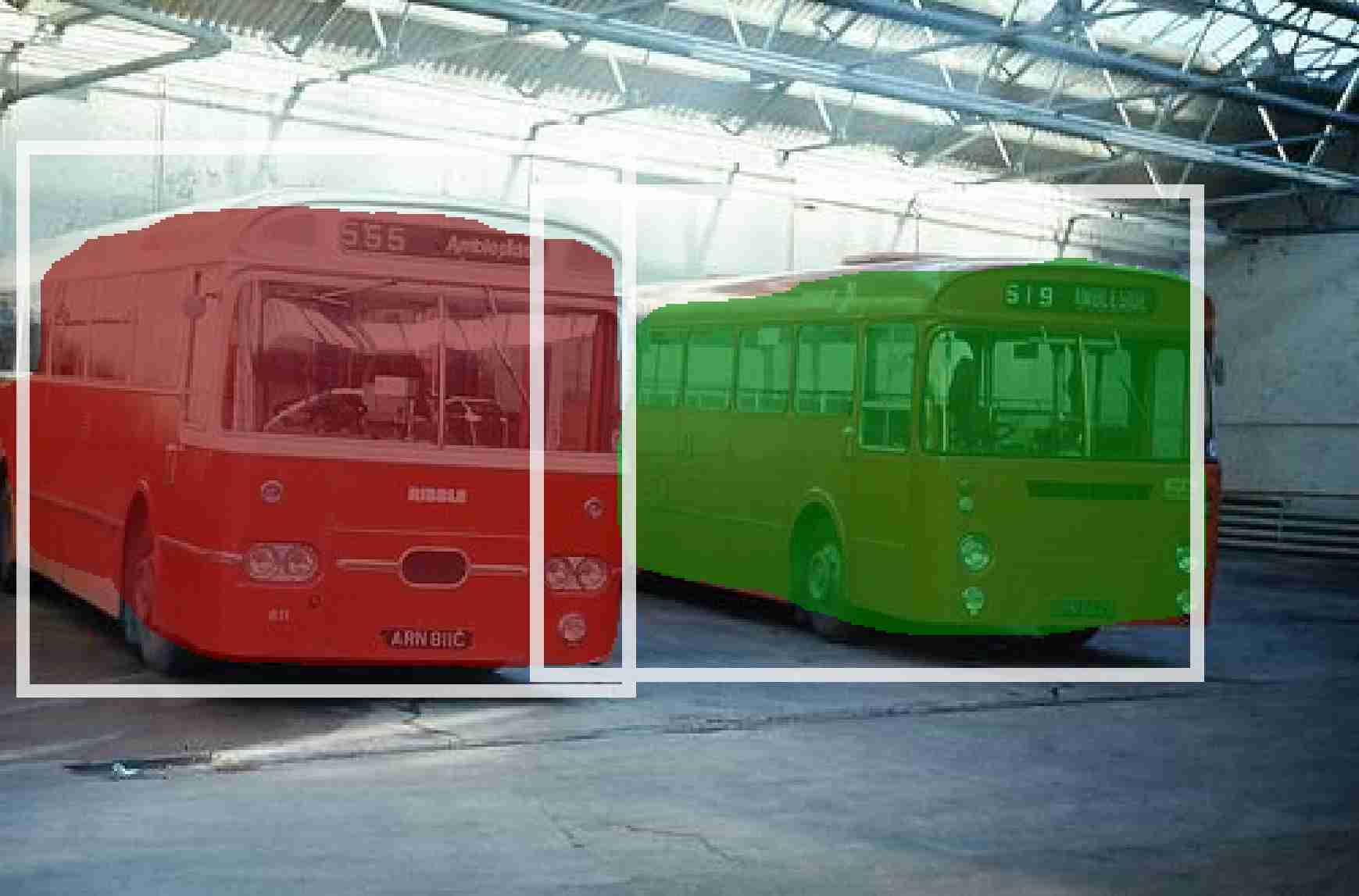} & \hspace*{0.1em} & \includegraphics[width=0.19\textwidth,height=0.09\textheight]{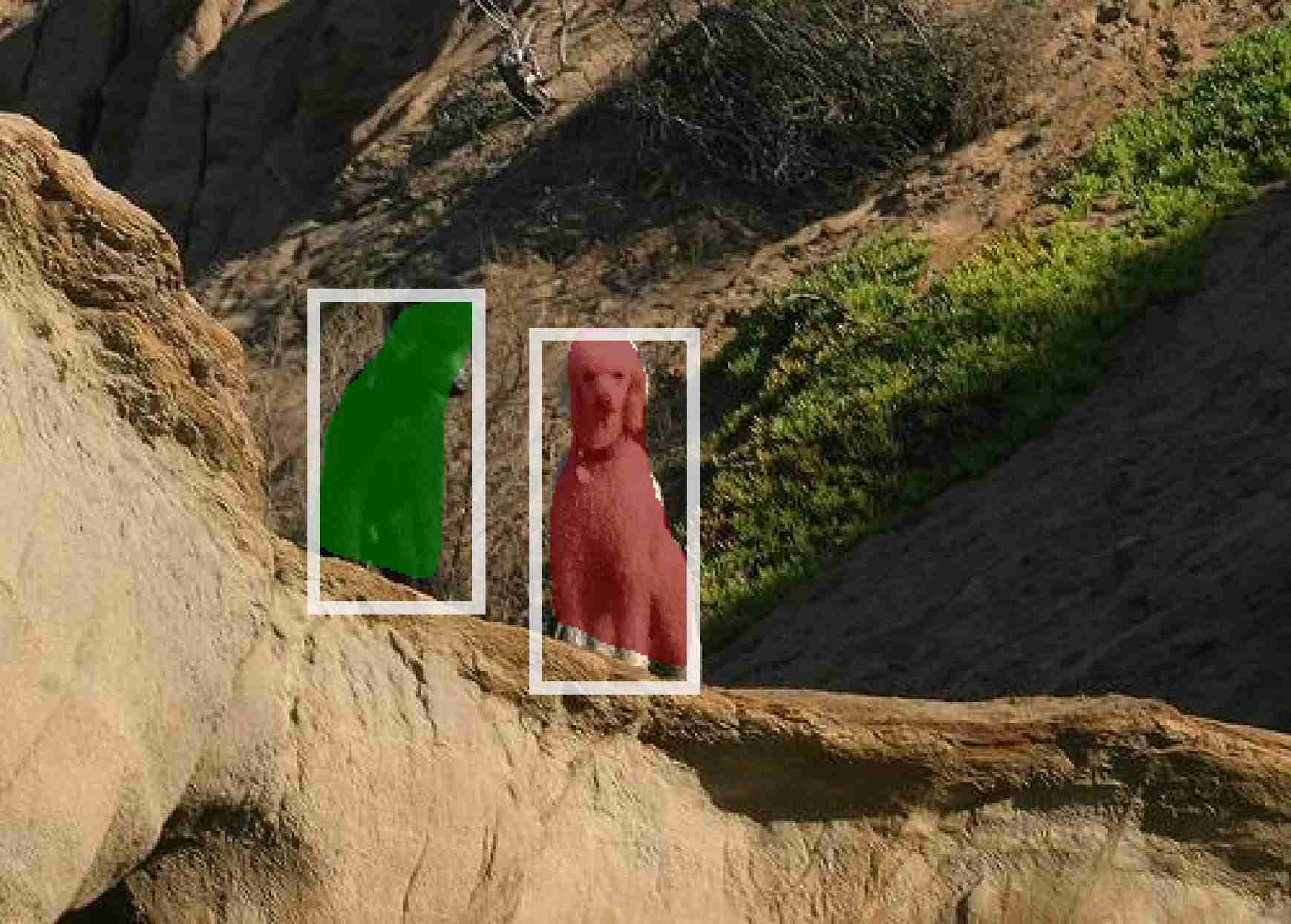}\tabularnewline
\includegraphics[width=0.19\textwidth,height=0.15\textheight]{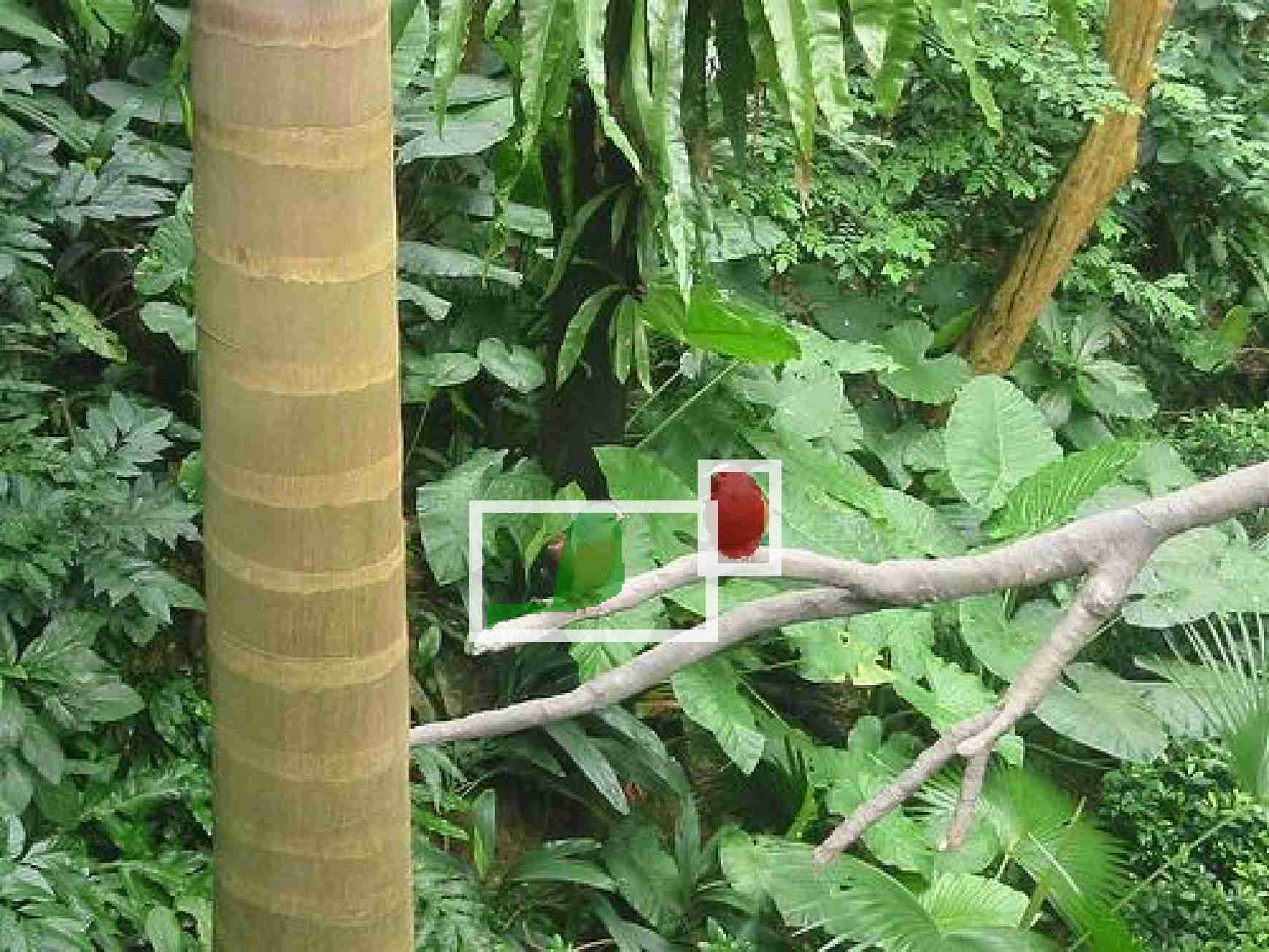} & \hspace*{0.1em} & \includegraphics[width=0.19\textwidth,height=0.15\textheight]{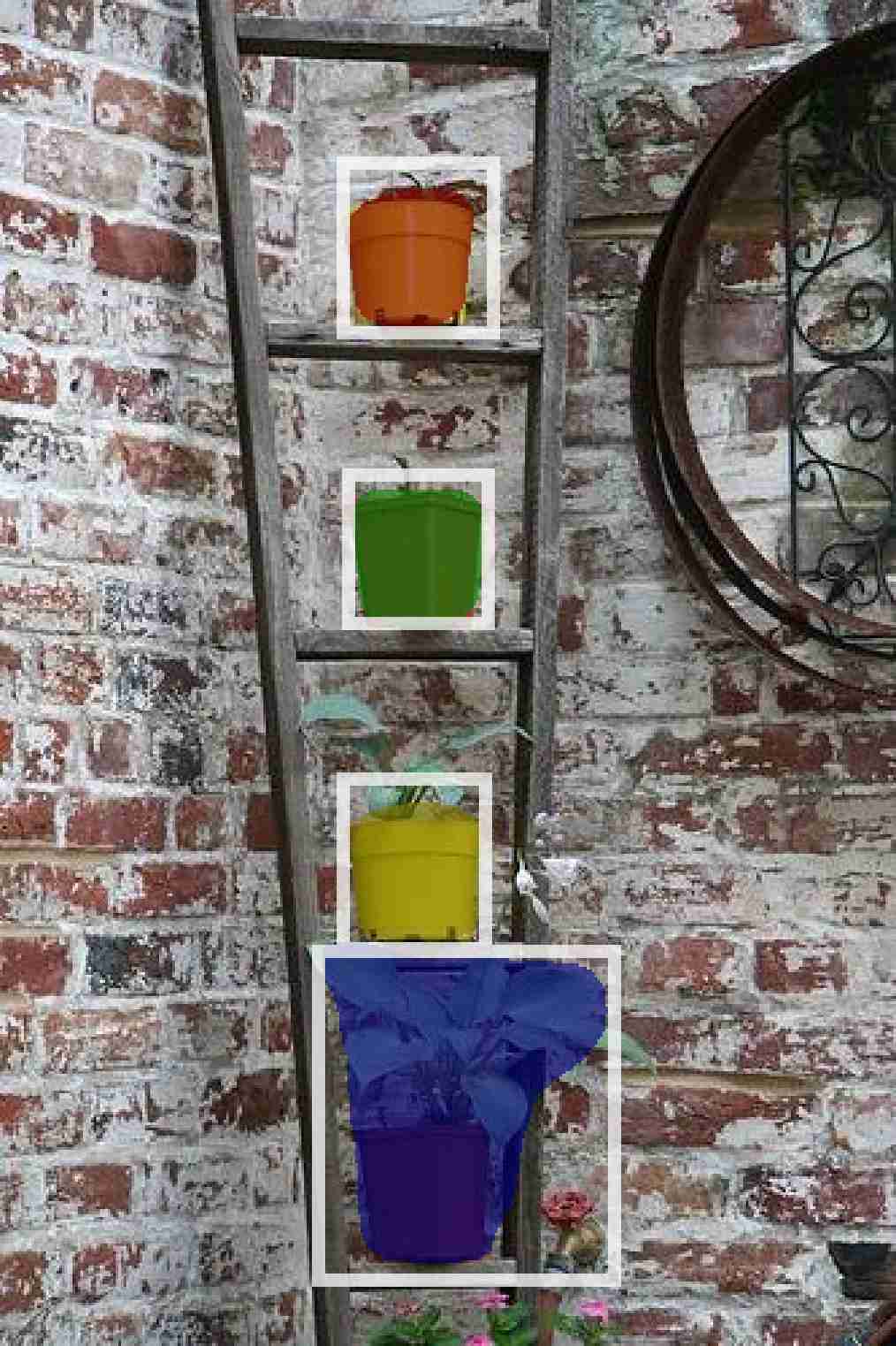} & \hspace*{0.1em} & \includegraphics[width=0.19\textwidth,height=0.15\textheight]{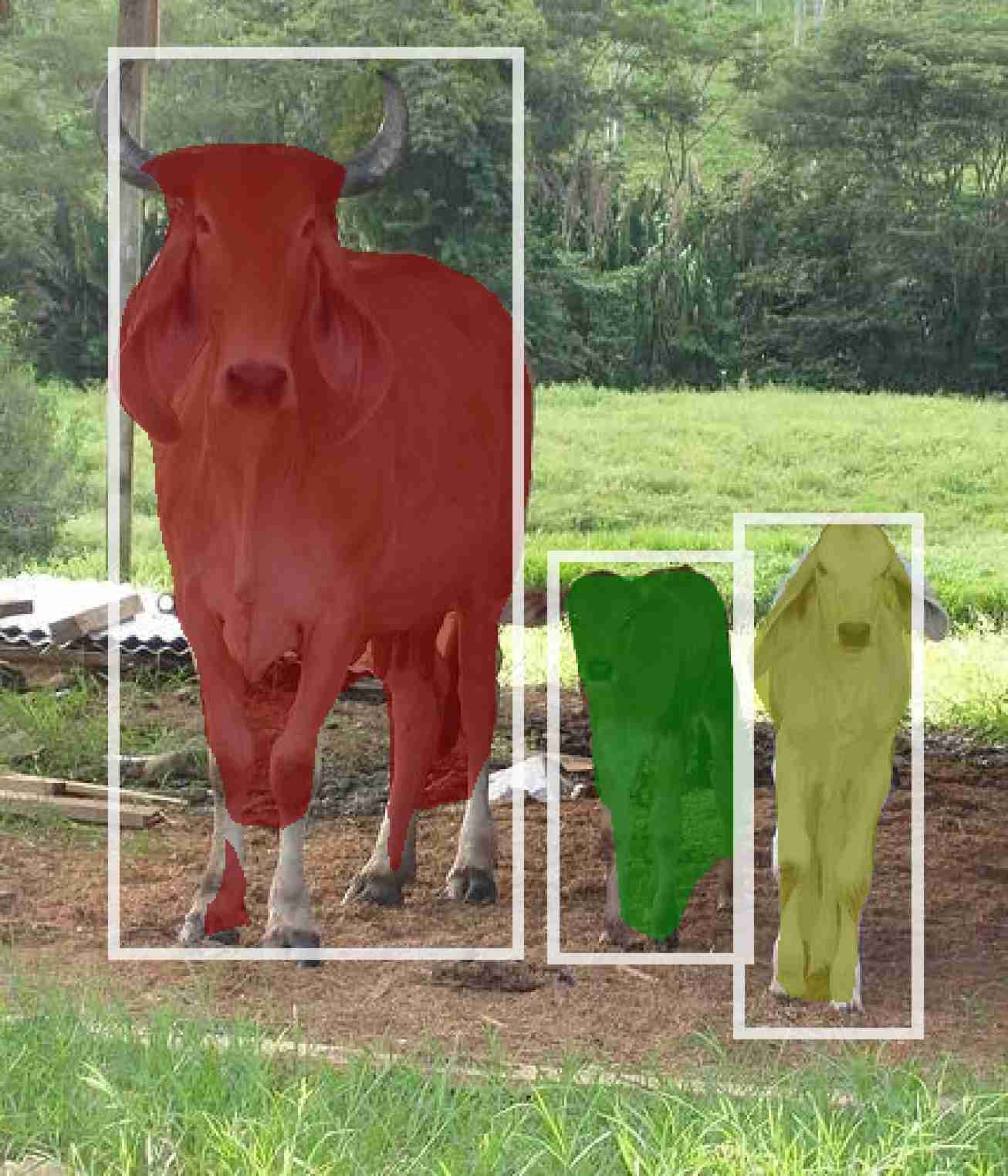} & \hspace*{0.1em} & \includegraphics[width=0.19\textwidth,height=0.15\textheight]{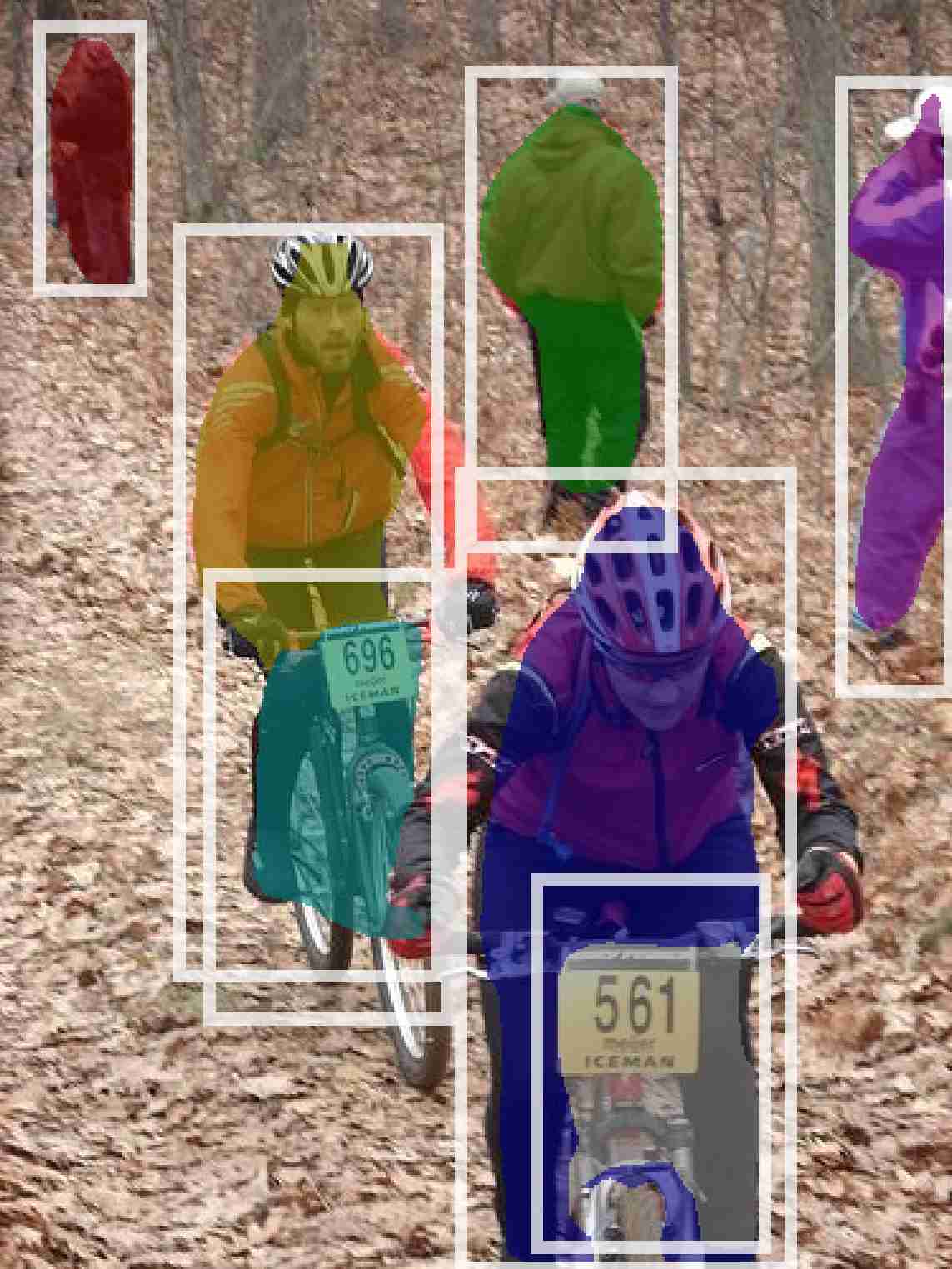} & \hspace*{0.1em} & \includegraphics[width=0.19\textwidth,height=0.15\textheight]{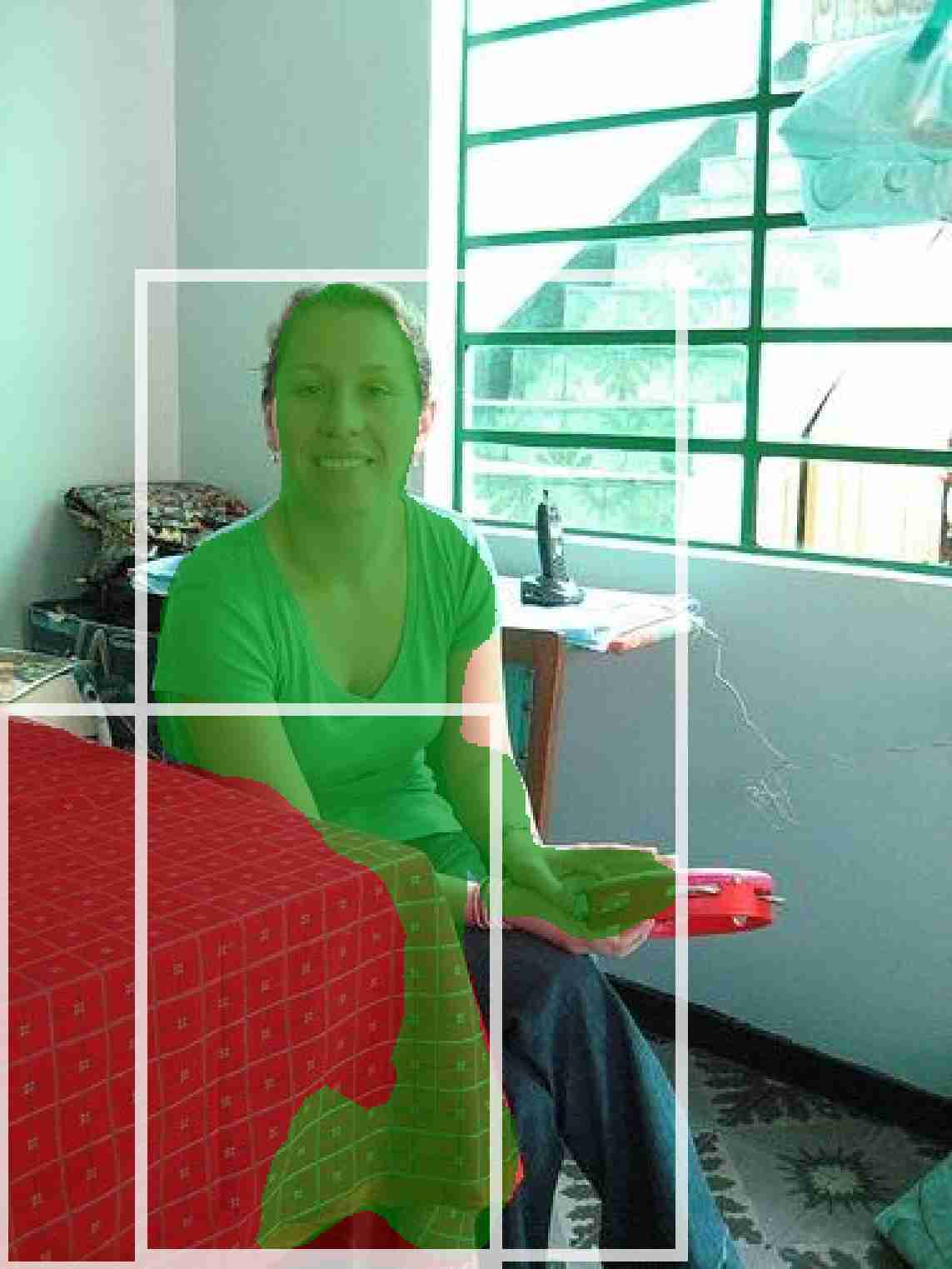}\tabularnewline
\vspace{0.1em}
 &  &  &  &  &  &  &  & \tabularnewline
\multicolumn{9}{c}{ $\mathrm{DeepLab_{BOX}}$}\tabularnewline
\vspace{0.1em}
 &  &  &  &  &  &  &  & \tabularnewline
 &  &  &  &  &  &  &  & \tabularnewline
\end{tabular}\hspace*{\fill}

\caption{\label{fig:Qualitative-results-1-1}Example results from the $\mathrm{DeepMask}$
and $\mathrm{DeepLab_{BOX}}$ models trained with Pascal VOC12 and
COCO using box supervision. White boxes illustrate Fast-RCNN detection
proposals used to output the segments which have the best overlap
with the ground truth segmentation mask.}
\vspace{-1em}
}\endgroup
\end{figure*}
\begin{figure*}[t]
\begingroup{
\setlength{\tabcolsep}{0pt} 
\renewcommand{\arraystretch}{0.2}
\captionsetup[subfigure]{labelformat=empty,font=scriptsize}

\hspace*{\fill}%
\begin{tabular}[b]{ccccccccccccc}
\includegraphics[width=0.14\textwidth]{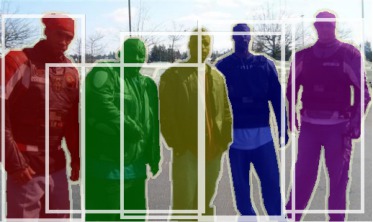} & \hspace*{0.1em} & \includegraphics[width=0.14\textwidth]{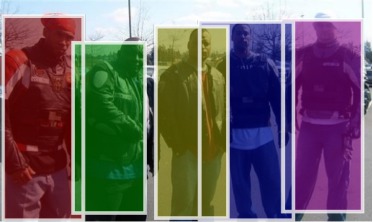} & \hspace*{0.1em} & \includegraphics[width=0.14\textwidth]{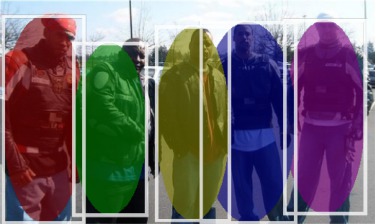} & \hspace*{0.1em} & \includegraphics[width=0.14\textwidth]{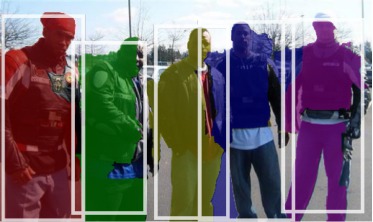} & \hspace*{0.1em} & \includegraphics[width=0.14\textwidth]{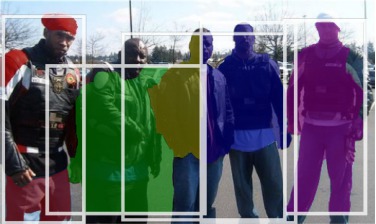} & \hspace*{0.1em} & \includegraphics[width=0.14\textwidth]{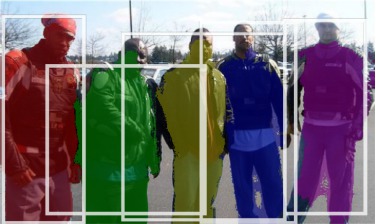} & \hspace*{0.1em} & \includegraphics[width=0.14\textwidth]{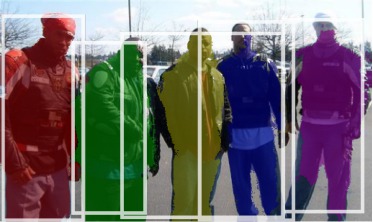}\tabularnewline
\includegraphics[width=0.14\textwidth]{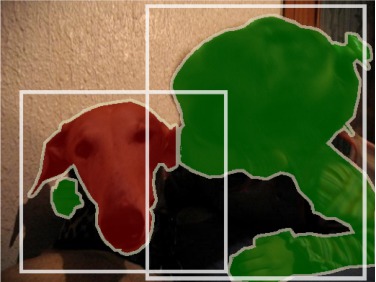} & \hspace*{0.1em} & \includegraphics[width=0.14\textwidth]{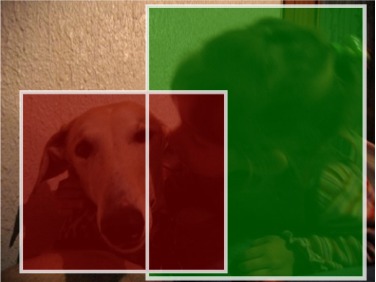} & \hspace*{0.1em} & \includegraphics[width=0.14\textwidth]{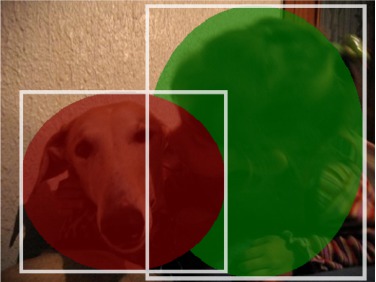} & \hspace*{0.1em} & \includegraphics[width=0.14\textwidth]{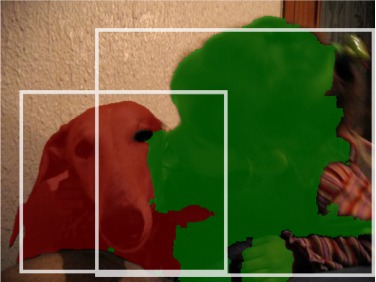} & \hspace*{0.1em} & \includegraphics[width=0.14\textwidth]{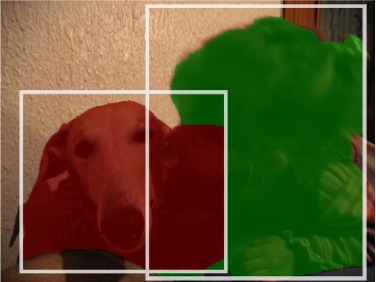} & \hspace*{0.1em} & \includegraphics[width=0.14\textwidth]{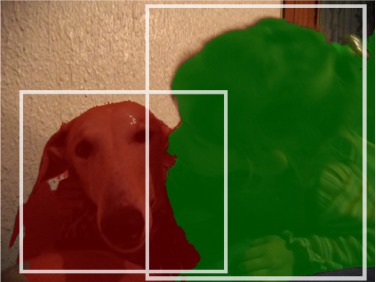} & \hspace*{0.1em} & \includegraphics[width=0.14\textwidth]{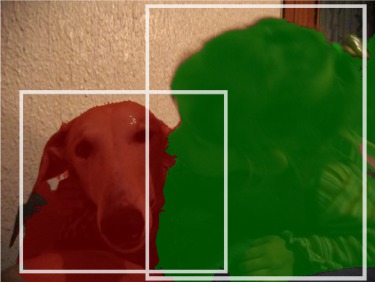}\tabularnewline
\includegraphics[width=0.14\textwidth]{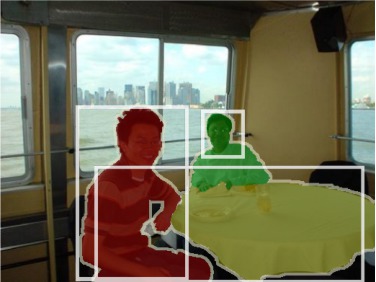} & \hspace*{0.1em} & \includegraphics[width=0.14\textwidth]{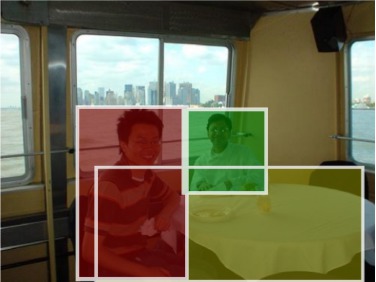} & \hspace*{0.1em} & \includegraphics[width=0.14\textwidth]{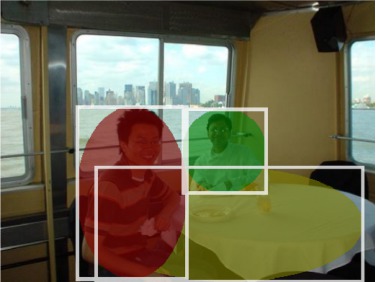} & \hspace*{0.1em} & \includegraphics[width=0.14\textwidth]{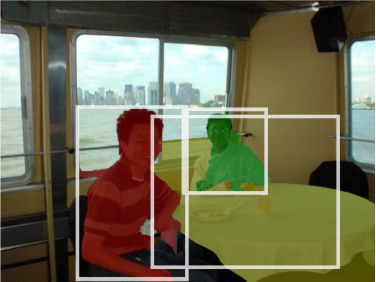} & \hspace*{0.1em} & \includegraphics[width=0.14\textwidth]{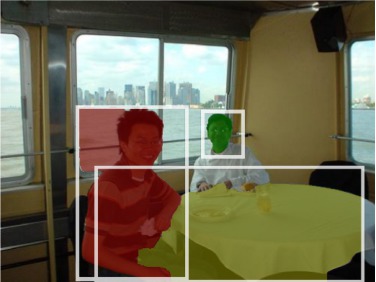} & \hspace*{0.1em} & \includegraphics[width=0.14\textwidth]{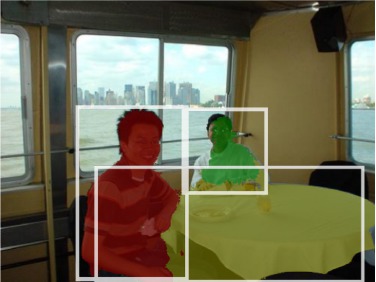} & \hspace*{0.1em} & \includegraphics[width=0.14\textwidth]{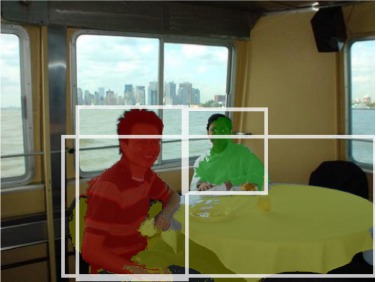}\tabularnewline
\includegraphics[width=0.14\textwidth]{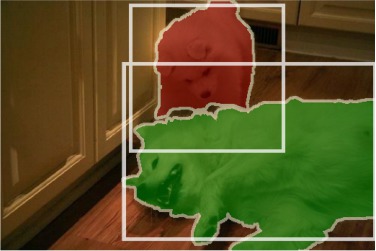} & \hspace*{0.1em} & \includegraphics[width=0.14\textwidth]{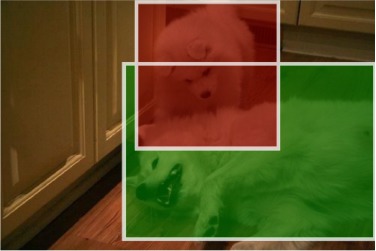} & \hspace*{0.1em} & \includegraphics[width=0.14\textwidth]{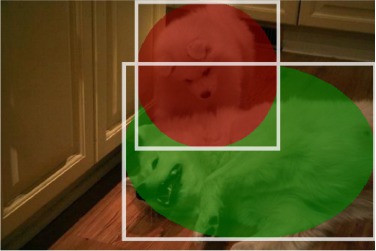} & \hspace*{0.1em} & \includegraphics[width=0.14\textwidth]{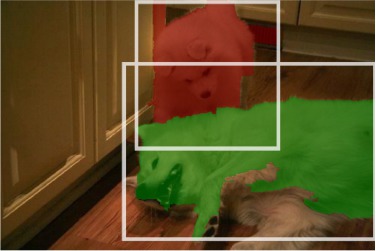} & \hspace*{0.1em} & \includegraphics[width=0.14\textwidth]{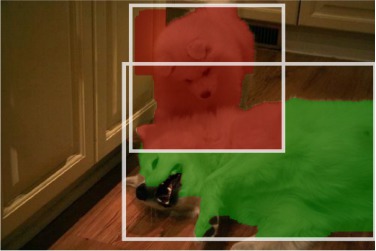} & \hspace*{0.1em} & \includegraphics[width=0.14\textwidth]{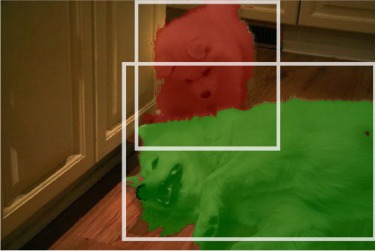} & \hspace*{0.1em} & \includegraphics[width=0.14\textwidth]{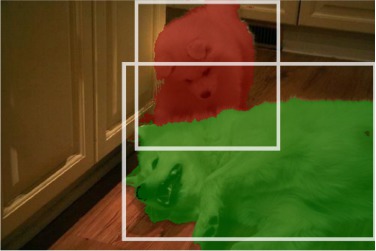}\tabularnewline
\vspace{0.1em}
 &  &  &  &  &  &  &  &  &  &  &  & \tabularnewline
\begin{tabular}{c}
Ground\tabularnewline
truth\tabularnewline
\end{tabular}  &  & Rectangles &  & Ellipse &  & MCG &  & GrabCut+ &  & %
\begin{tabular}{c}
Weakly\tabularnewline
supervised\tabularnewline
\end{tabular} &  & %
\begin{tabular}{c}
Fully\tabularnewline
supervised\tabularnewline
\end{tabular}\tabularnewline
\end{tabular}\hspace*{\fill}

\caption{\label{fig:Qualitative-results-DeepMask}Qualitative results of instance
segmentation on VOC12. Example result from the DeepMask model are
trained with Pascal VOC12 and COCO supervision. White boxes illustrate
Fast-RCNN detection proposals used to output the segments which have
the best overlap with the ground truth segmentation mask.}
\vspace{-1em}
}\endgroup
\end{figure*}

\end{document}